%% file: main.tex
\documentclass[10pt,twocolumn,letterpaper]{article}

\usepackage[final]{main/cvpr}

\input{main/mypackages}

\definecolor{cvprblue}{rgb}{0.21,0.49,0.74}
\usepackage[pagebackref,breaklinks,colorlinks,citecolor=cvprblue]{hyperref}

\def\paperID{880}
\def\confName{CVPR}
\def\confYear{2025}

\title{Functionality understanding and segmentation in 3D scenes}
\author{
\begin{minipage}{0.19\textwidth}
    \vspace*{-7mm}
    \centering
    Jaime Corsetti$^{1,2}$
\end{minipage}
\begin{minipage}{0.19\textwidth}
    \vspace*{-7mm}
    \centering
    Francesco Giuliari$^1$
\end{minipage}
\begin{minipage}{0.19\textwidth}
    \vspace*{-7mm}
    \centering
    Alice Fasoli$^1$
\end{minipage}
\begin{minipage}{0.19\textwidth}
    \vspace*{-7mm}
    \centering
    Davide Boscaini$^1$
\end{minipage}
\begin{minipage}{0.16\textwidth}
    \vspace*{-7mm}
    \centering
    Fabio Poiesi$^1$
\end{minipage} \\
\begin{minipage}[t]{0.45\textwidth}
\centering
     $^1$Fondazione Bruno Kessler
\end{minipage}
\hfill
\begin{minipage}[t]{0.45\textwidth}
\centering
    $^2$University of Trento
\end{minipage} \\
\begin{minipage}{\textwidth}
    \vspace*{4mm}
    \centering
    \tt \small \{jcorsetti,fgiuliari,alfasoli,dboscaini,poiesi\}@fbk.eu
\end{minipage}
}
\begin{document}

\include{main/mycommands}

% \maketitle

\twocolumn[{%
    \renewcommand\twocolumn[1][]{#1}%
    \vspace{-1cm}
    \maketitle
    \thispagestyle{empty}
    \input{main/figures/teaser/teaser}

}]

\input{main/sections/0_abstract}
\input{main/sections/1_introduction}

\input{main/sections/2_related}
\input{main/sections/3_method}

\input{main/sections/4_results}
\input{main/sections/5_conclusion}

{
    \small
    \bibliographystyle{main/ieeenat_fullname}
    \bibliography{main}
}

\end{document}

\include{supp/mycommands}

\maketitle

\input{supp/sections/intro}
\input{supp/sections/0_baselines}

\input{supp/sections/1_ablations}
\input{supp/sections/2_qualitatives}
\input{supp/sections/3_implementation}
{
    \small
    \bibliographystyle{supp/ieeenat_fullname}
    \bibliography{supp}
}

%% file: main/mypackages.tex
\usepackage[dvipsnames]{xcolor}
\usepackage{amsmath,amssymb,amsfonts}
\usepackage{multirow,multicol}
\usepackage{natbib}
\usepackage{algorithmic}
\usepackage{graphicx}
\usepackage{textcomp}
\usepackage{xspace}
\usepackage{url}
\usepackage{dsfont}
\usepackage{booktabs}
\usepackage{balance}
\usepackage{soul}
\usepackage{todonotes}
\usepackage{pifont}
\usepackage{colortbl}
\usepackage{overpic}
\usepackage{enumitem}
\usepackage{nccmath}
\usepackage{courier}
\usepackage{array}% http://ctan.org/pkg/array
\usepackage{tabularray}
\usepackage{pifont}% http://ctan.org/pkg/pifont
\usepackage[accsupp]{axessibility} % Improves PDF readability for those with visual impairments.
\usepackage{pgfplots}
\pgfplotsset{compat=1.18}
\usepackage{comment}
\usepackage{microtype}
\usepackage{transparent}

%% file: main/mycommands.tex
\newcommand{\higherbetter}[0]{{\color{black!50}{$\,\uparrow$}}}
\newcommand{\oracle}[1]{\textcolor{gray}{#1}}

\definecolor{visual}{HTML}{3399FF}
\definecolor{text}{HTML}{97D077}

\newcommand{\impp}[1]{{\textcolor{Green}{+#1}}}
\newcommand{\impn}[1]{{\textcolor{BrickRed}{-#1}}}

\newcommand{\correct}[1]{{\textcolor{Green}{#1}}}
\newcommand{\wrong}[1]{{\textcolor{BrickRed}{#1}}}

\newcommand{\acronym}{Fun3DU\xspace}
\definecolor{myazure}{rgb}{0.8509,0.8980,0.9412}
\newcommand{\cmark}{\ding{51}}%
\newcommand{\xmark}{\ding{55}}%

\newcommand{\assistant}[0]{vision and language model\xspace}
\newcommand{\ass}[0]{VLM\xspace}

\newcommand{\pcd}[0]{$\mathcal{C}$\xspace}
\newcommand{\allviews}[0]{$\mathcal{V}$\xspace}
\newcommand{\funmask}[0]{$\mathcal{M}$\xspace}

\newcommand{\descr}[0]{\texttt{D}\xspace}
\newcommand{\funobj}[0]{\texttt{F}\xspace}
\newcommand{\parobj}[0]{\texttt{O}\xspace}

\newcommand{\valid}[0]{split0\xspace}
\newcommand{\train}[0]{split1\xspace}

%% file: main/figures/teaser/teaser.tex
\begin{center}
\includegraphics[width=1.0\linewidth]{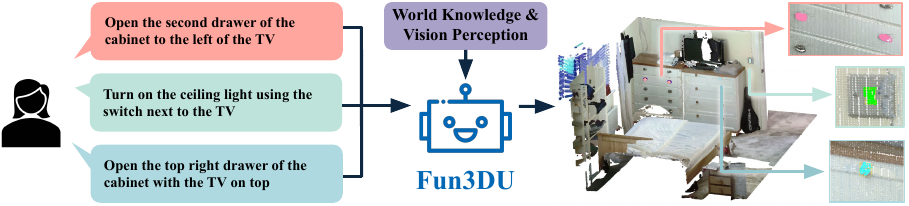}

\vspace{-2mm}
\captionof{figure}{
We present \acronym, the first method for functionality understanding and segmentation in 3D scenes.
\acronym interprets natural language descriptions (left-hand side) in order to segment functional objects in real-world 3D environments (right-hand side).
\acronym relies on world knowledge and vision perception capabilities of pre-trained vision and language models, without requiring 
task-specific finetuning.
}
\label{fig:teaser}
\end{center}

%% file: main/sections/0_abstract.tex
\begin{abstract}
Understanding functionalities in 3D scenes involves interpreting natural language descriptions to locate functional interactive objects, such as handles and buttons, in a 3D environment. 
Functionality understanding is highly challenging, as it requires both world knowledge to interpret language and spatial perception to identify fine-grained objects. 
For example, given a task like `turn on the ceiling light,' an embodied AI agent must infer that it needs to locate the light switch, even though the switch is not explicitly mentioned in the task description.
To date, no dedicated methods have been developed for this problem. 
In this paper, we introduce \acronym, the first approach designed for functionality understanding in 3D scenes. 
\acronym uses a language model to parse the task description through Chain-of-Thought reasoning in order to identify the object of interest. 
The identified object is segmented across multiple views of the captured scene by using a \assistant. 
The segmentation results from each view are lifted in 3D and aggregated into the point cloud using geometric information.
\acronym is training-free, relying entirely on pre-trained models.
We evaluate \acronym on SceneFun3D, the most recent and only dataset to benchmark this task, which comprises over 3000 task descriptions on 230 scenes. 
Our method significantly outperforms state-of-the-art open-vocabulary 3D segmentation approaches.
Project page: \url{https://tev-fbk.github.io/fun3du/}
\end{abstract}

%% file: main/sections/1_introduction.tex
%%%%%%%%%%%%%%%%%%%%%%%%%%%%%%%%%%%%%%%%%%%%%%%%%%%%%%%%%%%%%%%%%%%%%%%
%%%%%%%%%%%%%%%%%%%%%%%%%%%%%%%%%%%%%%%%%%%%%%%%%%%%%%%%%%%%%%%%%%%%%%%
%%%%%%%%%%%%%%%%%%%%%%%%%%%%%%%%%%%%%%%%%%%%%%%%%%%%%%%%%%%%%%%%%%%%%%%
\section{Introduction}\label{sec:introduction}

Functionality understanding in 3D scenes requires interpreting natural language descriptions in order to locate functional interactive elements, such as handles, knobs, and buttons, in a 3D environment~\cite{delitzas2024scenefun3d}. 
For example, given a task like `\texttt{turn on the ceiling light}', the agent must first understand that the functional goal is to increase the room’s brightness through a light switch, and then locate the switch in 3D by segmentation.
While functionality understanding has long been a research challenge \cite{Winston1983}, we argue that this dual requirement, i.e.~integrating language comprehension with spatial perception, still poses several unaddressed problems, especially when applied to 3D data.
A major bottleneck is the limited availability of datasets with fine-grained mask annotations for real-world environments focused on interactive functional objects.
The recent release of SceneFun3D~\cite{delitzas2024scenefun3d} contributes to address this gap, providing high-resolution point clouds and multi-view images of real-world domestic 3D scenes annotated with functionality masks. 
However, no existing method tackles this unique challenge.

In this paper, we introduce \acronym, the first approach designed specifically for \underline{fun}ctionality \underline{3D} scene \underline{u}nderstanding. 
Since 3D data is orders of magnitude smaller than 2D data, it is insufficient for training models capable of understanding real-world nuances, such as those needed for interpreting object functionalities in 3D scenes. 
Therefore, we design \acronym as a training-free method that leverages pre-trained vision and language models (VLMs) \cite{deitke2024molmo} to comprehend task descriptions and segment functional objects, often not explicitly mentioned in the description.
\acronym is based on four key modules that process multiple views of a given scene and project the results in 3D.
The first module interprets the task description to explain the functionality and context through Chain-of-Thought reasoning \cite{wei2022chain}. 
The second module locates contextual objects via open-vocabulary segmentation \cite{minderer2023owl2} to improve accuracy and efficiency in masking the functional objects within each view.
Moreover, it employs a novel visibility-based view selection approach to reduce the number of views from thousands to tens informative ones.
The third module segments the functional objects on this view subset using a 2D VLM~\cite{deitke2024molmo}. 
The fourth module performs multi-view agreement by lifting and aggregating the 2D masks into the 3D point cloud using point-to-pixel correspondences.
We validate \acronym on SceneFun3D~\cite{delitzas2024scenefun3d}, which includes 230 scenes and over 3000 task descriptions.
SceneFun3D presents challenging real-world scenarios, such as objects with similar appearances (e.g., dresser handles), ambiguous task descriptions that require spatial reasoning (e.g., distinguishing between the top and bottom drawers of a cabinet), and world knowledge (e.g., understanding that switches are typically located near doors).

We adapt open-vocabulary 3D segmentation methods~\cite{takmaz2024openmask3d, huang2023openins3d, kerr2023lerf} to this task and use them as comparison baselines.
\acronym outperforms these baselines by a significant margin (+13.2 mIoU on average), highlighting that 3D functionality understanding requires deeper reasoning capabilities than those provided by current open-vocabulary segmentation approaches.
In summary, our contributions are:
\begin{itemize}[noitemsep,nolistsep,leftmargin=*]
    \item We introduce the first approach specifically designed for functionality understanding in 3D scenes;
    \item We present a novel \textit{training-free} method that leverages pre-trained vision and language models to interpret task descriptions and segment functional objects jointly;
    \item We design a novel view-selection algorithm to make masking of functional objects on images effective.
\end{itemize}

\input{main/figures/diagram/diagram}

%% file: main/figures/diagram/diagram.tex
\begin{figure*}
    \centering
    \includegraphics[width=1\linewidth]{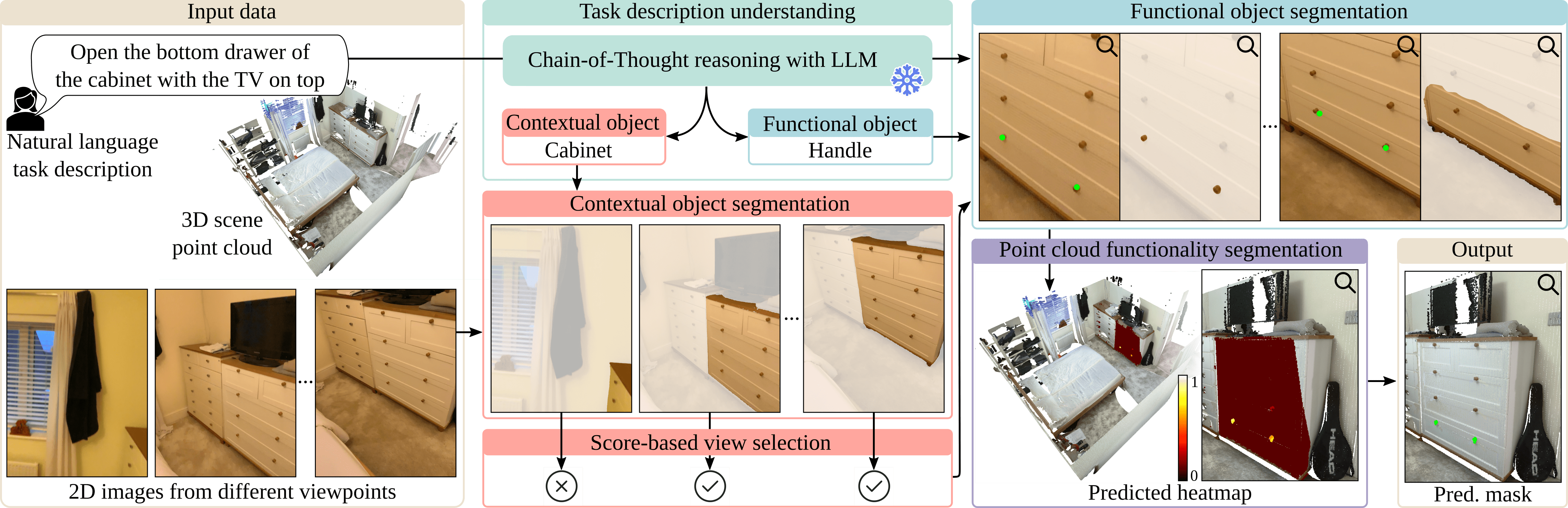}

    \vspace{-2mm}
    \caption{
    \acronym consists of four main modules.
    The first module (green) interprets the natural language task description using Chain-of-Thought reasoning with a frozen LLM, identifying the contextual object (pink) and functional object (azure) to segment.
    The second module (pink) segments the contextual object in input 2D images using an open-vocabulary segmentor, followed by a score-based view selection to discard views where the contextual object is occluded or absent (e.g., the leftmost image).
    In the third module (azure), the functional objects are segmented in the selected views by a VLM paired with a promptable segmentor.
    The fourth module (purple) lifts the 2D masks in 3D via 2D-3D correspondences, performs multi-view agreement, and outputs the 3D segmentation masks of the functional objects.
    }
    \label{fig:diagram}
\end{figure*}

%% file: main/sections/2_related.tex
%%%%%%%%%%%%%%%%%%%%%%%%%%%%%%%%%%%%%%%%%%%%%%%%%%%%%%%%%%%%%%%%%%%%%%%
%%%%%%%%%%%%%%%%%%%%%%%%%%%%%%%%%%%%%%%%%%%%%%%%%%%%%%%%%%%%%%%%%%%%%%%
%%%%%%%%%%%%%%%%%%%%%%%%%%%%%%%%%%%%%%%%%%%%%%%%%%%%%%%%%%%%%%%%%%%%%%%
\section{Related works}\label{sec:related}

%%%%%%%%%%%%%%%%%%%%%%%%%%%%%%%%%%%%%%%%%%%%%%%%%%%%%%%%%%%%%%%%%%%%%%%
\noindent \textbf{Functionality segmentation} in 3D scenes has been recently introduced by SceneFun3D~\cite{delitzas2024scenefun3d} as the task of segmenting functional elements that enable interaction within indoor environments, such as handles, buttons, and knobs.
SceneFun3D provides a dataset containing both 2D images and high-resolution 3D point clouds rich in fine-grained geometric details, which are essential for the precise localization of small functional objects.
Unlike other 3D segmentation datasets~\cite{deng20213daffnet}, SceneFun3D features challenging task descriptions that do not explicitly mention the objects to segment and require world knowledge to be correctly addressed.
Moreover, SceneFun3D focuses on indoor scenes instead of objects as in previous works~\cite{deng20213daffnet}, and thus it is an important benchmark for embodied-AI applications \cite{Engelbracht2024}.
In SceneFun3D~\cite{delitzas2024scenefun3d}, the authors empirically observe that open-vocabulary 3D segmentation methods, such as OpenMask3D~\cite{takmaz2024openmask3d} and LERF~\cite{kerr2023lerf}, perform poorly in functionality segmentation when applied without specific modifications or fine-tuning.
Experiments on OpenMask3D show that fine-tuning the model for this specific task yields slightly better performance, but it remains insufficient for practical use~\cite{delitzas2024scenefun3d}.
This shows that functional segmentation requires ad-hoc architectures to tackle its intrinsic challenges.
In our paper, we reproduce the results of the baselines introduced in SceneFun3D, and add the state-of-the-art open-vocabulary 3D segmentation method OpenIns3D~\cite{huang2023openins3d}.
Our results confirm that these methods cannot reliably segment functional objects, but rather segment common household furniture.
Instead, we introduce the first approach that can segment functional objects from natural language description without relying on task-specific training.

%%%%%%%%%%%%%%%%%%%%%%%%%%%%%%%%%%%%%%%%%%%%%%%%%%%%%%%%%%%%%%%%%%%%%%%
\noindent \textbf{Open-vocabulary 3D segmentation} (OV3DS) methods use natural language descriptions to segment objects in 3D scenes.
These approaches address both semantic segmentation~\cite{peng2023openscene, jiang2024openvocfound} and instance segmentation~\cite{yin2024sai3d, ton2024zeroshotdualpath, takmaz2024openmask3d, nguyen2023open3dis}.
Since OV3DS benchmarks provide both RGBD images and 3D point clouds, OV3DS methods are designed to fully exploit both 2D and 3D data.
They typically rely on a 3D proposal module for predicting 3D masks~\cite{takmaz2024openmask3d, nguyen2023open3dis} directly from the point cloud, while a 2D module extracts 2D masks~\cite{takmaz2024openmask3d, nguyen2023open3dis, ton2024zeroshotdualpath} from multi-view RGB images.
3D proposal modules include ad-hoc trained networks~\cite{choy2019minkowski, schult2023mask3d, hui2022graphcut}, while 2D modules rely on VLMs~\cite{clip, liu2024llava} combined with segmentation or detection components~\cite{kirillov2023segany, zou2024seem, liu2023groundingdino}.
Segmentation is achieved through fusion, either based on 2D-3D mask agreement~\cite{takmaz2024openmask3d, tai2024opensam3d, ton2024zeroshotdualpath} or learnable pooling~\cite{nguyen2023open3dis, peng2023openscene}.
Several OV3DS methods rely on pre-trained models and do not require task-specific training~\cite{takmaz2024openmask3d, yin2024sai3d, ton2024zeroshotdualpath}, while others opt for a distillation strategy~\cite{peng2023openscene, jiang2024openvocfound}. 
Methods based on Language Radiance Fields~\cite{kerr2023lerf} can be used for OV3DS~\cite{qin2023langsplat, engelmann2024opennerf}, but they follow the NeRF~\cite{mildenhall2021nerf} protocol as they require specific training on each scene.
OV3DS methods~\cite{takmaz2024openmask3d, ton2024zeroshotdualpath, nguyen2023open3dis} struggle to segment small object parts, as they rely on modules pre-trained on 3D datasets biased toward large objects, typically common household furniture~\cite{chang2017matterport3d, dai2017scannet}.
Additionally, these methods use concise descriptions that clearly specify object names for easy interpretation.
Instead, \acronym uses Chain-of-Thought reasoning to interpret complex natural language descriptions, and segments the functional elements in 2D to bypass the bias of pre-trained 3D models.

%%%%%%%%%%%%%%%%%%%%%%%%%%%%%%%%%%%%%%%%%%%%%%%%%%%%%%%%%%%%%%%%%%%%%%%
\noindent \textbf{Vision and language models} (VLMs) are multimodal models that can process natural language questions about visual data, reason across both modalities, and generate answers in free-form text.
2D VLMs typically rely on pre-trained unimodal backbones, i.e., a language model and a 2D vision encoder, and differ from each other primarily in how these components are connected.
For instance, LLaVA~\cite{liu2024llava} introduces an MLP-based adapter to align CLIP~\cite{clip} visual features with Vicuna~\cite{vicuna2023} textual features.
Molmo~\cite{deitke2024molmo} adds the ability to answer questions by pointing to the pixels that support the answer, thanks to specialized training on pixel-grounded data.
Due to the limited availability of 3D foundation models and large-scale 3D datasets to train them, only few 3D VLMs currently exist~\cite{hong20233dllm,he2024pointseg,chen2024grounded3dllm}.
However, these methods are trained to ground common objects.
In this paper, we overcome this limitation by using pre-trained 2D VLMs to process images and then aggregate the results onto the point cloud using geometric information.
\acronym can be adapted to other contexts in a zero-shot manner, eliminating the need for additional data collection or model retraining.

%% file: main/sections/3_method.tex
%%%%%%%%%%%%%%%%%%%%%%%%%%%%%%%%%%%%%%%%%%%%%%%%%%%%%%%%%%%%%%%
%%%%%%%%%%%%%%%%%%%%%%%%%%%%%%%%%%%%%%%%%%%%%%%%%%%%%%%%%%%%%%%
%%%%%%%%%%%%%%%%%%%%%%%%%%%%%%%%%%%%%%%%%%%%%%%%%%%%%%%%%%%%%%%
\section{Our approach}

%%%%%%%%%%%%%%%%%%%%%%%%%%%%%%%%%%%%%%%%%%%%%%%%%%%%%%%%%%%%%%%
%%%%%%%%%%%%%%%%%%%%%%%%%%%%%%%%%%%%%%%%%%%%%%%%%%%%%%%%%%%%%%%
\subsection{Overview}\label{sec:overview}

The input to \acronym is a scene point cloud, a set of posed images captured in the same scene, and a task description. 
We first parse the task description with a Large Language Model (LLM), in order to extract the names of the objects to segment (Sec.~\ref{sec:nlpunderstanding}).
Then, we segment the relevant contextual objects in all the views of the scene and select the best views according to the quality and spatial attributes of the mask.
The selected views are processed by a \ass to obtain fine-grained masks of the functional object (Sec.~\ref{sec:reasoning}), which are then assigned to 3D points via multi-view agreement (Sec.~\ref{sec:lifting}).
This results in a heatmap on the point cloud, on which we apply a threshold to obtain the final 3D mask.

%%%%%%%%%%%%%%%%%%%%%%%%%%%%%%%%%%%%%%%%%%%%%%%%%%%%%%%%%%%%%%%
%%%%%%%%%%%%%%%%%%%%%%%%%%%%%%%%%%%%%%%%%%%%%%%%%%%%%%%%%%%%%%%
\subsection{Problem formulation}\label{sec:probform}

Let $\mathcal{C} = \{c_i\}_{i=1}^C$ denote the input point cloud of a scene, where each $c_i \in \mathbb{R}^3$ represents a point coordinate, and $C$ is the total number of points.
Let $\mathcal{V} = \{v_i\}_{i=1}^V$ denote the input views, where $V$ is the total number of RGBD posed images taken from different viewpoints within the scene.
$C$ and $V$ can vary based on the scene.
Let \descr represent the task description that can be used to infer the objects to interact with.
We define two types of objects: the \textit{functional object} is the ultimate object(s) to segment in 3D, and the \textit{contextual object} is the object that contains or is related to the functional object in the space.
For example, if the task description \descr states `\texttt{open the bottom drawer of the cabinet with the TV on top}', the functional object is the `\texttt{knob}' or `\texttt{handle}', and the contextual object is the `\texttt{cabinet}'.
Let \funobj denote the functional object(s), and \parobj denote the contextual object.
\acronym processes \pcd, \allviews and \descr and outputs \funmask, the 3D mask of \funobj on \pcd.

%%%%%%%%%%%%%%%%%%%%%%%%%%%%%%%%%%%%%%%%%%%%%%%%%%%%%%%%%%%%%%%
%%%%%%%%%%%%%%%%%%%%%%%%%%%%%%%%%%%%%%%%%%%%%%%%%%%%%%%%%%%%%%%
\subsection{Task description understanding}\label{sec:nlpunderstanding}

The task description \descr may not explicitly mention \funobj.
E.g.,~in the case of `\texttt{open the bottom drawer of the cabinet with the TV on top}', the agent should only segment the knobs or handles (which are not mentioned in \descr) of the bottom drawer of the cabinet under the TV.

To extract the information about \funobj, we use the LLM to decompose the task description into elements that can be identified in images.
Note that \funobj can also be ambiguous, as there can be other similar objects in the scene.
E.g.,~if the LLM outputs that \funobj is a `\texttt{knob}', this can be confused with that of a nightstand next to the bed; 
whereas, if the LLM outputs that \funobj is a `\texttt{handle}', this can be confused with that of the door.
To address this ambiguity, we query the LLM to provide information about the contextual object \parobj (e.g.,~the cabinet) in addition to \funobj. 
We found this approach to be effective, but we also experienced two issues.
Firstly, the model often fails to correctly identify \funobj due to a lack of a ``stopping criteria" for task understanding and an unclear level of abstraction. 
E.g.,~when queried with `\texttt{open the door}', it may return `\texttt{door}' as \funobj, stopping at contextual-level abstraction. 
Occasionally, it may output `\texttt{hand}', focusing on the finger movements needed to manipulate the handle.
Secondly, when the model identifies \funobj correctly, the LLM may output a contextual object that is typically associated with \funobj, but is unrelated to the specific context of \descr. E.g.,~when queried with `\texttt{open the door}' it correctly returns `\texttt{handle}' as functional object \funobj but returns `\texttt{cabinet}' as contextual object \parobj due to hallucination.

We address the first issue by providing the LLM with a stopping criterion by defining a system description that clarifies its role in assisting a robotic manipulator, which can perform limited actions to interact with the scene.
This list of actions is predefined and based on the annotations of SceneFun3D~\cite{delitzas2024scenefun3d}.
We address the second issue by adopting a Chain-of-Thought strategy~\cite{wei2022chain}, where we query the LLM to first list the actions to complete the task and then identify \funobj along with the hierarchy of objects related to it. 
We select the first item in the hierarchy as the contextual object~\parobj.
Fig.~\ref{fig:cot} reports an example of a conversation with the LLM.

%%%%%%%%%%%%%%%%%%%%%%%%%%%%%%%%%%%%%%%%%%%%%%%%%%%%%%%%%%%%%%%
%%%%%%%%%%%%%%%%%%%%%%%%%%%%%%%%%%%%%%%%%%%%%%%%%%%%%%%%%%%%%%%
\subsection{Multi-view functionality understanding}\label{sec:reasoning}

The main challenge in understanding functionalities from the views in  \allviews lies in handling real-world nuances, such as drawers with different numbers of knobs or perspectives where one knob appears lower than the other along the vertical axis of the image (as shown in Fig.~\ref{fig:mask_ranking}). 
On one hand, multiple functional objects of the same type may be present in a view, but only certain ones should be segmented. 
On the other hand, there are cases where identifying all functional objects in a view is correct; e.g.,~if the query is `\texttt{open a drawer on a cabinet}', then all knobs on that drawer should be segmented.
These nuances require an approach to segmentation that accounts for the spatial location of objects in the environment, rather than relying solely on semantic detection or segmentation.
To achieve this, we leverage \ass's capability to interpret image semantics in order to accurately locate \funobj within the scene \cite{deitke2024molmo}.
Moreover, \allviews may consists of thousands of views, hence processing them all can be computationally costly.
Therefore, we determine which views contain \parobj to reduce the number of views processed in the subsequent steps.
We also prioritize views where \parobj is well visible, thus promoting and fine-grained \funobj localization.

\input{main/figures/llm_conversation/cot}

%%%%%%%%%%%%%%%%%%%%%%%%%%%%%%%%%%%%%%%%%%%%%%%%%%%%%%%%%%%%%%
\noindent\textbf{Contextual object segmentation.}
Given \parobj inferred from \descr by the LLM (Sec.~\ref{sec:nlpunderstanding}), we use open-vocabulary segmentation to locate \parobj in each view. 
In some views, multiple segmentations may appear, as there can be multiple instances of \parobj.
Formally, for a view, we define the $n$-th segmentation mask as ${m^n_\parobj=\{(x^n_\parobj,y^n_\parobj\})}$, that is the set of pixel coordinates belonging to the mask.
For each mask, the segmentation method also outputs a confidence score $S^n_{m_\parobj}$.

%%%%%%%%%%%%%%%%%%%%%%%%%%%%%%%%%%%%%%%%%%%%%%%%%%%%%%%%%%%%%%
\noindent\textbf{Score-based view selection.}
In order to rank the views, we first assign to each mask a visibility score based on both the mask coverage and confidence score.
Given all masks in the view, we use the highest score among the masks as the view score. 
We then select a subset of the views that score the highest to use for the functional object segmentation.

Specifically, to compute the visibility score of a mask $m^n_\parobj$, we use a polar coordinate representation of pixels, assigning a higher score when \parobj's mask is more centered in the image and features a uniform pixel distribution around the center.
Let $d^n_\parobj$ and $\alpha^n_\parobj$ denote the distance and angle of $(x^n_\parobj, y^n_\parobj)$ relative to the image center, which are computed as
%---------------------------------------
\begin{align}
    & d^n_\parobj = \| (x^n_\parobj, y^n_\parobj)\|
    & \alpha^n_\parobj = \arctan(x^n_\parobj, y^n_\parobj).
\end{align}
%---------------------------------------

We analyze the distributions of distance and angle in the mask to assess the visibility of \parobj. 
Let $\texttt{P}^n_{d_\parobj}$ and $\texttt{P}^n_{\alpha_\parobj}$ denote the distance and angle distributions, respectively. 
We want to prioritize a uniform distribution for both the distances and the angles, as it implies that the object is close to the image center and uniformly distributed around it. 
Let $\texttt{U}_d$ and $\texttt{U}_\alpha$ denote the reference uniform distributions for distance and angle, respectively.
We compute the similarity between the measured distributions and the reference one as
%---------------------------------------
\begin{align}
& S^n_{d_\parobj} = 1 - D_{KL}(\texttt{P}^n_{d_\parobj} \mid \mid \texttt{U}_d)
& S^n_{\alpha_\parobj} =  1 - D_{KL}(\texttt{P}^n_{\alpha_\parobj} \mid \mid \texttt{U}_\alpha)
\end{align}
%---------------------------------------
where $D_{KL}$ is the Kullback-Leibler divergence~\cite{1975kldiv}.
$S^n_{d_\parobj}$ is higher when the distance distribution is uniform between 0 and the maximum $d^n_\parobj$ in $m^n_\parobj$, which implies a mask closer to the image center.
$S^n_{\alpha_\parobj}$ is higher when the distribution of angles is closer to a uniform distribution, i.e., when the mask points are uniformly distributed around the image center.
We combine $S^n_{m_\parobj}$, $S^n_{d_\parobj}$, and $S^n_{\alpha_\parobj}$ into a single score as
%---------------------------------------
\begin{equation}\label{eq:maskscoreeq}
S^n_\parobj = \lambda_m S^n_{m_\parobj} + \lambda_d S^n_{d_\parobj} + \lambda_\alpha S^n_{\alpha_\parobj},
\end{equation}
%---------------------------------------
where $S^n_\parobj \in [0,1]$, and $\lambda_{d}$, $\lambda_{\alpha}$ and $\lambda_{m}$ are hyperparameters.
We then assign the maximum score to the view as
%----------------------------------------
\begin{equation}
    S_\parobj=max_n(\{S^n_\parobj\}).
\end{equation}
%----------------------------------------
%
Fig.~\ref{fig:mask_ranking} shows two examples with \parobj (i.e., the cabinet).
In the top row,  $\texttt{P}^n_{d_\parobj}$ and $\texttt{P}^n_{\alpha_\parobj}$ are closer to our reference distribution $\texttt{U}$, thus resulting in higher scores $S^n_{\alpha}$ and $S^n_{d}$. 
Lastly, to process the scene, we create a subset of top-scoring views, defined as $\hat{\mathcal{V}} \subset \mathcal{V}$, where $|\hat{\mathcal{V}}| = \hat{V}$ represents the number of views in which \parobj is both present and well visible. 
In the subsequent step, we localize the functional object within $\hat{\mathcal{V}}$.

%%%%%%%%%%%%%%%%%%%%%%%%%%%%%%%%%%%%%%%%%%%%%%%%%%%%%%%%%%%%%%%
\noindent\textbf{Functional object segmentation.} Accurately localizing functional objects requires a semantic understanding of objects' structural composition.
We can leverage the processing we performed in the previous step as a prior knowledge that a functional object is present and close to the contextual object.
In practice, given the task description \descr and the functional object \funobj, we query our \ass as `\texttt{Point to all the} \funobj \texttt{in order to} \descr'.
The \ass will respond with a set of points $\{ (x_i, y_i) \in \mathbb{Z}^2\}$ on the image plane, which can be empty if the object is not present.
In our previous example, our query would be `\texttt{Point to all the handles in order to open the bottom drawer of the cabinet with the TV on top}'.
Compared to only providing \funobj, adding \descr in the query allows the VLM to disambiguate objects that are consistent with the functional object, but not with the task description as a whole (e.g.,~the handles on the cabinet next to the TV should not be considered).
To obtain the functional object 2D masks, we fed the resulting points to a promptable segmentor~\cite{kirillov2023segany}.

\input{main/figures/ranking/ranking}

%%%%%%%%%%%%%%%%%%%%%%%%%%%%%%%%%%%%%%%%%%%%%%%%%%%%%%%%%%%%%%%
%%%%%%%%%%%%%%%%%%%%%%%%%%%%%%%%%%%%%%%%%%%%%%%%%%%%%%%%%%%%%%%
\subsection{Point cloud functionality segmentation}\label{sec:lifting}

Using camera poses for each view, 3D segmentation masks can be obtained by lifting the predicted 2D functional masks onto the point cloud.
However, this process may introduce false positives if the 2D masks are inaccurate.
To mitigate this, we exploit multi-view agreement: when lifting the 2D masks, we accumulate on each 3D point the count of all the 2D pixels that project onto it, producing a heatmap.
Formally, let $m^k$ represent the functional object mask in the $k$-th view, and let $p^k \in m^k$ be a 2D pixel within the mask.
Let $\Gamma^{k} \colon \mathbb{Z}^2 \to \mathbb{R}^3$ be the 2D-3D mapping from pixels of the $k$-th view to points of the point cloud \pcd, so that $\Gamma^{k}(p^k) = c_i \in$ \pcd.
The score $s_i$ of a point $c_i$ is computed as
%----------------------------------------
\begin{equation}
    s_i = \sum_{k=1}^K \lvert \{p^k \text{ s.t. } \Gamma^{k}(p^k) = c_i \}_{p^k \in m^k} \rvert,
\end{equation}
%----------------------------------------
i.e., the number of pixels of $m^k$ that map to the point $c_i$, across all the $K$ masks.
Finally, we normalize these counts over the point cloud so that $s_i \in [0, 1]$, and define \funmask as the set of points with $s_i>\tau$, where $\tau$ is a hyperparameter.

%% file: main/figures/llm_conversation/cot.tex
\begin{figure}[h]
    \centering
    \begin{overpic}[width=\columnwidth]{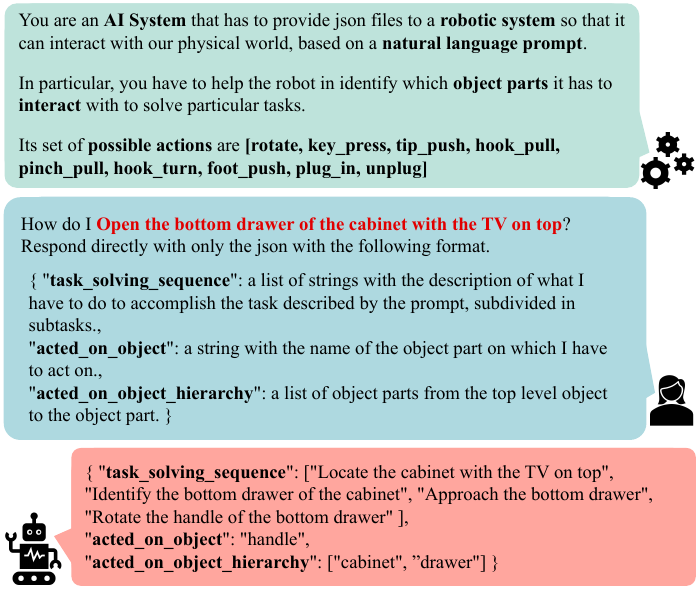}
    \end{overpic}

    \vspace{-2mm}
    \caption{Example of LLM reasoning on a task description (in {\color{red} red}).
    First, we pass a system message to condition the LLM, using the \textbf{possible actions} defined by SceneFun3D.
    Then, we ask to respond with a JSON structure, that includes a \textbf{``task\_solving\_sequence"} field to perform Chain-of-Thought reasoning.
    }
    \vspace{-4mm}
    \label{fig:cot}
\end{figure}

%% file: main/figures/ranking/ranking.tex
\begin{figure}[h]
    \centering
    \begin{overpic}[trim=0 0 0 0, width=1.0\columnwidth]{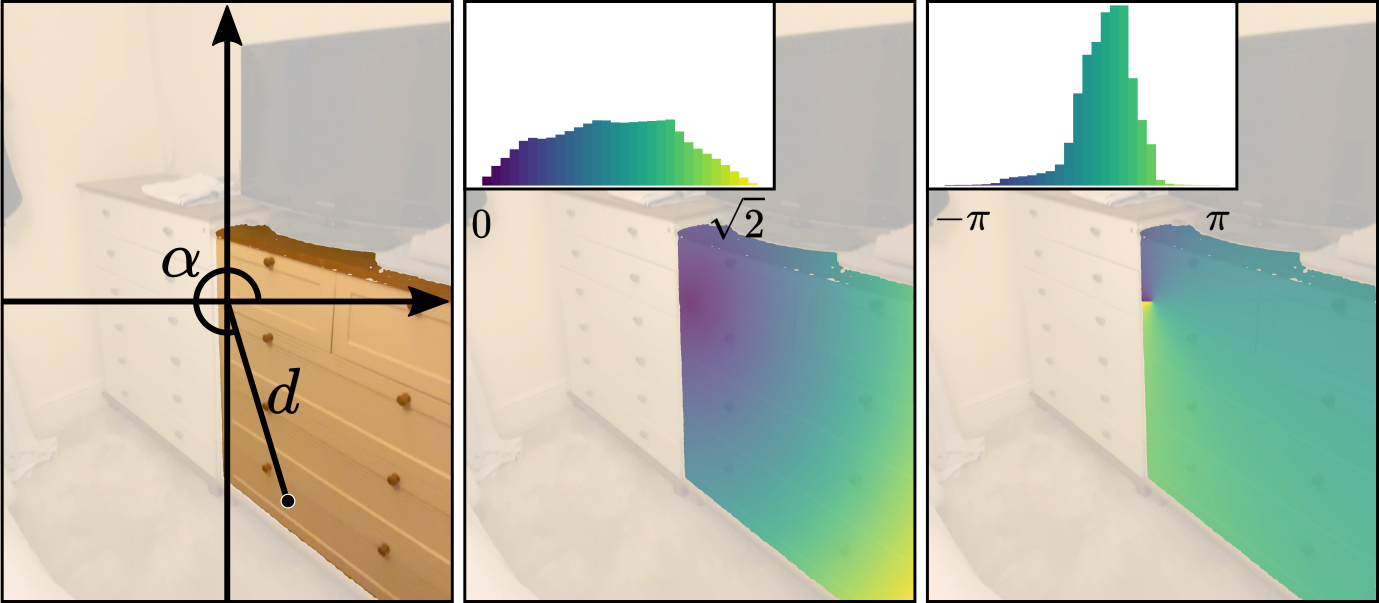}
    \put(6, -4){$S_\texttt{O}=0.63 \quad (S_{m_\texttt{O}}=0.40, S_{d_\texttt{O}}=0.98, S_{\alpha_\texttt{O}}=0.75)$}
    \end{overpic}

    \vspace{6mm}
    \begin{overpic}[trim=0 0 0 0, width=1.0\columnwidth]{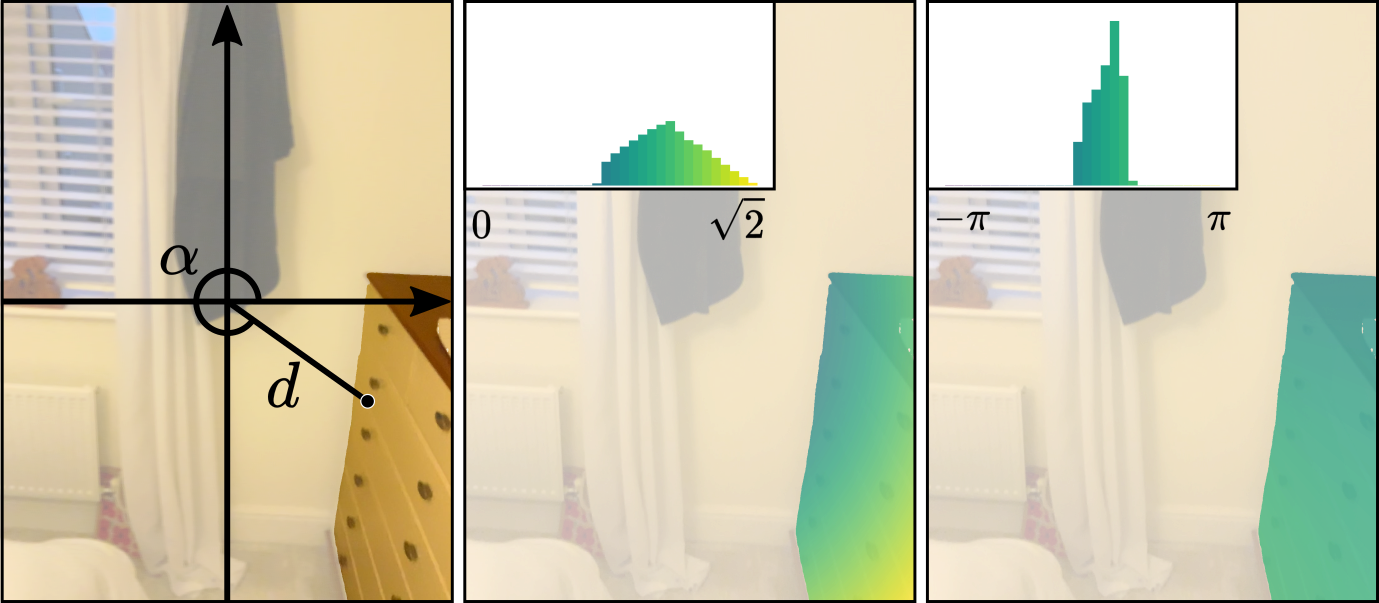}
    \put(6, -4){$S_\texttt{O}=0.45 \quad (S_{m_\texttt{O}}=0.30, S_{d_\texttt{O}}=0.66, S_{\alpha_\texttt{O}}=0.54)$}
    \end{overpic}

    \vspace{2mm}
    \caption{
    Given the example masks in the first column, the second and third column show respectively the distance distribution $\texttt{P}_{d_\parobj}$ and the angle distribution $\texttt{P}_{\alpha_\parobj}$.
    The coordinates are normalized, so that $d_\parobj \in [0,\sqrt{2}]$.
    }
    \vspace{-2mm}
    \label{fig:mask_ranking}
\end{figure}

%% file: main/sections/4_results.tex
%%%%%%%%%%%%%%%%%%%%%%%%%%%%%%%%%%%%%%%%%%%%%%%%%%%%%%%%%%%%%%%%%%%%%
%%%%%%%%%%%%%%%%%%%%%%%%%%%%%%%%%%%%%%%%%%%%%%%%%%%%%%%%%%%%%%%%%%%%%
%%%%%%%%%%%%%%%%%%%%%%%%%%%%%%%%%%%%%%%%%%%%%%%%%%%%%%%%%%%%%%%%%%%%%
\section{Experiments}

%%%%%%%%%%%%%%%%%%%%%%%%%%%%%%%%%%%%%%%%%%%%%%%%%%%%%%%%%%%%%%%%%%%%%
%%%%%%%%%%%%%%%%%%%%%%%%%%%%%%%%%%%%%%%%%%%%%%%%%%%%%%%%%%%%%%%%%%%%%
\subsection{Experimental setup}

%%%%%%%%%%%%%%%%%%%%%%%%%%%%%%%%%%%%%%%%%%%%%%%%%%%%%%%%%%%%%%%%%%%%%
\noindent\textbf{Dataset.}
SceneFun3D~\cite{delitzas2024scenefun3d} is currently the only dataset providing annotations for functionality segmentation in 3D scenes.
It includes high-resolution scans of 3D indoor environments divided into two splits: \valid contains 30 scenes, and \train contains 200 scenes.
Each scene comprises an average of 1800 high-resolution RGBD images, with corresponding intrinsic and extrinsic camera parameters.
SceneFun3D provides an average of 15 task descriptions per scene, along with the associated ground-truth masks.
A task description may correspond to multiple objects within a scene, e.g.,~if the task is `\texttt{Open the bottom drawer}' and that drawer has two handles, the ground-truth mask includes both handles. 
We report results on both \valid and \train.

%%%%%%%%%%%%%%%%%%%%%%%%%%%%%%%%%%%%%%%%%%%%%%%%%%%%%%%%%%%%%%%%%%%%%
\noindent\textbf{Evaluation metrics.}
Following SceneFun3D~\cite{delitzas2024scenefun3d}, we report the Average Precision (AP) at IoU thresholds of 0.25 and 0.5 (denoted as AP$_{25}$ and AP$_{50}$), as well as the mean AP (mAP), calculated over IoUs from 0.5 to 0.95 in 0.05 increments.
Additionally, we report the corresponding Average Recall metrics (AR$_{25}$, AR$_{50}$ and mAR), and the mean IoU (mIoU).

%%%%%%%%%%%%%%%%%%%%%%%%%%%%%%%%%%%%%%%%%%%%%%%%%%%%%%%%%%%%%%%%%%%%%
\noindent\textbf{Implementation details.}
For task description understanding, we use LLama3.1-9B~\cite{dubey2024llama} with 4-bit quantization.
For the contextual object segmentation, we rely on a pipeline built with OWLv2~\cite{minderer2023owl2} and RobustSAM~\cite{chen2024robustsam}.
To segment the functional objects, we first use Molmo~\cite{deitke2024molmo} to point to the objects, then prompt SAM~\cite{kirillov2023segany} with these points to generate the final masks.
We choose Molmo for its point-grounding capability, and because we empirically found that it leads to better results than other VLMs, such as LLaVA~\cite{liu2024llava}.
Although Molmo could also be used for task description understanding, its text-only input capabilities have not been released yet.
\textit{All the above-mentioned models are frozen and used without finetuning.}
For view selection, we compute the score $S_{\parobj}$ with weights $\lambda_m=0.5$ and $\lambda_d=\lambda_{\alpha}=0.25$, and select the top $\hat{V}=50$ views.
The final 3D mask \funmask is obtained by thresholding the heatmap at $\tau=0.7$.

%%%%%%%%%%%%%%%%%%%%%%%%%%%%%%%%%%%%%%%%%%%%%%%%%%%%%%%%%%%%%%%%%%%%%
\noindent\textbf{Baselines}
We use open-vocabulary 3D segmentation methods as baselines for comparison~\cite{huang2023openins3d,nguyen2023open3dis}. 
The baselines include OpenMask3D~\cite{takmaz2024openmask3d} and LERF~\cite{kerr2023lerf} as in SceneFun3D. 
We reproduce their results on \valid and \train.
We incorporate the more recent method OpenIns3D~\cite{huang2023openins3d} as a baseline, modifying its original implementation to better suit the SceneFun3D data. 
In particular, we replace its 2D segmentation module, which operates on rendered views, with the original (real) views provided by SceneFun3D. 
For all baselines, we input the original task description \descr, as it outperformed other alternative descriptions in our testing. 
More details can be found in the Supp.~Mat..

%%%%%%%%%%%%%%%%%%%%%%%%%%%%%%%%%%%%%%%%%%%%%%%%%%%%%%%%%%%%%%%%%%%%%
%%%%%%%%%%%%%%%%%%%%%%%%%%%%%%%%%%%%%%%%%%%%%%%%%%%%%%%%%%%%%%%%%%%%%
\subsection{Quantitative results}

Tabs.~\ref{tab:main_val} and~\ref{tab:main_train} present the results of \acronym and the baselines on \valid and \train of SceneFun3D~\cite{delitzas2024scenefun3d}, respectively.
All baselines score high recall but near-zero precision, indicating a tendency to undersegment functional objects and struggle with capturing fine-grained details.
In both splits, \acronym is the only method to achieve an AP$_{25}$ higher than 0.4, surpassing the nearest competitor, OpenMask3D, by 32.9 and 23.1 points in \valid and \train, respectively.
While OpenIns3D and LERF achieve relatively high AR$_{25}$ scores (51.5 and 36.0 on \valid, 39.9 and 25.1 on \train), their precision remains at zero. 
In fact, \acronym is the only method to achieve an mIoU greater than 0.2 in both splits, outperforming the closest competitor by 15 and 11.4 points on \valid and \train, respectively.
Qualitative results show that all baselines exhibit a strong bias toward contextual objects; e.g,~when tasked with segmenting a cabinet handle, they often segment the entire cabinet. 
This highlights the limitations of such models, which lack the reasoning capabilities needed to accurately interpret the task descriptions of SceneFun3D.
Moreover, we observe that all methods perform worse on \train than on \valid.
This is due to the generally higher complexity of the \train scenes, which exhibit point clouds with up to 13 million points, while \valid is limited to 8 million points.
This complexity leads to a large drop in performance in OpenMask3D, which mostly relies on the 3D encoder~\cite{choy2019minkowski}, and thus is more affected.
Instead, LERF, OpenIns3D, and \acronym perform masking on the 2D views, which makes them more robust to the higher scene complexity.

\input{main/tables/results_unified}

\input{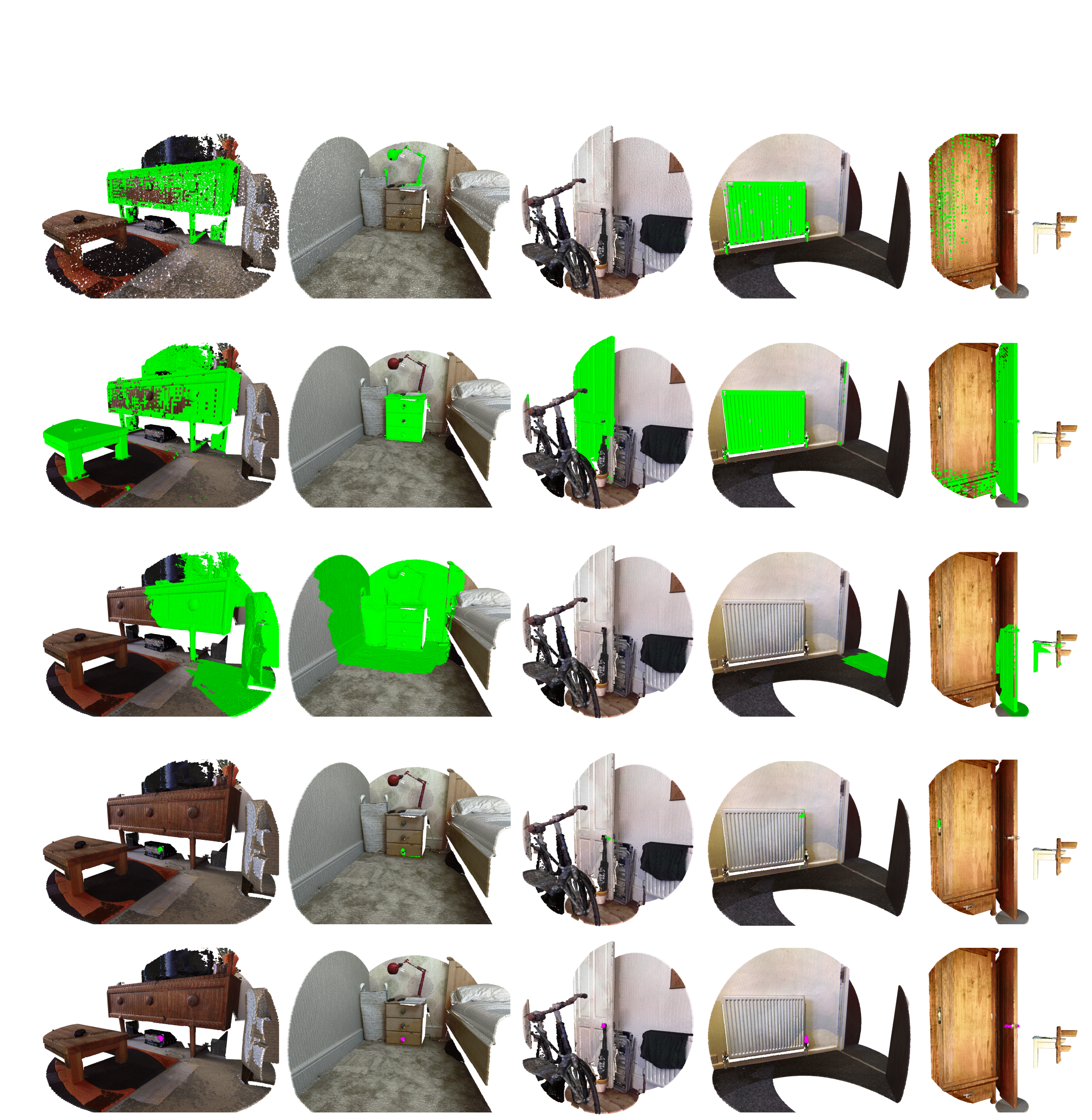}
%%%%%%%%%%%%%%%%%%%%%%%%%%%%%%%%%%%%%%%%%%%%%%%%%%%%%%%%%%%%%%%%%%%%%
%%%%%%%%%%%%%%%%%%%%%%%%%%%%%%%%%%%%%%%%%%%%%%%%%%%%%%%%%%%%%%%%%%%%%
\subsection{Qualitative results}

Fig.~\ref{fig:qualitative} provides examples of predictions by \acronym and the baselines on \valid of SceneFun3D~\cite{delitzas2024scenefun3d}.
As the high-recall and low-precision behavious suggested in Tab.~\ref{tab:main_val}, the baselines tends to focus on the contextual object rather than the functional object.
For example, in the second column OpenMask3D and OpenIns3D segment respectively the red lamp and the nightstand mentioned in the task description, but cannot find the bottom drawer handle.
Similarly, the other methods can locate the TV stand in the first column, but they miss the button of the music system underneath it.
In contrast, \acronym is able to segment the functional objects with a good degree of accuracy, although some spurious segmentation masks are present, such as on the cabinet in the second column.
These errors can be caused by the limited precision of the points provided by the VLM when segmenting the functional object.
When the objects is distant from the camera, the VLM may generate imprecise points that sometimes land on the object beneath (the drawer, in this case), resulting in spurious masks. 
However, the multi-view agreement module filters out most of these errors.
In the fourth column, we observe that \acronym fails to accurately segment the functional object (the radiator dial), and instead segments a portion of the contextual object.
This case is challenging due to the size of the functional object and its similar texture to the surrounding area.
Lastly, the fifth column shows a case where visual ambiguity results in an incorrect mask.
The door mentioned in the task description is positioned near a wardrobe with a very similar door.
As a result, \acronym segments the wardrobe handle instead of the door handle.
In contrast, both LERF and OpenIns3D correctly identify the bedroom door, although their masks do not precisely locate the handle.
See the Supp.~Mat. for additional qualitative results with \acronym.

%%%%%%%%%%%%%%%%%%%%%%%%%%%%%%%%%%%%%%%%%%%%%%%%%%%%%%%%%%%%%%%%%%%%%
%%%%%%%%%%%%%%%%%%%%%%%%%%%%%%%%%%%%%%%%%%%%%%%%%%%%%%%%%%%%%%%%%%%%%
\subsection{Ablation studies}

Unless otherwise specified, all the experiments in this section are conducted on \valid of the SceneFun3D~\cite{delitzas2024scenefun3d} dataset, using \acronym with its standard settings.

%%%%%%%%%%%%%%%%%%%%%%%%%%%%%%%%%%%%%%%%%%%%%%%%%%%%%%%%%%%%%%%%%%%%%
\noindent\textbf{Architectural design.}
Tab.~\ref{tab:ablation_arc} analyzes the impact of each architectural design choice in \acronym on the final performance.
In row 1, we replace the VLM used for functional element segmentation with the open-vocabulary segmentation pipeline employed for localizing contextual objects.
Since OWLv2~\cite{minderer2023owl2} can only process a limited number of text tokens, it cannot handle the full task description.
To address this, we create a shorter description by concatenating \parobj and \funobj (e.g., `\texttt{cabinet handle}').
This variant achieves significantly lower performance compared to our standard method (row 4), with drops of 19.4 AP$_{25}$ and 17.3 AR$_{25}$.
We attribute this performance drop to the limited reasoning capabilities of the open-vocabulary segmentor compared to the VLM.
In row 2, we replace the original task description \descr with a simplified version, asking the VLM to `\texttt{Point to all the} \funobj.'
This removes the contextual information needed to distinguish the correct functional object from other similar instances in the same environment, causing a drop of 14.4 AP$_{25}$ and 12.3 AR$_{25}$ with respect to our standard method (row 4).
Lastly, in row 3, we do not use the view selection module, and instead randomly sample $\hat{V}=100$ input images (instead of our default $\hat{V}=50$).
This also removes the influence of the contextual object segmentation module, as the random sampling does not consider the presence of the contextual object.
Compared to our standard method (row 4), this results in a drop of 12.8 AP$_{25}$ and 4.9 AR$_{25}$.
It is clear that, for accurate segmentation of functional objects, the views quality is far more important than their quantity.

%%%%%%%%%%%%%%%%%%%%%%%%%%%%%%%%%%%%%%%%%%%%%%%%%%%%%%%%%%%%%%%%%%%%%
\noindent\textbf{View selection analysis.}
Tab.~\ref{tab:ablation_scoring} assesses the impact of the hyperparameters in Eq.~\eqref{eq:maskscoreeq}, which influence how input views are ranked by emphasizing different attributes of the contextual object masks.
In row 1, we set $\lambda_{m}=1$ and all other weights to 0, therefore relying solely on the detection confidence score to rank the views.
This results in a drop of 6.3 AP$_{25}$ and 2.4 AR$_{25}$.
Ignoring the position of the contextual object mask often leads to suboptimal view selection, as it may include distant or occluded contextual objects where the functional object is partially visible or even absent.
In rows 2 to 4, we explore the opposite scenario by setting $\lambda_{m}=0$ and varying $\lambda_{d}$ and $\lambda_{\alpha}$.
Row 2 ranks views based exclusively on the distance distribution ($\lambda_{d}=1$), row 3 is based solely on the angle distribution ($\lambda_{\alpha}=1$), and finally row 4 considers both distributions equally ($\lambda_{d}=\lambda_{\alpha}=0.5$).
All configurations underperform compared to our standard method, with row 4 showing the smallest drop, losing 2.5 AP$_{25}$ and 1.5 AR$_{25}$.
Intuitively, the detection confidence score captures the semantic relevance of the predicted mask, while angle and distance scores reflect the quality of the contextual object masks.
By combining these attributes, \acronym leverages the strengths of both in order to obtain optimal performance.

%%%%%%%%%%%%%%%%%%%%%%%%%%%%%%%%%%%%%%%%%%%%%%%%%%%%%%%%%%%%%%%%%%%%%
\noindent\textbf{Hyperparameter sensitivity.}
Fig.~\ref{fig:abl:views} analyzes the sensitivity of our model to the number of views $\hat{V}$ and the final mask threshold $\tau$.
It is clear that an higher $\hat{V}$ allows \acronym to output more accurate functionality masks, as the consensus from many views raises the score of correct masks and removes spurious predictions.
Nonetheless, even when provided with 10 or less views, our model does not completely fail: with the standard $\tau=0.7$, using 10 views results in an mIoU above 10 (light green line), while with 4 or less views (cyan and blue lines) the mIoU is still above 7.
We attribute this result to the view selection procedure, which allows to retain a reasonable performance even with a very sparse, but accurately selected, set of views.
Increasing $\hat{V}$ over 50 results in a marginally better performance (see the Supp.~Mat.), therefore for better trade-off with computational cost we keep $\hat{V}=50$ in our standard setting.

\input{main/tables/ablation_unified}

%% file: main/tables/results_unified.tex
\begin{table}[t]
\centering
\tabcolsep 2pt
\caption{
Results obtained with \acronym and our baselines on \valid of the SceneFun3D dataset.}

\vspace{-2mm}
\label{tab:main_val}
\resizebox{\columnwidth}{!}{%
    \begin{tabular}{rl|ccc|ccc|c}
    \toprule
        & Method & mAP & AP$_{50}$ & AP$_{25}$ & mAR & AR$_{50}$ & AR$_{25}$ & mIoU\\
        \midrule

        \small{\color{gray} 1} & OpenMask3D~\cite{takmaz2024openmask3d} & 0.2 & 0.2 & 0.4 & 20.3 & 24.5 & 27.0 & 0.2 \\ 
        \small{\color{gray} 2} & OpenIns3D~\cite{huang2023openins3d}  & 0.0 & 0.0 & 0.0 & 40.5 & 46.7 & 51.5 & 0.1 \\
        \small{\color{gray} 3} & LERF~\cite{kerr2023lerf} & 0.0 & 0.0 & 0.0 & 34.2 & 35.1 & 36.0 & 0.0 \\ 
        \midrule
        \small{\color{gray} 4} & \acronym        & 7.6 & 16.9 & 33.3   & 27.4 & 38.2 & 46.7 & 15.2 \\

        \bottomrule   
    \end{tabular}
}

\vspace{4mm}
\tabcolsep 2pt
\caption{
Results obtained with \acronym and our baselines on \train of the SceneFun3D dataset.
}

\vspace{-2mm}
\label{tab:main_train}
\resizebox{\columnwidth}{!}{%
    \begin{tabular}{ll|ccc|ccc|c}
\toprule
        & Method & mAP & AP$_{50}$ & AP$_{25}$ & mAR & AR$_{50}$ & AR$_{25}$ & mIoU\\
        \midrule
        
        \small{\color{gray} 1} & OpenMask3D~\cite{takmaz2024openmask3d} & 0.0 & 0.0 & 0.0 & 1.2 & 1.4 & 2.6 & 0.1 \\  
        \small{\color{gray} 2} & OpenIns3D~\cite{huang2023openins3d}    & 0.0 & 0.0 & 0.0 & 32.3 & 37.1 & 39.9 & 0.1 \\ 
 
        \small{\color{gray} 3} & LERF~\cite{kerr2023lerf}                & 0.0 & 0.0 & 0.0 & 23.9 & 24.6 & 25.1 & 0.0 \\ 

        \midrule
        \small{\color{gray} 4} & \acronym & 6.1 & 12.6 & 23.1 & 23.9 & 32.9 & 40.5 & 11.5 \\

        \bottomrule
    \end{tabular}
}
\vspace{-4mm}
\end{table}

%% file: main/figures/qualitative/qualitative.tex
%\newcommand{\OPyopenmask}{88.7}
%\newcommand{\OPyopenins}{70}
%\newcommand{\OPylerf}{51.5}
%\newcommand{\OPyfun}{33}

%correct only for IoU
%\newcommand{\OPxi}{10}
%\newcommand{\OPxii}{30}
%\newcommand{\OPxiii}{49}
%\newcommand{\OPxiiii}{68}
%\newcommand{\OPxiiiii}{85}
\newcommand{\miuold}[3]{\put(#1,#2){
\scriptsize
\text{AP}: $#3$, \text{AR}: $#3$, \text{IoU}: $#3$
}}

\newcommand{\OPyopenmask}{75}
\newcommand{\OPyopenins}{57}
\newcommand{\OPylerf}{38}
\newcommand{\OPyfun}{19.7}

\newcommand{\OPxi}{2}
\newcommand{\OPxii}{24}
\newcommand{\OPxiii}{45}
\newcommand{\OPxiiii}{62}
\newcommand{\OPxiiiii}{82}

\newcommand{\miu}[5]{\put(#1,#2){
%\begin{mdframed}[backgroundcolor=lightgray, roundcorner=10pt]
                \transparent{0.75}
                {\colorbox{white}{%
                \parbox[c][0.8cm][c]{1.6cm}{%
                \transparent{1.0}
                \scriptsize
                \texttt{\textbf{Prc:#3}}\\
                \texttt{\textbf{Rec:#4}}\\
                \texttt{\textbf{IoU:#5}}
                }%
            }%
            }%
}}

\newcommand{\OPprompt}[1]{{\parbox[t][20mm][c]{26mm}{\centering \footnotesize {\texttt{#1}}}}}

\begin{figure*}[t]
    \centering
    \begin{overpic}[width=\textwidth]{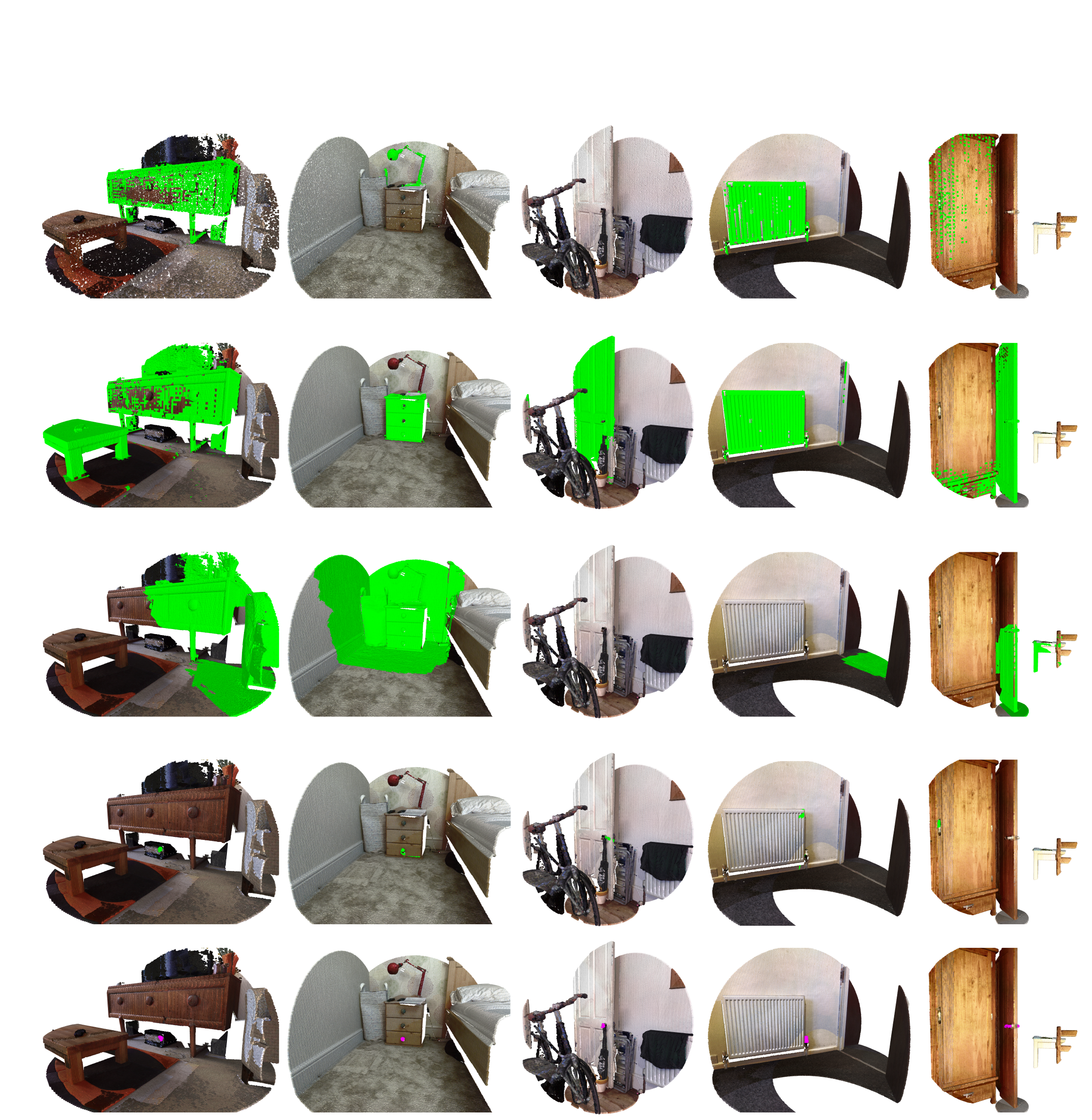}
        % Side names
        \put(0,3){\rotatebox{90}{Ground truth}}
        \put(0,19){\rotatebox{90}{\acronym (ours)}}
        \put(0,39){\rotatebox{90}{LERF \cite{kerr2023lerf}}}
        \put(0,55){\rotatebox{90}{OpenIns3D \cite{huang2023openins3d}}}
        \put(0,73){\rotatebox{90}{OpenMask3D \cite{takmaz2024openmask3d}}}

        % prompts
        \put(7,100){\OPprompt{ Control the volume on the mini music system under the TV stand}}
        \put(28,100){\OPprompt{ Open the bottom drawer of the nightstand with the red table lamp on top}}
        \put(47,100){\OPprompt{ Close the door}}
        \put(65,100){\OPprompt{ Adjust the room's temperature using the radiator dial}}
        \put(82,100){\OPprompt{ Close the\\ bedroom door}}

        % OpenMask
        \miu{\OPxi}{\OPyopenmask}{\ \ 0.00}{\ \ 0.00}{\ \ 0.00}
        \miu{\OPxii}{\OPyopenmask}{\ \ 0.00}{\ \ 0.00}{\ \ 0.00}
        \miu{\OPxiii}{\OPyopenmask}{\ \ 0.00}{\ \ 0.00}{\ \ 0.00}
        \miu{\OPxiiii}{\OPyopenmask}{\ \ 0.62}{\ 63.16}{\ \ 0.62}
        \miu{\OPxiiiii}{\OPyopenmask}{\ \ 0.00}{\ \ 0.00}{\ \ 0.00}
        
        % openins
        \miu{\OPxi}{\OPyopenins}{\ \ 0.00}{\ \ 0.00}{\ \ 0.00}
        \miu{\OPxii}{\OPyopenins}{\ \ 0.08}{100.00}{\ \ 0.08}
        \miu{\OPxiii}{\OPyopenins}{\ \ 0.14}{\ 40.00}{\ \ 0.14}
        \miu{\OPxiiii}{\OPyopenins}{\ \ 0.13}{\ 22.61}{\ \ 0.13}
        \miu{\OPxiiiii}{\OPyopenins}{\ \ 0.03}{\ 96.84}{\ \ 0.03}

        % lerf
        \miu{\OPxi}{\OPylerf}{\ \ 0.00}{\ \ 0.00}{\ \ 0.00}
        \miu{\OPxii}{\OPylerf}{\ \ 0.05}{100.00}{\ \ 0.05}
        \miu{\OPxiii}{\OPylerf}{\ \ 0.00}{\ \ 0.00}{\ \ 0.00}
        \miu{\OPxiiii}{\OPylerf}{\ \ 0.00}{\ \ 0.00}{\ \ 0.00}
        \miu{\OPxiiiii}{\OPylerf}{\ \ 0.06}{100.00}{\ \ 0.06}

        % ours
        \miu{\OPxi}{\OPyfun}{\ 49.54}{\ 67.51}{\ 40.00}
        \miu{\OPxii}{\OPyfun}{\ 32.30}{\ 69.33}{\ 28.26}
        \miu{\OPxiii}{\OPyfun}{\ 18.94}{\ 33.33}{\ 13.73}
        \miu{\OPxiiii}{\OPyfun}{\ \ 1.61}{\ \ 2.01}{\ \ 0.90}
        \miu{\OPxiiiii}{\OPyfun}{\ \ 0.00}{\ \ 0.00}{\ \ 0.00}

    \end{overpic}

    \vspace{-2mm}
    \caption{
    Qualitative examples of \acronym and its baselines on \valid of SceneFun3D~\cite{delitzas2024scenefun3d}.
    Point clouds are cropped around the functional object for better visualization. We report mask-level Precision (Prc), Recall (Rec) and IoU.
    }
    \label{fig:qualitative}
\end{figure*}

%% file: main/tables/ablation_unified.tex
\begin{table}[t]
\centering
\tabcolsep 2pt
\caption{
Ablation on the architectural components of \acronym on \valid of the SceneFun3D dataset.
}

\vspace{-2mm}
\label{tab:ablation_arc}
\resizebox{\columnwidth}{!}{%
    \begin{tabular}{ll|ccc|ccc|c}
\toprule
        & Component & mAP & AP$_{50}$ & AP$_{25}$ & mAR & AR$_{50}$ & AR$_{25}$ & mIoU\\
        \toprule
        \small{\color{gray} 1} & w/o \ass & 1.9 & 4.5 & 13.9 & 17.3 & 23.1 & 29.4 & 6.3 \\ 
        \small{\color{gray} 2} & w/o \descr & 3.8 & 7.9 & 18.9 & 19.6 & 28.3 & 34.4 & 9.5 \\ 
        \small{\color{gray} 3} & w/o \parobj, $\hat{V}=100$ & 5.8 & 11.2 & 20.5 & 24.5 & 33.9 & 41.8 & 11.2 \\ 

        \midrule
        \small{\color{gray} 4} & Standard & 7.6 & 16.9 & 33.3   & 27.4 & 38.2 & 46.7 & 15.2 \\
        \bottomrule
        
    \end{tabular}
}

\vspace{3mm}

\centering
\tabcolsep 3pt
\caption{
Ablation on the view selection hyperparameters of \acronym on \valid of the SceneFun3D dataset.
}

\vspace{-2mm}
\label{tab:ablation_scoring}
\resizebox{\columnwidth}{!}{%
    \begin{tabular}{lccc|ccc|ccc|c}
\toprule
        & $\lambda_m$ & $\lambda_d$ & $\lambda_\alpha$ & mAP & AP$_{50}$ & AP$_{25}$ & mAR & AR$_{50}$ & AR$_{25}$ & mIoU\\
        \toprule
        \small{\color{gray} 1} & 1 & 0 & 0      & 6.0 & 14.2 & 27.0 & 26.1 & 35.3 & 44.3 & 13.2 \\ 
        \small{\color{gray} 2} & 0 & 1 & 0      & 5.8 & 13.7 & 27.2 & 24.8 & 35.9 & 42.5 & 13.3 \\  
        \small{\color{gray} 3} & 0 & 0 & 1      & 6.3 & 15.7 & 30.1 & 24.7 & 36.4 & 44.5 & 14.3 \\
        \small{\color{gray} 4} & 0 & .5 & .5    & 6.4 & 15.5 & 30.8 & 24.5 & 36.6 & 45.2 & 14.3 \\ 
        \midrule
        \small{\color{gray} 5} & .5 & .25 & .25 & 7.6 & 16.9 & 33.3   & 27.4 & 38.2 & 46.7 & 15.2 \\  

        \bottomrule
        
    \end{tabular}
}

\vspace{3mm}

\definecolor{color1}{HTML}{d53e4f}
\definecolor{color2}{HTML}{fc8d59}
\definecolor{color3}{HTML}{fdcf4e}
\definecolor{color4}{HTML}{d1ed45}
\definecolor{color5}{HTML}{99d594}
\definecolor{color6}{HTML}{4575b4}

% ref per colorpalette https://colorbrewer2.org/?
\begin{tikzpicture}
    \begin{axis}[
        xlabel={$\tau$},
        ylabel={mIoU},
        ylabel style={yshift=-0.1cm},
        tick label style={font=\footnotesize},
        legend style={at={(1.05,1.0)}, anchor=north west, font=\footnotesize},
        legend cell align={left},
        xmin=0.1, xmax=0.9,
        ymin=4, ymax=18,
        xtick={0.1,0.3,0.5,0.7,0.9},
        width=0.83\linewidth,
        height=4cm,
        cycle list name=color list,
        grid=both,
        grid style={line width=.1pt, draw=gray!40},
    ]

    % Data for n_samples = 50
    \addplot+[thick, color=color1,mark=o] coordinates {
        (0.1, 5.5)  (0.3, 10.1) (0.5, 13.1) (0.7, 15.3)  (0.9, 13.2)
    };
    % Data for n_samples = 30
    \addplot+[thick, color=color2,mark=square] coordinates {
        (0.1, 5.8)  (0.3, 9.8) (0.5, 12.6) (0.7, 13.9)  (0.9, 12.6)
    };
    % Data for n_samples = 20
    \addplot+[thick, color=color3,mark=triangle] coordinates {
        (0.1, 6.2)  (0.3,9.8 ) (0.5, 12.1) (0.7, 12.9)  (0.9,10.9 )
    };
    % Data for n_samples = 10
    \addplot+[thick, color=color4,mark=diamond] coordinates {
        (0.1, 6.4)  (0.3, 8.7) (0.5, 10.8) (0.7, 10.7 )  (0.9, 10.0)
    };
    % Data for n_samples = 4
    \addplot+[thick, color=color5,mark=star] coordinates {
    (0.1, 6.7)  (0.3, 7.6) (0.5, 9.3) (0.7, 9.4 )  (0.9, 9.6)
    };

    % Data for n_samples = 2
    \addplot+[thick, color=color6,mark=pentagon] coordinates {
        (0.1, 7.1)  (0.3, 7.1 ) (0.5, 8.1) (0.7, 8.1)  (0.9,8.1)
    };

    \legend{$\hat{V}$=50, $\hat{V}$=30,$\hat{V}$=20,$\hat{V}$=10,$\hat{V}$=4,$\hat{V}$=2}
    \end{axis}
\end{tikzpicture}
\vspace{-3mm}
\captionof{figure}{Effect on \valid of the number of views $\hat{V}$ on segmentation performance, reported in mIoU. Each experiment set the threshold $\tau$ to different values to obtain \funmask.}
\label{fig:abl:views}

\vspace{-5mm}
\end{table}

%% file: main/sections/5_conclusion.tex
\section{Conclusions}
We introduced \acronym, the first method for functionality segmentation in 3D scenes.
\acronym is training-free as it relies on a large language model and a vision and language model, both pre-trained with world knowledge.
Key components of our approach are the reasoning module for processing task descriptions and the view selection module for pruning low-quality views.
\acronym is capable of accurate functionality segmentation, but errors can occur when the location of the contextual object alone is insufficient to identify the functional object.
In future work, we will extend \acronym to incorporate additional objects mentioned in the task description beyond the contextual object to act as priors.
This will help resolve ambiguous cases where segmentation currently fails and prove valuable in more complex environments, such as those with multiple rooms.

\noindent\textbf{Acknowledgements.} 
This work was supported by the European Union’s Horizon Europe research
and innovation programme under grant agreement No 101058589 (AI-PRISM).
We also acknowledge ISCRA for awarding this project access to the LEONARDO supercomputer, owned by the EuroHPC Joint Undertaking, hosted by CINECA (Italy).
% https://www.hpc.cineca.it/acknowledgements-in-publications/
%\jaimecomment{According to CVPR25 camera ready instructions, the acknowledgements section can go on the 8th page.}

%% file: supp/mycommands.tex
\newcommand{\fabiocomment}[1]{\todo[color=purple!20, inline, author=Fabio]{#1}}
\newcommand{\davidecomment}[1]{\todo[color=blue!20, inline, author=Davide]{#1}}
\newcommand{\francescocomment}[1]{\todo[color=orange!20, inline, author=Francesco]{#1}}
\newcommand{\jaimecomment}[1]{\todo[color=red!20, inline, author=Jaime]{#1}}

\newcommand{\fabio}[1]{\textbf{\textcolor{purple!75}{#1}}}
\newcommand{\davide}[1]{\textbf{\textcolor{blue!75}{#1}}}
\newcommand{\francesco}[1]{\textbf{\textcolor{orange!75}{#1}}}
\newcommand{\jaime}[1]{\textbf{\textcolor{red!75}{#1}}}

\newcommand{\higherbetter}[0]{{\color{black!50}{$\,\uparrow$}}}
\newcommand{\oracle}[1]{\textcolor{gray}{#1}}

\definecolor{visual}{HTML}{3399FF}
\definecolor{text}{HTML}{97D077}

\newcommand{\impp}[1]{{\textcolor{Green}{+#1}}}
\newcommand{\impn}[1]{{\textcolor{BrickRed}{-#1}}}

\newcommand{\correct}[1]{{\textcolor{Green}{#1}}}
\newcommand{\wrong}[1]{{\textcolor{BrickRed}{#1}}}

\newcommand{\acronym}{Fun3DU\xspace}
\definecolor{myazure}{rgb}{0.8509,0.8980,0.9412}
\newcommand{\cmark}{\ding{51}}%
\newcommand{\xmark}{\ding{55}}%

\newcommand{\assistant}[0]{vision and language model\xspace}
\newcommand{\ass}[0]{VLM\xspace}

\newcommand{\pcd}[0]{$\mathcal{C}$\xspace}
\newcommand{\allviews}[0]{$\mathcal{V}$\xspace}
\newcommand{\funmask}[0]{$\mathcal{M}$\xspace}

\newcommand{\descr}[0]{\texttt{D}\xspace}
\newcommand{\funobj}[0]{\texttt{F}\xspace}
\newcommand{\parobj}[0]{\texttt{O}\xspace}

\newcommand{\valid}[0]{split0\xspace}
\newcommand{\train}[0]{split1\xspace}

%%%%
% Francesco stuff for qualitatives visualization
%%%%

\newcommand{\miuold}[3]{\put(#1,#2){
\scriptsize
\text{AP}: $#3$, \text{AR}: $#3$, \text{IoU}: $#3$
}}

\newcommand{\OPyopenmask}{75}
\newcommand{\OPyopenins}{57}
\newcommand{\OPylerf}{38}
\newcommand{\OPyfun}{19.7}

\newcommand{\OPxi}{2}
\newcommand{\OPxii}{24}
\newcommand{\OPxiii}{45}
\newcommand{\OPxiiii}{62}
\newcommand{\OPxiiiii}{82}

\newcommand{\miu}[5]{\put(#1,#2){
%\begin{mdframed}[backgroundcolor=lightgray, roundcorner=10pt]
                \transparent{0.75}
                {\colorbox{white}{%
                \parbox[c][0.8cm][c]{1.6cm}{%
                \transparent{1.0}
                \scriptsize
                \texttt{\textbf{Prc:#3}}\\
                \texttt{\textbf{Rec:#4}}\\
                \texttt{\textbf{IoU:#5}}
                }%
            }%
            }%
}}

\newcommand{\OPprompt}[1]{{\parbox[t][20mm][c]{26mm}{\centering \footnotesize {\texttt{#1}}}}}

%% file: supp/sections/intro.tex
\noindent In this supplementary material, we provide an analysis of the methods used for functionality segmentation  (Sec.~\ref{sec:suppl:Baselines}), delve deeper into the ablation studies highlighting the capabilities of our proposed approach  (Sec.~\ref{sec:suppl:abl}), and present additional qualitative results (Sec.~\ref{sec:suppl:quali}).
We finally report some hardware and implementation details in Sec.~\ref{sec:details}.

%% file: supp/sections/0_baselines.tex
\section{Baselines}
\label{sec:suppl:Baselines}

In the following section, we report the architecture of the baseline methods, as well as their implementation details (Sec.~\ref{sec:method_detail}).
We then show the results obtained with variations in the prompt and architecture of the baselines (Sec.~\ref{sec:method_results}).

\subsection{Methods details}
\label{sec:method_detail}
As SceneFun3D did not release the code used for OpenMask3D and LERF, we use their original code and follow SceneFun3D's instructions to reproduce the results.

\par\noindent\textbf{OpenMask3D~\cite{takmaz2024openmask3d}} is an open-vocabulary 3D instance segmentation method that operates jointly on 3D and 2D data.
As first step, Mask3D~\cite{schult2023mask3d} is used on the scene to obtain a set of 3D class-agnostic masks.
The 3D masks are projected on the scene views, and the top K views are selected according to the visibility of the projected mask on the frames.
The selected views are segmented via SAM~\cite{kirillov2023segany} and CLIP~\cite{clip} is used to extract multi-scale embedding from each SAM mask.
By leveraging 2D-3D correspondences, the features for each mask are aggregated in 3D, so that a CLIP feature vector characterizes each original 3D mask.
Open-vocabulary segmentation is obtained by embedding the description \descr with CLIP and retrieving the most similar masks, selecting the ones with a similarity higher than a threshold $\tau$.

Following SceneFun3D~\cite{delitzas2024scenefun3d}, we test OpenMask3D without retraining, with the original implementation\footnote{\url{https://github.com/OpenMask3D/openmask3d}} and checkpoint trained on ScanNet200~\cite{dai2017scannet}.
To account for the high dimension of SceneFun3D point clouds with respect to the ones used for training, we subsample each point cloud to a maximum of 2 million points and retain the top 200 masks for the 2D segmentation procedure.

\input{supp/tables/baselines_ext}

\par\noindent\textbf{OpenIns3D~\cite{huang2023openins3d}} is a 3D-only framework for 3D open-vocabulary instance segmentation. 
It features three modules, named `Mask', `Snap', and `Lookup'.  
The `Mask' module generates class-agnostic mask proposals in 3D point clouds, while the `Snap' module generates synthetic images of the scene at multiple scales, in order to cover the locations of the previously extracted 3D masks.
The generated views are processed by an open-vocabulary 2D localization module~\cite{xu2023odise} to extract the objects from the textual description \descr.
Finally, the `Lookup' module searches through the outcomes of each extracted 2D mask, and by using Mask2Pixels correspondences transfers the semantic labels from 2D to 3D.
%As we do for OpenMask3D, 
We test OpenIns3D on SceneFun3D without retraining, using the official implementation\footnote{\url{https://github.com/Pointcept/OpenIns3D}} and the checkpoint trained on ScanNet200~\cite{dai2017scannet}.
However, we found that using the `Snap' module is not beneficial in our case (see Tab.~\ref{tab:openins}), and instead we use the 2D images provided by SceneFun3D. 
Specifically, we sample 120 views per scene at uniform time intervals, and render 24 views when using the `Snap' module, following the OpenIns3D's standard setting.
We found that rendering more than 24 views did not lead to better performance.

OpenIns3D reported results with both Yolo-world~\cite{cheng2024yoloworld} and ODISE~\cite{xu2023odise} as 2D detector.
We use Yolo-world in our standard setting for better performance. 
The open-vocabulary 2D detector is used together with CLIP for ranking and filtering the masks, in order to reduce the false positives.

\par\noindent\textbf{LERF~\cite{kerr2023lerf}} is a method based on neural radiance fields~\cite{mildenhall2021nerf} that produces pixel-level feature maps from arbitrary viewpoints within a scene.
The features are trained to align with CLIP~\cite{clip}, so that a textual description can be used to obtain the most relevant pixels.
By lifting and aggregating the pixels on the 3D point cloud, open-vocabulary 3D semantic segmentation can be performed.
For each high-resolution image sequence in SceneFun3D, we train a LERF model using the official implementation\footnote{\url{https://github.com/kerrj/lerf}} integrated with Nerfstudio\footnote{\url{https://docs.nerf.studio/}}. 
We first sample each high-resolution video sequence at 2Hz, and run Colmap to estimate and refine the camera poses. 
Next, we train LERF models on sequences in which Colmap matches at least 10\% of the views and identifies at least 10 valid views. 
Finally, we use the task description as the prompt to extract relevancy maps for each view.
The final 3D mask is obtained by lifting and accumulating the score of each pixel on each point, and finally applying a threshold on points with a positive score.

\subsection{Additional results}
\label{sec:method_results}
In Tab.~\ref{tab:baselines_ext} we report the extended results on the baselines, obtained with the prompt composed by concatenating the contextual and functional object names (e.g., `\texttt{cabinet handle}').
In principle, using a shorter description is beneficial as all the baselines have been tested with relatively short prompts, i.e., describing single objects.
Instead, compared to using the standard prompt provided by SceneFun3D, this leads to a worse performance in all cases, with the exception of LERF which remains on par (rows 5 vs 6).
In particular, OpenMask3D loses 0.2 AP$_{25}$ and 7.2 AR$_{25}$ (rows 1 vs 2), while OpenIns3D loses 12.8 AR$_{25}$ (rows 3 vs 4).
This suggests that the text encoder used by the baselines~\cite{clip} is to some extent capable of processing more advanced prompts, and therefore removing the context given from the original prompt hinders the performance.

In Tab.~\ref{tab:openins} we report additional results on OpenIns3D~\cite{huang2023openins3d}, showing how our modifications influenced its performance on SceneFun3D.
Row 1 is the version we use as baseline, in which the RGB frames are used in place of the `Snap' module and Yolo-world~\cite{cheng2024yoloworld} replaces ODISE~\cite{xu2023odise} to perform 2D localization.
In row 2, we use the original `Snap' module, which renders a set of views of the scene according to the 3D mask proposals.
Using this module results in a large drop in performance, as the AR$_{25}$ loses 46.6 points.
We observe that scene renders are designed to focus on the common furniture objects retrieved from the mask proposal module, often from a top-down perspective.
This is suboptimal for finding the small functional objects, and therefore OpenIns3D benefits from using the original frames of SceneFun3D, which often show close-up views of the objects.
In rows 3 and 4 we replace Yolo-world with ODISE~\cite{xu2023odise}, respectively without and with the `Snap' module.
Both configurations fall short of the baseline at row 1.
As the authors of OpenIns3D observed, Yolo-world performs better on real images (i.e., without the `Snap' module), while ODISE performs better on rendered images (i.e., with the `Snap' module).

%% file: supp/tables/baselines_ext.tex
\begin{table}[t]
\centering
\tabcolsep 3pt
\caption{
Results obtained by OpenMask3D, LERF, and OpenIns3D with different types of prompts, on \valid of SceneFun3D.
}

\vspace{-3mm}
\label{tab:baselines_ext}
\resizebox{\columnwidth}{!}{%
    \begin{tabular}{rcl|ccc|ccc|c}
    \toprule
        & Method & Prompt & mAP & AP$_{50}$ & AP$_{25}$ & mAR & AR$_{50}$ & AR$_{25}$ & mIoU\\
        \midrule
        \small{\color{gray} 1} & \multirow{2}{*}{\rotatebox{12}{OpenMask3D~\cite{takmaz2024openmask3d}}} & original & 0.2 & 0.2 & 0.4 & 20.3 & 24.5 & 27.0 & 0.2 \\ 
        \small{\color{gray} 2} & & parsed & 0.2 & 0.2 & 0.2 & 13.9 & 17.1 & 19.8 & 0.2 \\
        \midrule
        \small{\color{gray} 3} & \multirow{2}{*}{\rotatebox{12}{OpenIns3D~\cite{huang2023openins3d}}}  & original  & 0.0 & 0.0 & 0.0 & 40.5 & 46.7 & 51.5 & 0.1 \\
        \small{\color{gray} 4} &  & parsed & 0.0 & 0.0 & 0.0 & 30.9 & 35.5 & 38.7 & 0.1 \\ 
        \midrule
        \small{\color{gray} 5} & \multirow{2}{*}{\rotatebox{12}{LERF~\cite{kerr2023lerf}}} & original & 0.0 & 0.0 & 0.0 & 34.2 & 35.1 & 36.0 & 0.0 \\ 
        \small{\color{gray} 6} &  & parsed & 0.0 & 0.0 & 0.0 & 34.2 & 35.1 & 35.9 & 0.0 \\ 
        \midrule
        \small{\color{gray} 7} & \acronym (ours) & original & 7.6 & 16.9 & 33.3   & 27.4 & 38.2 & 46.7 & 15.2 \\      
        \bottomrule   
    \end{tabular}
}
\vspace{5mm}

\caption{
Results obtained by OpenIns3D~\cite{huang2023openins3d}, with and without the original `Snap' rendering module used for 2D segmentation, and with different segmentors, on \valid of the SceneFun3D dataset.
}
\label{tab:openins}
\vspace{-3mm}
\resizebox{\columnwidth}{!}{%
    \begin{tabular}{rcl|ccc|ccc|c}
    \toprule
        & Snap & Segmentor & mAP & AP$_{50}$ & AP$_{25}$ & mAR & AR$_{50}$ & AR$_{25}$ & mIoU\\
        \midrule
        \small{\color{gray} 1} & & Yolo-world~\cite{cheng2024yoloworld} & 0.0 & 0.0 & 0.0 & 40.5 & 46.7 & 51.5 & 0.1 \\
        \small{\color{gray} 2} & \cmark & Yolo-world~\cite{cheng2024yoloworld} & 0.0 & 0.0 & 0.0 & 4.1 & 4.3 & 4.9 & 0.0 \\ 
        \small{\color{gray} 3} & & ODISE~\cite{xu2023odise} & 0.0 & 0.0 & 0.0 & 24.8 & 29.8 & 31.9 & 0.0 \\ 
        \small{\color{gray} 4} & \cmark & ODISE~\cite{xu2023odise} & 0.0 & 0.0 & 0.0 & 4.1 & 5.6 & 6.7 & 0.0 \\
        \bottomrule   
    \end{tabular}
}
\end{table}

%% file: supp/sections/1_ablations.tex
\section{Extended ablation studies}
\label{sec:suppl:abl}
In Sec.~\ref{sec:ablation} we extend the ablation study on the number of views $\hat{V}$, by showing additional results and reporting how this hyperparameter influences the processing time.
\subsection{Hyperparameter sensitivity analysis}
\label{sec:ablation}

\input{supp/tables/ablation_views}

We report in Tab.~\ref{tab:abl_views} the mIoU values obtained by varying the threshold $\tau$ and the number of sampled views $\hat{V}$.
Additionally to the values reported in the main paper, we show the change in performance with $\hat{V}=100$ and $\hat{V}=200$.
We observe that, compared to our standard setting (row 4, $\hat{V}=50$), sampling $\hat{V}=100$ views causes a small increment of 0.6 points of mIoU, but almost doubles the processing time per task description (118.4 vs 204.6 seconds).
Instead, setting $\hat{V}=200$ is not beneficial, as it lowers the mIoU of 1.3 points with respect to our baseline with the default $\tau=0.7$.
When so many views are sampled, spurious masks are more easily retrieved, thereby lowering the final performance.
As a trade-off between performance and execution time, we set $\hat{V}=50$ in our standard setting.

%% file: supp/tables/ablation_views.tex
 \begin{table}[b]
\centering
\tabcolsep 3pt
\caption{
Results in mIoU on \valid of SceneFun3D, obtained by sampling different numbers of views $\hat{V}$ and using different thresholds $\tau$.
The last row reports the average time per task description \descr for retrieving the top views, performing functional object segmentation, and multi-view agreement to obtain the final point clouds.
The timings for task description understanding and contextual object segmentation are excluded, as they are constant.
}

\vspace{-3mm}
\label{tab:abl_views}
\resizebox{\columnwidth}{!}{%
    \begin{tabular}{rl|cccccccc}
    \toprule
             \multicolumn{2}{c|}{\multirow{2}{*}{$\tau$}} &\multicolumn{8}{c}{Number of views $\hat{V}$ used for View Selection}\\
             \multicolumn{2}{c|}{} & 2 & 4 & 10 & 20 & 30 & 50 & 100 & 200 \\
             \midrule
             \small{\color{gray} 1} & 0.1 & 7.1 & 6.7 & 6.4 & 6.2 & 5.8 & 5.5 & 4.8 & 3.4 \\
             \small{\color{gray} 2} & 0.3 & 7.1 & 7.6 & 8.7 & 9.8 & 9.8 & 10.1 & 10.1 & 7.6 \\
             \small{\color{gray} 3} & 0.5 & 8.1 & 9.3 & 10.8 & 12.1 & 12.6 & 13.1 & 14.0 & 11.4 \\
             \small{\color{gray} 4} & 0.7 & 8.1 & 9.4 & 10.7 & 12.9 & 13.9 & 15.3 & 15.9 & 14.0 \\
             \small{\color{gray} 5} & 0.9 & 8.1 & 9.6 & 10.0 & 10.9 & 12.6 & 13.2 & 13.3 & 11.1 \\
             \midrule
             \multicolumn{2}{c|}{Time (s)} & 28.4 & 34.7 & 42.3 & 60.7 & 81.9 & 118.4 & 204.6 & 237.6 \\
        \bottomrule   
    \end{tabular}
}
\end{table}

%% file: supp/sections/2_qualitatives.tex
\section{Additional qualitative results}
\label{sec:suppl:quali}
In Sec.~\ref{sec:task_description} we report examples of outputs of the task description understanding module, showing how the contextual and functional object names are retrieved.
Sec.~\ref{sec:view_selection} reports examples of view selection, by showing the retrieved contextual masks along with their scores.
In Sec.~\ref{sec:molmo_seg}, we show examples of points extracted with Molmo~\cite{deitke2024molmo}, along with the final segmentation mask provided by SAM~\cite{kirillov2023segany}.
Finally, in Sec.~\ref{sec:segmentation} we extend the qualitative results on functionality segmentation provided in the main paper.

\subsection{Task description understanding}
\label{sec:task_description}

\begin{figure*}
    \input{supp/figures/llm_examples/llm_tikz_pic}
\end{figure*}

We report in Fig.~\ref{fig:supl:LLM}, some examples of task description understanding, where we use the LLM to extract the functional object \funobj and contextual object \parobj from a given text description \descr. The description is shown in the blue box, and the LLM response is shown in the red box. For the sake of brevity, in the figure we omit the system message ``\emph{You are an AI System that has to provide JSON files to a robotic system so that it can interact with our physical world, based on a natural language prompt. In particular, you have to help the robot in identify which object parts it has to interact with to solve particular tasks. Its set of possible actions are [rotate, key\_press, tip\_push, hook\_pull, pinch\_pull, hook\_turn, foot\_push, plug\_in, unplug]}", as it is the same for all conversations. For the same reason, we omit also the part of the user message where we ask the LLM to respond with a structured json: ``\emph{How do I \{\texttt{Task description \descr}\}? Respond directly with only the json with the following format. \{ "task\_solving\_sequence": a list of strings with the description of what I have to do to accomplish the task described by the prompt, subdivided in subtasks.,
"acted\_on\_object": a string with the name of the object part on which I have to act on.,
"acted\_on\_object\_hierarchy": a list of object parts from the top level object to the object part. \}}".

We observe that in most cases, the LLM successfully provides an object hierarchy that is useful for the view selection process. 
Typically, the first object in the hierarchy, which we use as ``contextual object" (\parobj), serves this purpose effectively.
However, there are some failure cases, specifically (c) and (d). 
In case (c), the object hierarchy is reversed; an uncommon but occasionally observed behavior. 
Despite this reversal, the `telephone' is still correctly identified as the contextual object. In case (d), however, the contextual object is identified as `ceiling light', which is unsuitable for view selection. 
This happens because the functional object `light switch' would more likely be mounted on the wall.

An interesting observation arises in cases (e) and (f), where a small change in the description, from ``left" to ``right", leads to significantly different responses. These examples highlight how the LLM reasons at different levels of abstraction. In case (e), it assumes that the ``left tap" can be acted upon, identifying it as the functional object. Instead, in case (f), it interprets the tap as part of the object hierarchy, identifying the functional object as the ``switch" on the tap.

These examples underline the LLM's overall effectiveness in interpreting task descriptions and generating useful object hierarchies, while also highlighting certain limitations in handling ambiguous or nuanced descriptions. 
Understanding these strengths and weaknesses is crucial for refining its reasoning capabilities and ensuring more consistent performance in diverse scenarios.

\subsection{Score-based view selection}
\label{sec:view_selection}
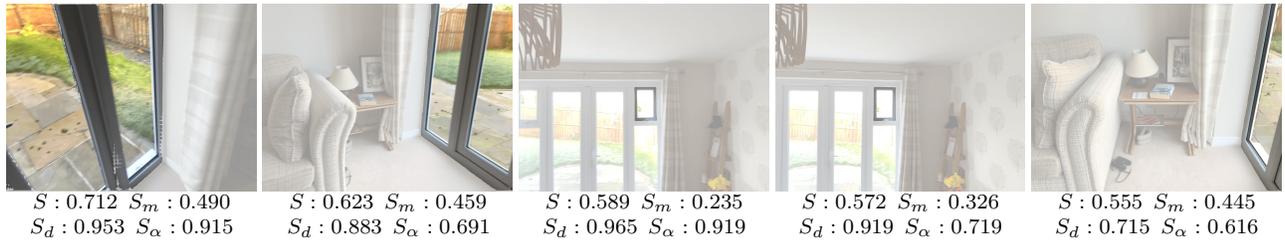
\begin{figure*}[]
    \centering
    \input{supp/figures/qualit_views/qualit_views}
    %\smallskip
    \caption{
    Examples of selected contextual object masks with their scores, from highest rank (left) to lowest rank (right).
    We highlighted the segmented regions by alpha blending with a white background, to enhance the result visibility.
    On top of each view set we report the contextual object.
    }
    \label{fig:qualit_views}
\end{figure*}

We report in Fig.~\ref{fig:qualit_views} some examples of contextual object masks retrieved by our view selection mechanism, from the highest score (leftmost) to the lowest score (rightmost).
For each mask, we report the total score used for ranking $S$, the detection confidence score $S_m$, the distance distribution score $S_d$, and the angle distribution score $S_\alpha$.
In the first row, we can observe an ideal case, in which the first three views show the contextual object (the kitchen range hood) well visible in the images, while the fourth and fifth images only show a portion of the object.
Similarly, in the second row the first two images show a close-up view of the nightstand, while the remaining images show a progressively far and occluded mask.
Note that the last two images are the ones with the highest detection confidence $S_m$ among the five samples. 
This shows that only considering the detection confidence could lead to selecting views that are not optimal for functionality segmentation.

In the third row, the TV stand is well-visible and centered in the first three views.
In the last two views instead, partial occlusions lead to a lower score, mostly due to the low angle scores $S_\alpha$.
Finally, the last row shows a more complex case, in which multiple instances of the contextual object (window) are present.
In this case, the correct masks (the third and fourth) have high $S_d$ and $S_\alpha$ scores, but due to a lower $S_m$ score, they are not ranked in the first positions.
This shows that weighting all the scores can provide the correct masks that would be discarded when only the confidence score $S_m$ is considered. 

\subsection{Functional object segmentation}
\label{sec:molmo_seg}

Fig.~\ref{fig:qualit_molmo} shows some examples of functional object segmentation on the scene views.
For each column, we report on the left the point(s) provided by Molmo~\cite{deitke2024molmo}, and on the right the mask(s) obtained by SAM~\cite{kirillov2023segany}, using the point(s) as input.

In the first row, the \ass can correctly point to the functional objects (the bottom drawer handles), but due to the limited image resolution, the whole drawer is segmented in the first case.
Due to our view selection strategy, in most cases, the contextual object (the nightstand) is clearly visible in the view, as in the second column, and therefore the final 3D mask accurately segments the functional object.

The second row shows a very ambiguous case, in which three radiator parts look similar to the functional object (the dial).
In the first case, all objects are pointed and segmented, while in the second case, only the correct one is segmented.
These spurious errors are effectively removed by our multi-view agreement strategy, by accumulating the 2D masks and applying a threshold.

In the third row, we can observe a particularly difficult case, in which the functional object (the remote) is very occluded in the first view.
Nonetheless, the \ass is capable of pointing to it, resulting in a correct 2D mask.
This shows that the \ass is effective not only in finding small objects such as knobs and buttons, but also in recognizing heavily occluded objects, such as the remote in the second case.

In the final row, we observe how the \ass is capable of recognizing the humidifier, which is an unusual object that can have very different appearances based on its design.
In both views, the \ass can point to the central button of the humidifier and SAM provides an accurate mask.

\subsection{Point cloud segmentation}
\label{sec:segmentation}

We report additional qualitative results on \valid of SceneFun3D in Fig.~\ref{fig:qualit_3d_split0}.
In the first column, \acronym is the only method to obtain an accurate segmentation mask of the light switch.
Note that the precision in this case is relatively low, as the ground-truth mask only considers the light switch itself, while our method also segments the panel around it.
In all the other cases, \acronym obtains precision higher than 70 and recall higher than 65, while the baselines either segment the whole contextual object (e.g., OpenIns3D in all cases), or fail in finding relevant points (e.g., OpenMask3D in the third and fourth column, and LERF in the second and fifth column).
These examples show particularly small functional objects, which are extremely difficult to handle without ad-hoc techniques as in \acronym.

We report qualitative results on \train of SceneFun3D in Fig.~\ref{fig:qualit_3d_split1}.
This partition is more difficult than \valid, as it features more complex scenes with more points, on average.
In the first column, \acronym can accurately segment the handle on the oven, while other methods fail at finding a relevant mask (OpenMask3D, LERF) or segment a large portion of the kitchen (OpenIns3D).
Similarly, in the second column, our method is the only one to locate the small valve at the bottom of the radiator, while OpenIns3D is only capable of segmenting the whole object.
The third column shows a case in which \acronym fails: instead of segmenting the lock used to open the blue case, it segments the handle just underneath it.
This could be due to an error in the task description understanding, in which `\texttt{handle}' has been provided by the LLM instead of `\texttt{lock}'.
Instead, the other methods either segment only the audio system (OpenIns3D) or cannot find any relevant mask.
In the fourth column our method correctly identifies the telephone on the TV, but segments the whole phone instead of the dial pad, resulting in high recall and relatively low precision.
Finally, the fifth column is a particularly difficult case, in which the task description requires adjusting the seat height of the exercise bike, on which two knobs are located.
\acronym segments both knobs, still resulting in a high mIoU of 44.12, while the other methods fail at providing accurate masks.
In Fig.~\ref{fig:qualit_big}, we report two examples of segmentation masks of \acronym on the complete point cloud of the scene, to show the complexity of the context surrounding the functional objects.

\begin{figure*}[]
    \centering
    \input{supp/figures/qualit_molmo/qualit_molmo}
    %\smallskip
    \caption{
    Examples points (in green) extracted with the \ass~\cite{deitke2024molmo}, along with the masks produced by SAM~\cite{kirillov2023segany}.
    We highlighted the segmented regions by alpha blending with a white background, to enhance the result visibility.
    }
    \label{fig:qualit_molmo}
\end{figure*}

\input{supp/figures/qualit_3d/qualitative_split0}

\input{supp/figures/qualit_3d/qualitative_split1}

\begin{figure*}[]
    \centering
    \input{supp/figures/qualit_3d/qualitative_big}
    %\smallskip
    \caption{
    Qualitative results of \acronym on the whole point cloud (left), with detail on the portion within the red circle, that shows the segmented object (right).
    For better visualization, we removed the room ceilings from the point clouds.
    }
    \label{fig:qualit_big}
\end{figure*}

%% file: supp/figures/llm_examples/llm_tikz_pic.tex
\definecolor{myBlue}{HTML}{aed9e0}
\definecolor{myRed}{HTML}{ffa69e}

\newcommand{\lmmConv}[2]{
\vspace{0pt}
\begin{tikzpicture}[node distance=0.4cm]
% Top bubble (blue)
    \node[draw, fill=myBlue, rounded corners,  minimum height=1.6cm,  text width=0.9\linewidth, align=left] (top) {
        %\small
        #1
    };
% Bottom right bubble (red)
    \node[draw, fill=myRed, rounded corners,  minimum height=4.4cm, text width=0.9\linewidth, align=left, below=of top] (bottom)  {
        %\small
        #2
    };
\end{tikzpicture}
}

\newcommand{\lmmConvii}[2]{
\vspace{0pt}
\begin{tikzpicture}[node distance=0.4cm]
% Top bubble (blue)
    \node[draw, fill=myBlue, rounded corners,  minimum height=1.3cm,  text width=0.9\linewidth, align=left] (top) {
        %\small
        #1
    };
% Bottom right bubble (red)
    \node[draw, fill=myRed, rounded corners,  minimum height=3.3cm, text width=0.9\linewidth, align=left, below=of top] (bottom)  {
        %\small
        #2
    };
\end{tikzpicture}
}

\hfill
\begin{minipage}[t]{0.33\linewidth}
\lmmConv{Open the bottom drawer of the nightstand with the red table lamp on top}{
      \textbf{task\_solving\_sequence}:\newline
      [``Locate the nightstand with the red table lamp on top", 
      ``Identify the bottom drawer of the nightstand", 
      ``Approach the bottom drawer and position yourself to interact with it" ]\\
      \textbf{acted\_on\_object}: ``drawer handle",\\
      \textbf{acted\_on\_object\_hierarchy}:\newline [ ``nightstand", ``drawer" ]
}
\end{minipage}
\hfill
\begin{minipage}[t]{0.33\linewidth}
\lmmConv{Turn on the TV using one of the remotes on the headboard}{
      \textbf{task\_solving\_sequence}:\newline [ 
            ``Locate the remote control on the headboard",
            ``Identify the power button on the remote control",
            ``Press the power button with your finger"
      ]\\
      \textbf{acted\_on\_object}: ``power button",\\
      \textbf{acted\_on\_object\_hierarchy}:\newline [ 
            ``headboard",
            ``remote control",
            ``power button"
      ]
}
\end{minipage}
\hfill
\begin{minipage}[t]{0.33\linewidth}
\lmmConv{Dial a number on the telephone on the nightstand}{
      \textbf{task\_solving\_sequence}:\newline [ 
             ``Locate the telephone on the nightstand",
            ``Identify the dial pad on the telephone",
            ``Place your finger on the correct digit to press"
      ]\\
      \textbf{acted\_on\_object}: ``dial pad",\\
      \textbf{acted\_on\_object\_hierarchy}:\newline [
                  ``telephone",
            ``nightstand"
      ]
}
\end{minipage}\\
\begin{minipage}[t]{0.33\linewidth}
\centering
(a)
\end{minipage}
\hfill
\begin{minipage}[t]{0.33\linewidth}
\centering
(b)
\end{minipage}
\hfill
\begin{minipage}[t]{0.33\linewidth}
\centering
(c)
\end{minipage}\\\vspace{0.2cm}\\
\begin{minipage}[t]{0.33\linewidth}
\lmmConvii{Turn on the ceiling light}{
      \textbf{task\_solving\_sequence}:\newline [ ``Locate the light switch" ]\\
      \textbf{acted\_on\_object}: ``light switch",\\
      \textbf{acted\_on\_object\_hierarchy}:\newline [ 
            ``ceiling light",
            ``light fixture",
            ``electrical panel"
            ]
}
\end{minipage}
\hfill
\begin{minipage}[t]{0.33\linewidth}
\lmmConvii{Turn on the left tap so that hot water flows into the sink}{
      \textbf{task\_solving\_sequence}:\newline [ 
      ``Locate the sink", 
      ``Identify the left tap", 
      ``Move towards the left tap" ]\\
      \textbf{acted\_on\_object}: ``left tap",\\
      \textbf{acted\_on\_object\_hierarchy}:\newline [ ``sink", ``tap" ]
}
\end{minipage}
\hfill
\begin{minipage}[t]{0.33\linewidth}
\lmmConvii{Turn on the right tap so that cold water flows into the sink}{
      \textbf{task\_solving\_sequence}:\newline [ 
      ``Locate the sink", 
      ``Identify the right tap",
            ``Find the switch or handle for the right tap" ]\\
      \textbf{acted\_on\_object}: ``switch",\\
      \textbf{acted\_on\_object\_hierarchy}:\newline [ ``sink",
            ``tap",
            ``switch" ]
}
\end{minipage}\\
\begin{minipage}[t]{0.33\linewidth}
\centering
(d)
\end{minipage}
\hfill
\begin{minipage}[t]{0.33\linewidth}
\centering
(e)
\end{minipage}
\hfill
\begin{minipage}[t]{0.33\linewidth}
\centering
(f)
\end{minipage}

\caption{Examples of LLM conversations. The blue box shows the task description \descr, and the red box is the LLM response. We omit the system message and the JSON structure in the user prompt, which can be observed in Fig. 3 of the main paper.}\label{fig:supl:LLM}

%% file: supp/figures/qualit_views/qualit_views.tex
%%%%%%%%%%%%%%%%%%%%%%%%%%%%%%%%%5
%467115 7b8dd2e9-dd7e-4356-bfb3-ac2dbfb0347b Select a function on the kitchen range hood
%%%%%%%%%%%%%%%%%%%%%%%%%%%%%%%%%

\begin{minipage}[t]{\textwidth}
    \vspace{0pt}
    \centering{\texttt{Kitchen range hood}}
\end{minipage}\\[0pt]
%\hspace*{0.00mm}
%\vspace*{1mm}
\begin{minipage}[t]{0.19\textwidth}
\vspace{0pt}
    \begin{overpic}[width=1\textwidth]{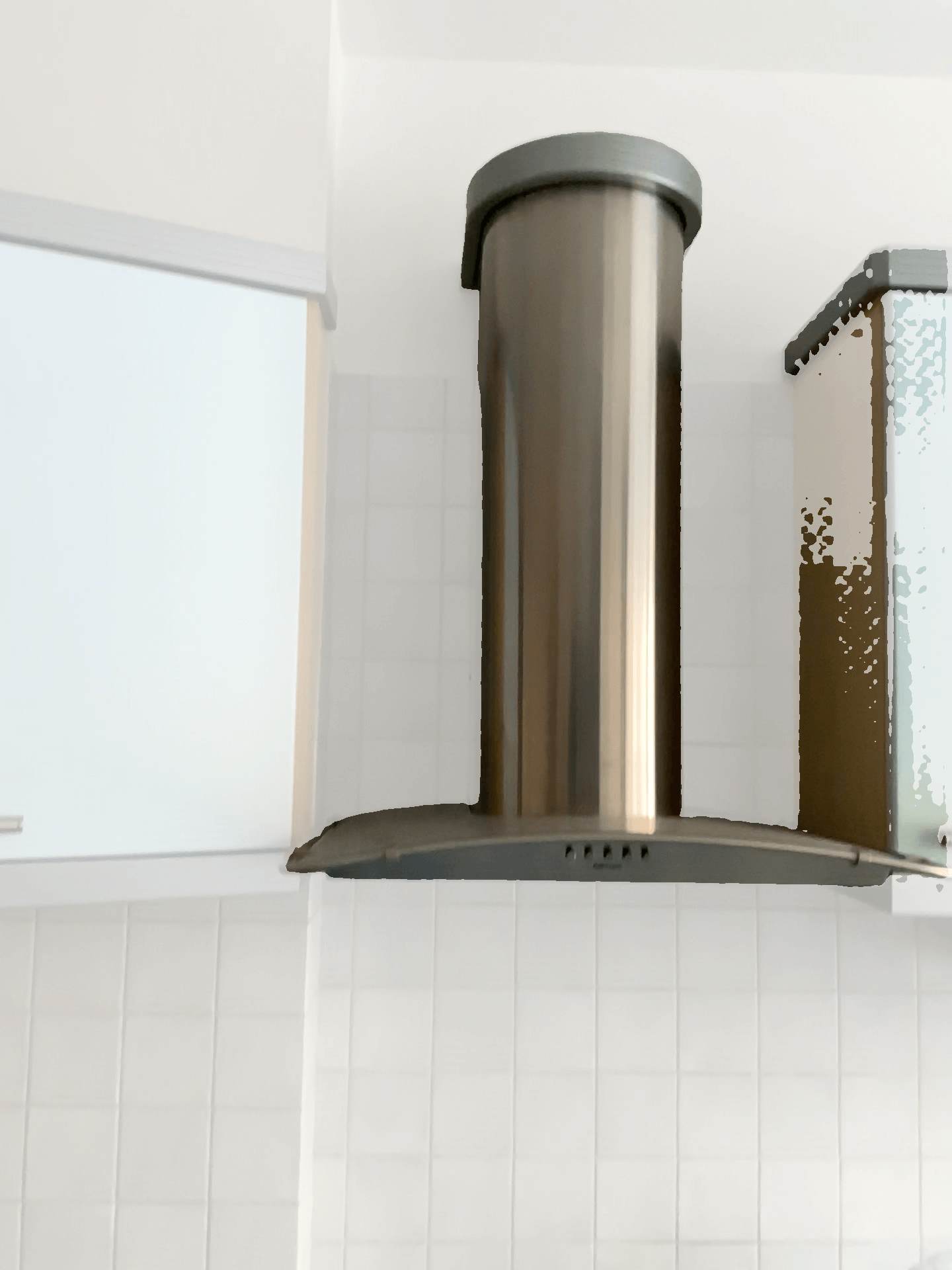}
    % \put(6,43){\small{\textcolor{magenta}{$\mathbf{Q}$}}}
    % \put(54,43){\small{\textcolor{magenta}{$\mathbf{A}$}}}
    \end{overpic}
\end{minipage}
\begin{minipage}[t]{0.19\textwidth}
\vspace{0pt}
    \begin{overpic}[width=1\textwidth]{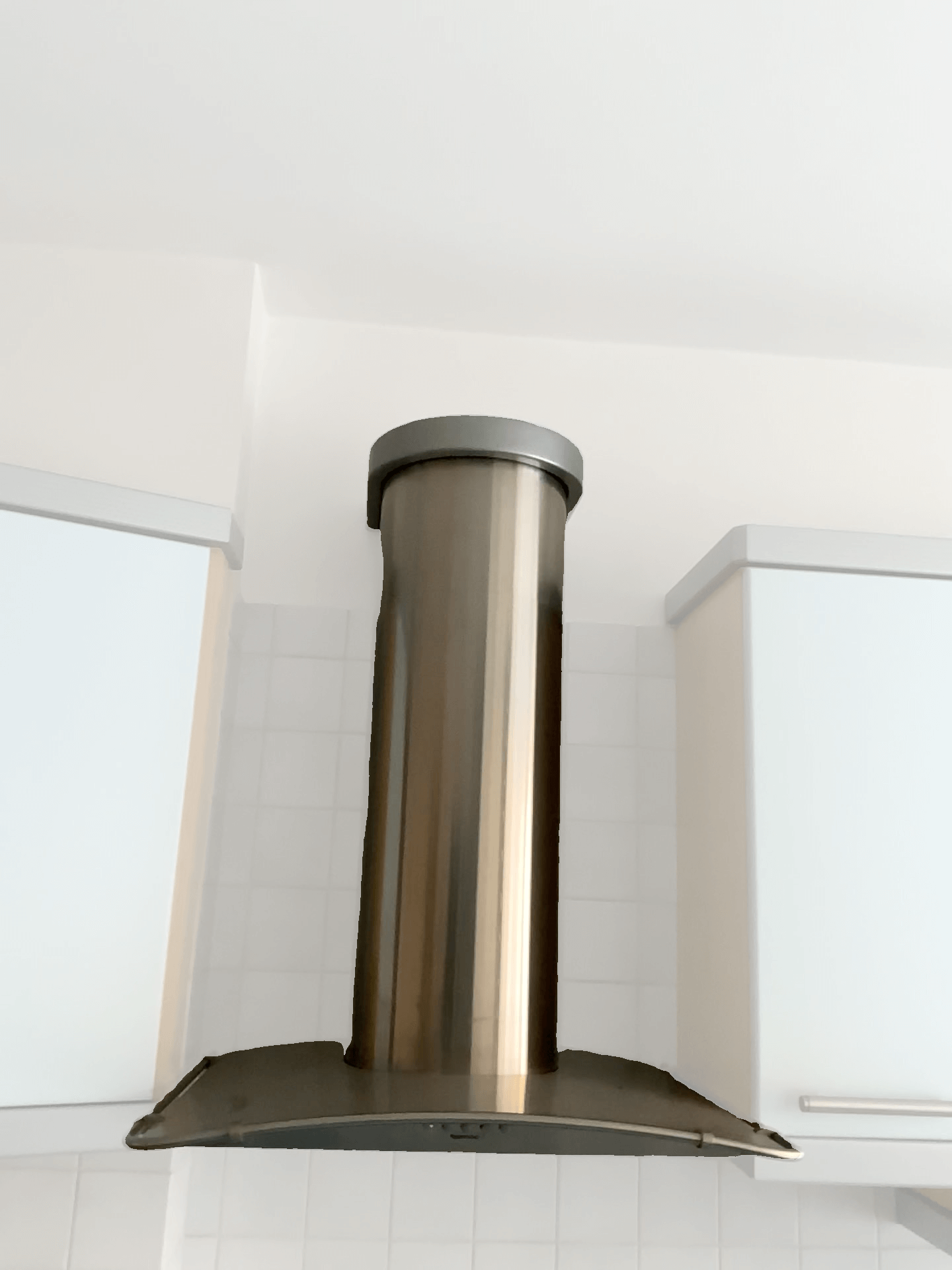}
    % \put(6,43){\small{\textcolor{magenta}{$\mathbf{Q}$}}}
    % \put(54,43){\small{\textcolor{magenta}{$\mathbf{A}$}}}
    \end{overpic}
\end{minipage}
\begin{minipage}[t]{0.19\textwidth}
\vspace{0pt}
    \begin{overpic}[width=1\textwidth]{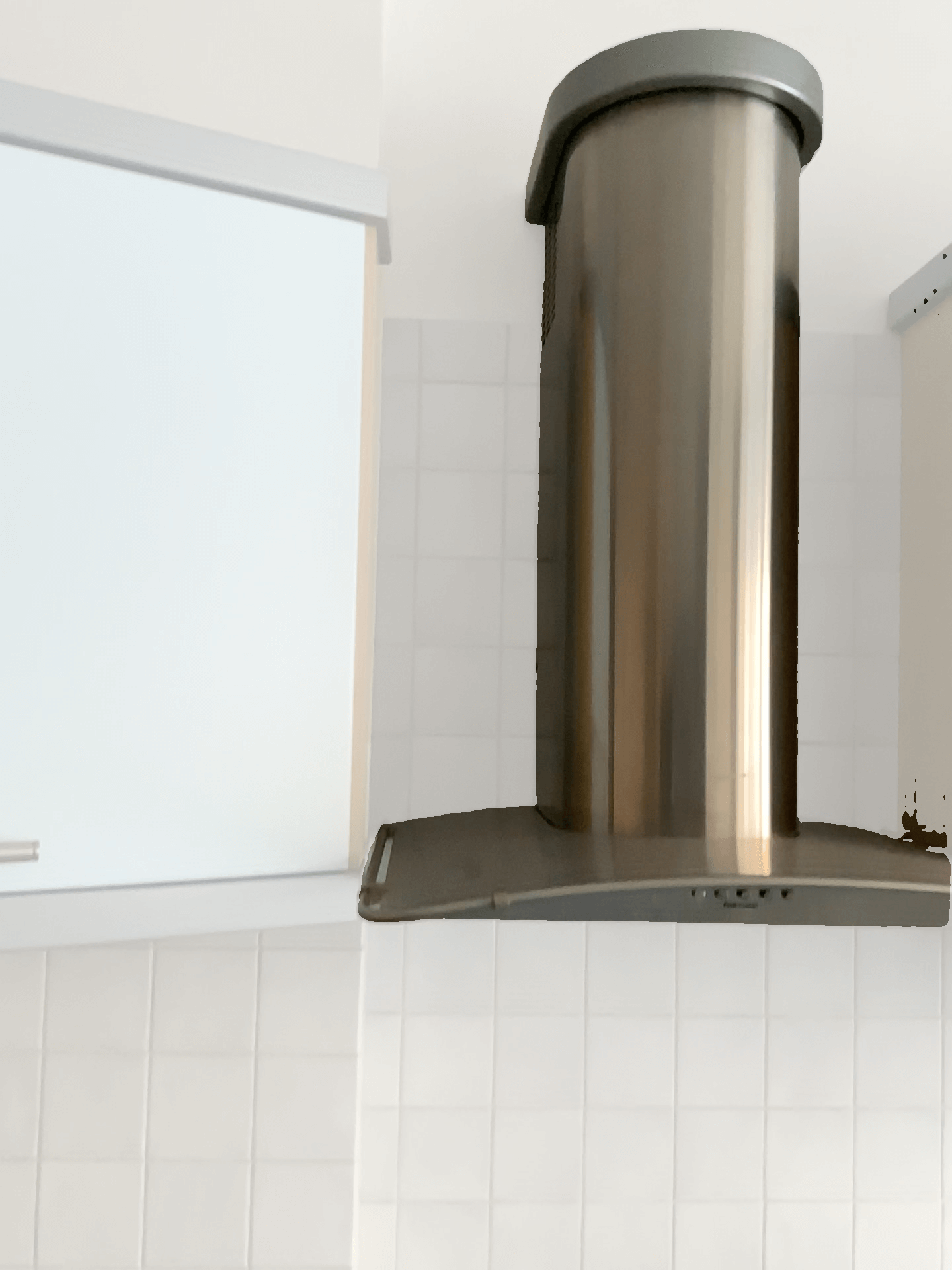}
    % \put(6,43){\small{\textcolor{magenta}{$\mathbf{Q}$}}}
    % \put(54,43){\small{\textcolor{magenta}{$\mathbf{A}$}}}
    \end{overpic}
\end{minipage}
\begin{minipage}[t]{0.19\textwidth}
\vspace{0pt}
    \begin{overpic}[width=1\textwidth]{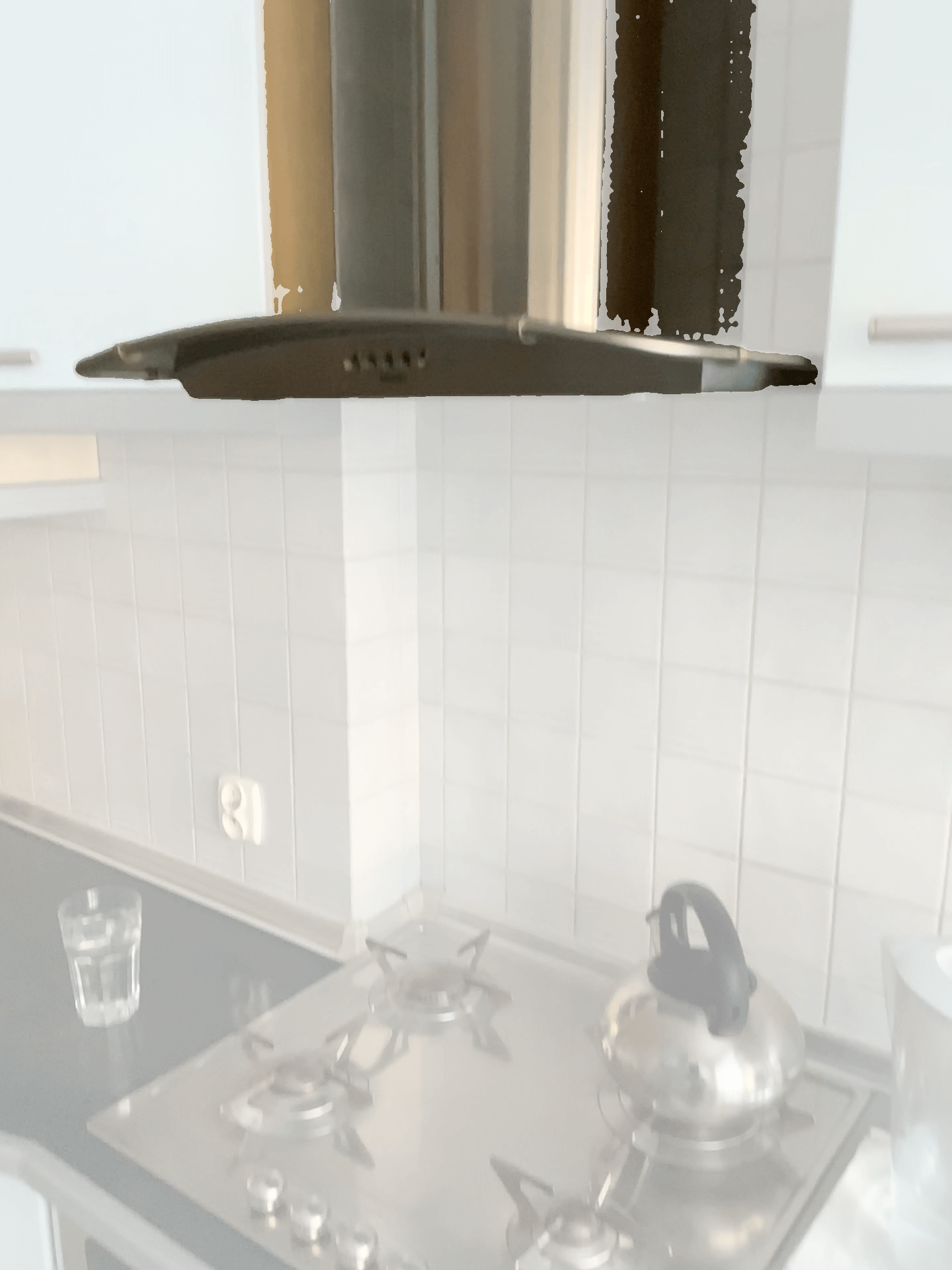}
    % \put(6,43){\small{\textcolor{magenta}{$\mathbf{Q}$}}}
    % \put(54,43){\small{\textcolor{magenta}{$\mathbf{A}$}}}
    \end{overpic}
\end{minipage}
\begin{minipage}[t]{0.19\textwidth}
\vspace{0pt}
    \begin{overpic}[width=1\textwidth]{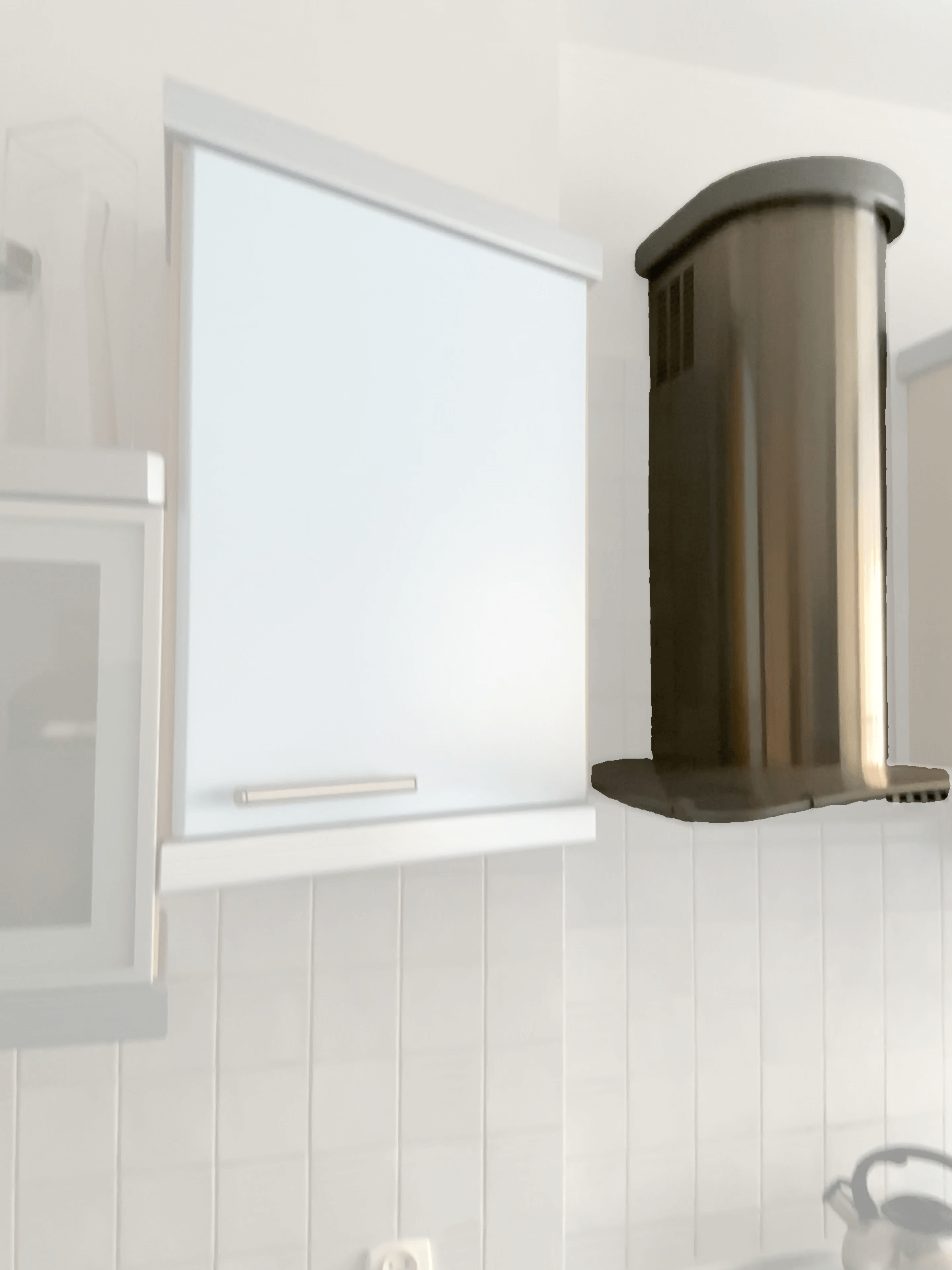}
    % \put(6,43){\small{\textcolor{magenta}{$\mathbf{Q}$}}}
    % \put(54,43){\small{\textcolor{magenta}{$\mathbf{A}$}}}
    \end{overpic}
\end{minipage}\\
\begin{minipage}{0.19\textwidth}
    \centering{\footnotesize{
$S :0.908$ \ $S_{m}: 0.878$\\
$S_{d}: 0.945$ \ $S_{\alpha}: 0.929$\\
    }}
\end{minipage}
\begin{minipage}{0.19\textwidth}
        \centering{\footnotesize{
$S :0.866$ \ $S_{m}: 0.777$\\
$S_{d}: 0.953$ \ $S_{\alpha}: 0.956$\\
    }}
\end{minipage}
\begin{minipage}{0.19\textwidth}
        \centering{\footnotesize{
$S :0.826$ \ $S_{m}: 0.748$\\
$S_{d}: 0.926$ \ $S_{\alpha}: 0.879$\\
    }}
\end{minipage}
\begin{minipage}{0.19\textwidth}
        \centering{\footnotesize{
$S :0.776$ \ $S_{m}: 0.766$\\
$S_{d}: 0.842$ \ $S_{\alpha}: 0.730$\\
    }}
\end{minipage}
\begin{minipage}{0.19\textwidth}
    \centering{\footnotesize{
$S :0.746$ \ $S_{m}: 0.727$\\
$S_{d}: 0.848$ \ $S_{\alpha}: 0.684$\\
    }}
\end{minipage}\\[5mm]
%%%%%%%%%%%%%%%%%%%%%%%%%%%%%%%%%5
%421254 55b56254-acf7-47f9-80f4-368e10816622 Open the bottom drawer of the nightstand with the red table lamp on top
%%%%%%%%%%%%%%%%%%%%%%%%%%%%%%%%%
\begin{minipage}[t]{\textwidth}
    \vspace{0pt}
    \centering{\texttt{Nightstand}}
\end{minipage}\\[0pt]
\begin{minipage}[t]{0.19\textwidth}
    \vspace{0pt}
    \begin{overpic}[width=1\textwidth]{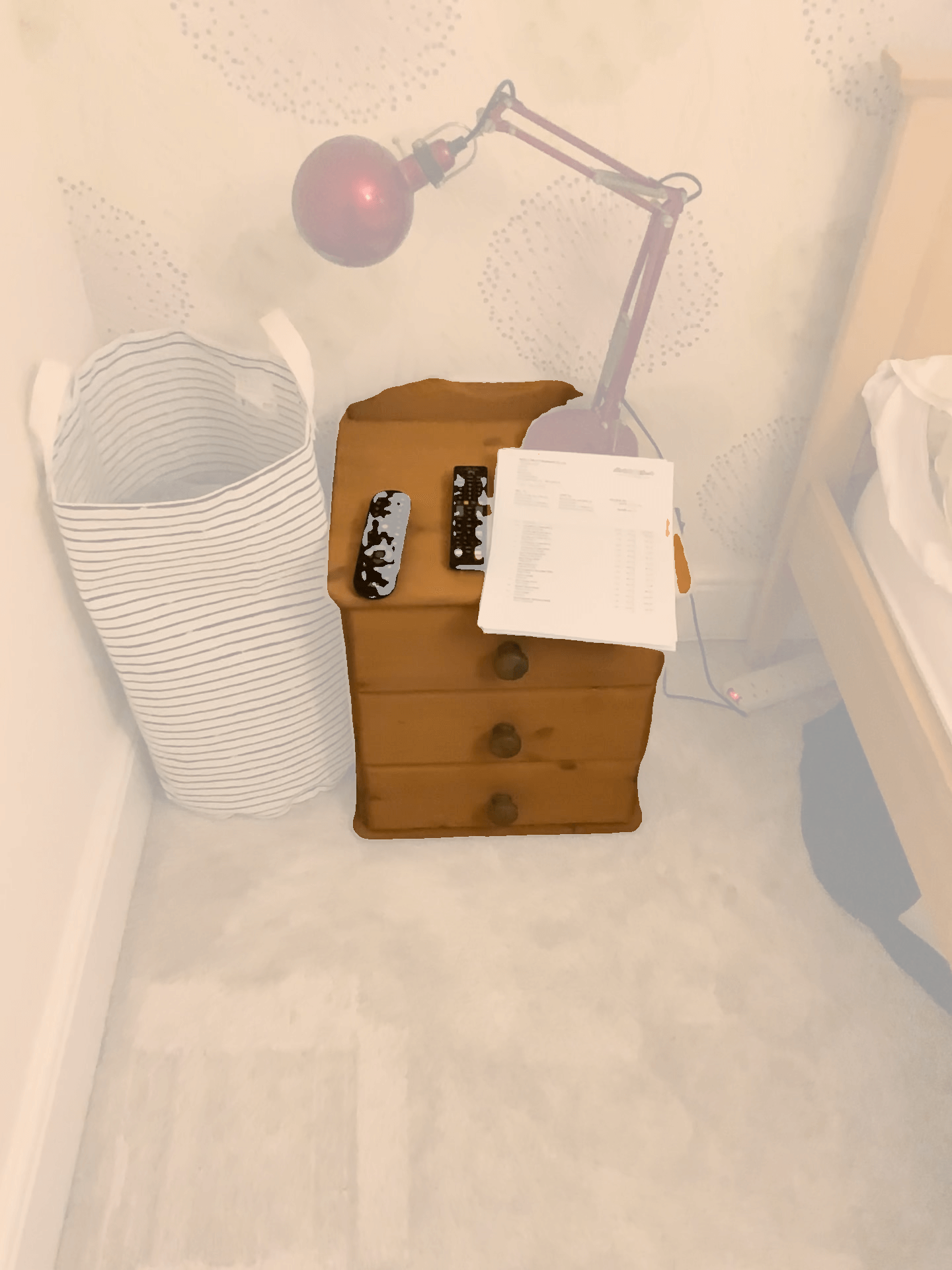}
    % \put(6,43){\small{\textcolor{magenta}{$\mathbf{Q}$}}}
    % \put(54,43){\small{\textcolor{magenta}{$\mathbf{A}$}}}
    \end{overpic}
\end{minipage}
\begin{minipage}[t]{0.19\textwidth}
    \vspace{0pt}
    \begin{overpic}[width=1\textwidth]{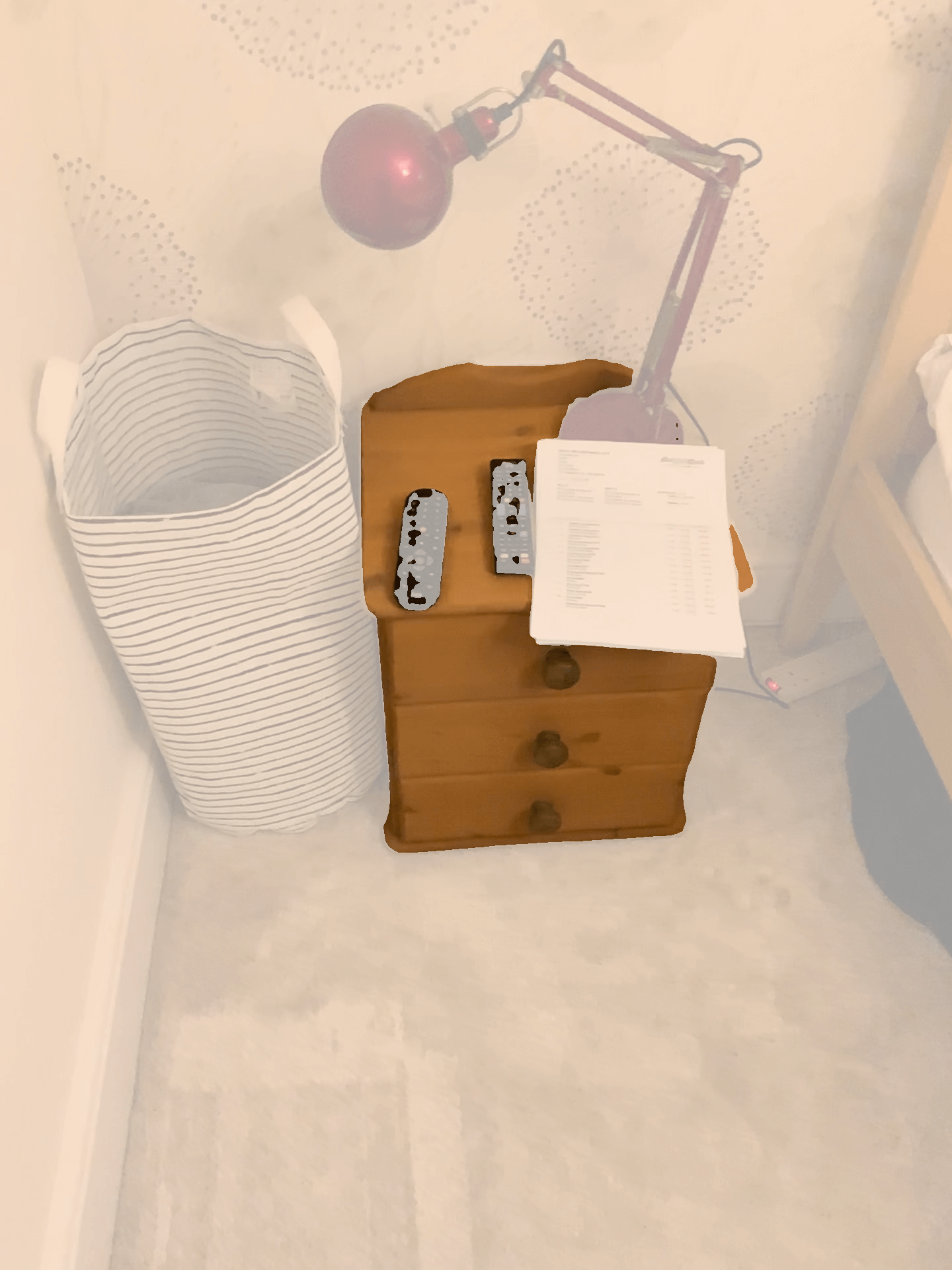}
    % \put(6,43){\small{\textcolor{magenta}{$\mathbf{Q}$}}}
    % \put(54,43){\small{\textcolor{magenta}{$\mathbf{A}$}}}
    \end{overpic}
\end{minipage}
\begin{minipage}[t]{0.19\textwidth}
    \vspace{0pt}
    \begin{overpic}[width=1\textwidth]{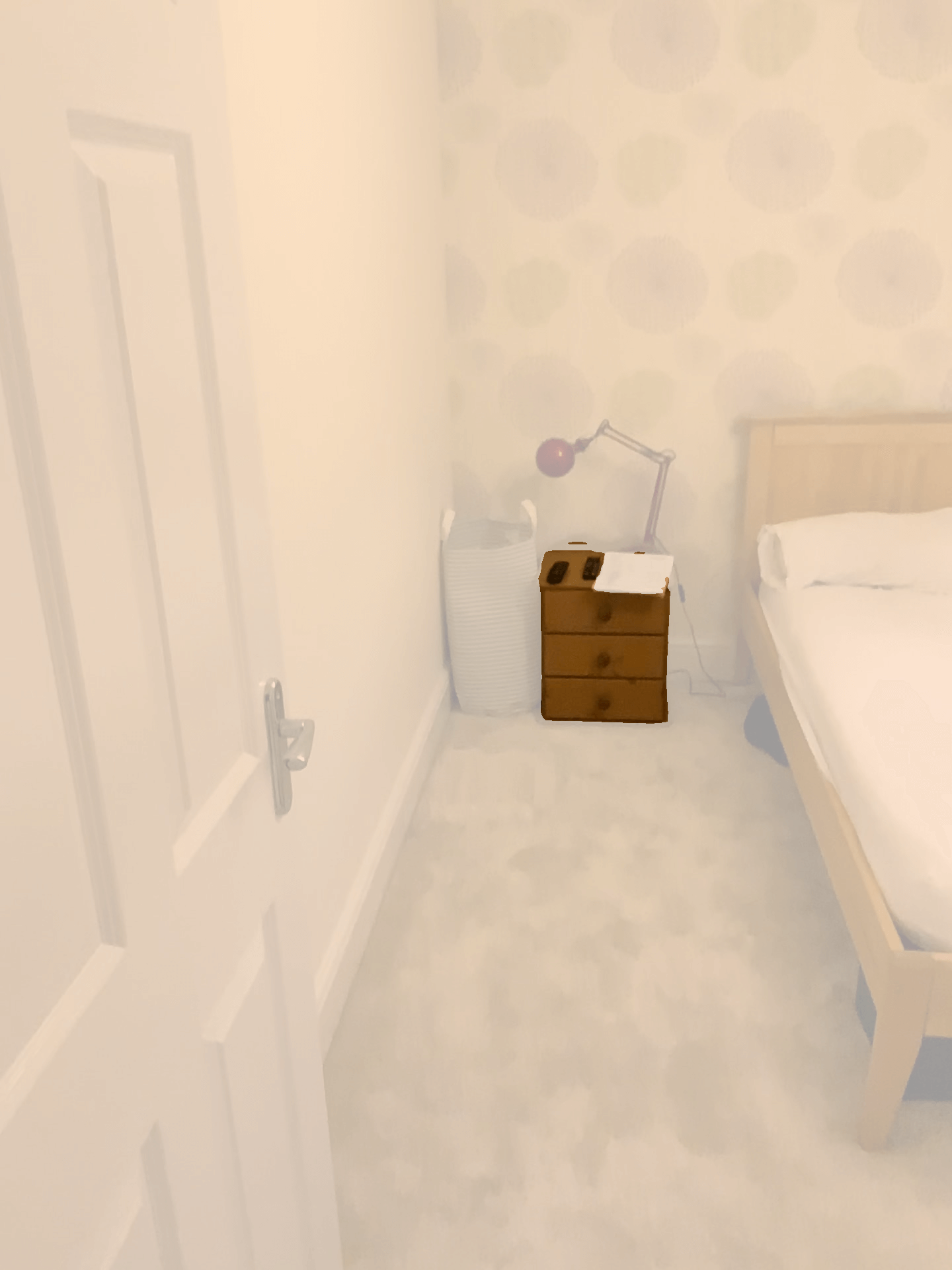}
    % \put(6,43){\small{\textcolor{magenta}{$\mathbf{Q}$}}}
    % \put(54,43){\small{\textcolor{magenta}{$\mathbf{A}$}}}
    \end{overpic}
\end{minipage}
\begin{minipage}[t]{0.19\textwidth}
    \vspace{0pt}
    \begin{overpic}[width=1\textwidth]{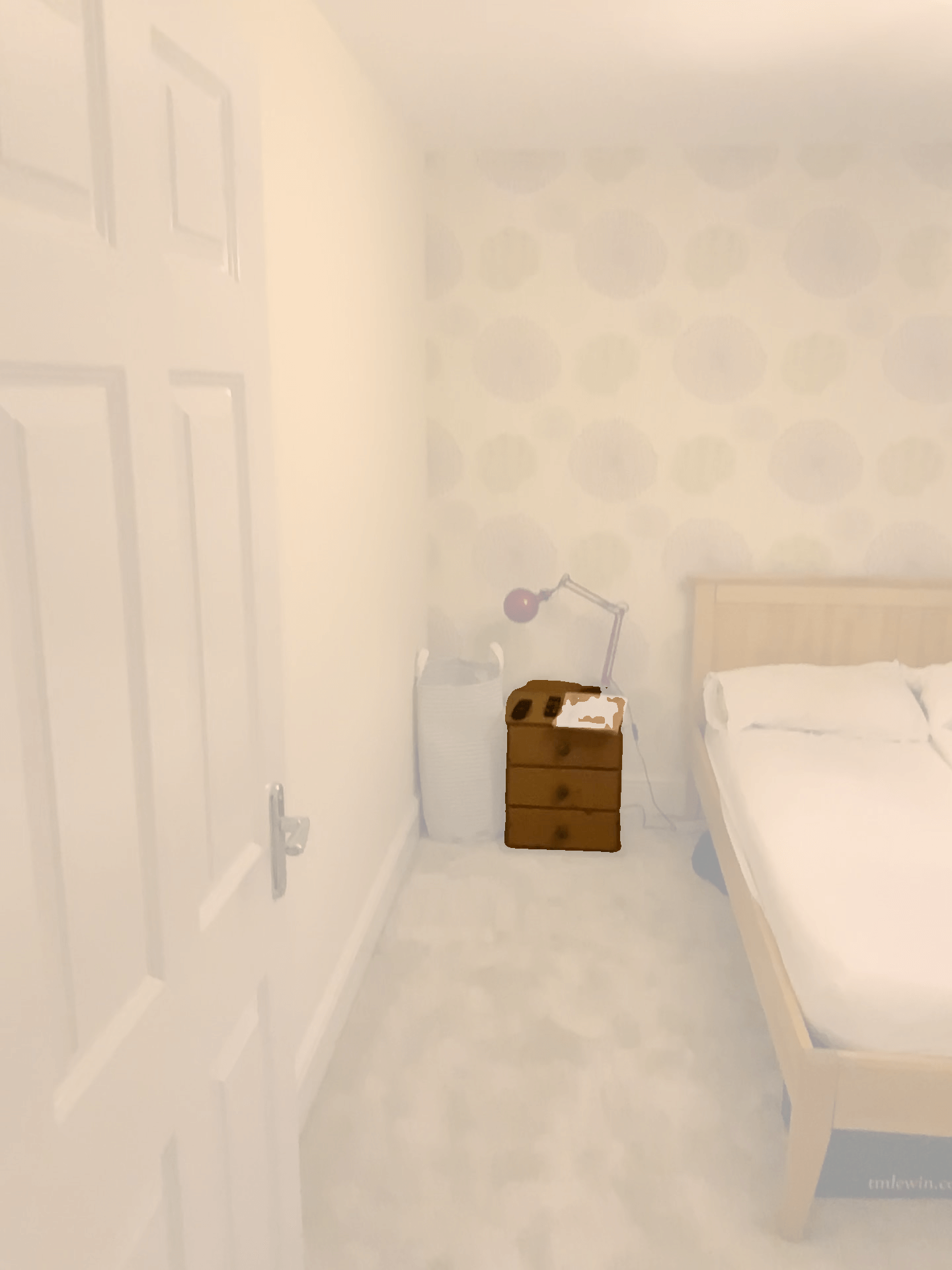}
    % \put(6,43){\small{\textcolor{magenta}{$\mathbf{Q}$}}}
    % \put(54,43){\small{\textcolor{magenta}{$\mathbf{A}$}}}
    \end{overpic}
\end{minipage}
\begin{minipage}[t]{0.19\textwidth}
    \vspace{0pt}
    \begin{overpic}[width=1\textwidth]{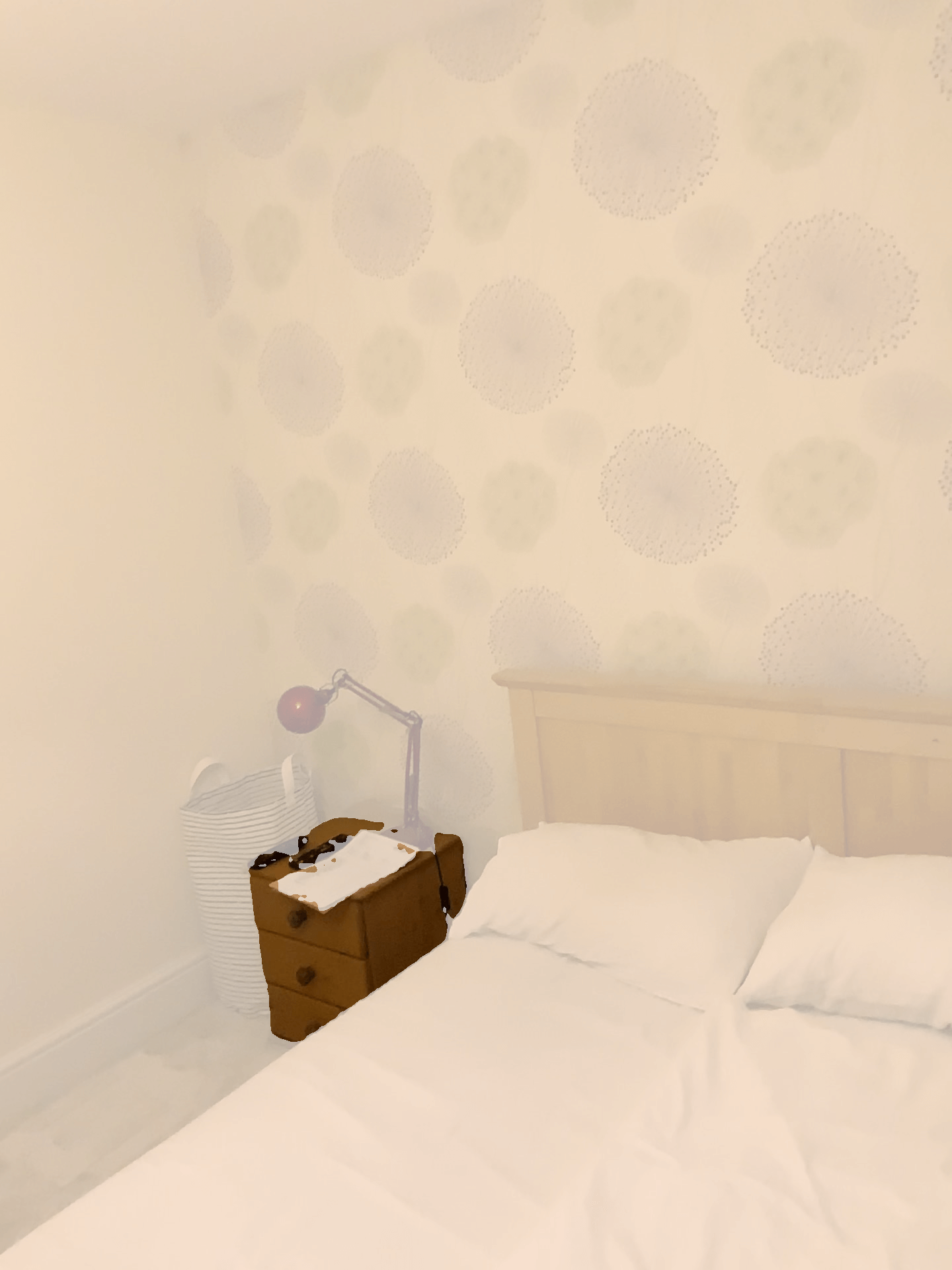}
    % \put(6,43){\small{\textcolor{magenta}{$\mathbf{Q}$}}}
    % \put(54,43){\small{\textcolor{magenta}{$\mathbf{A}$}}}
    \end{overpic}
\end{minipage}\\
\begin{minipage}{0.19\textwidth}
    \centering{\footnotesize{
$S :0.843$ \ $S_{m}: 0.715$\\
$S_{d}: 0.968$ \ $S_{\alpha}: 0.972$\\
    }}
\end{minipage}
\begin{minipage}{0.19\textwidth}
        \centering{\footnotesize{
$S :0.789$ \ $S_{m}: 0.617$\\
$S_{d}: 0.947$ \ $S_{\alpha}: 0.975$\\
    }}
\end{minipage}
\begin{minipage}{0.19\textwidth}
        \centering{\footnotesize{
$S :0.755$ \ $S_{m}: 0.798$\\
$S_{d}: 0.887$ \ $S_{\alpha}: 0.540$\\
    }}
\end{minipage}
\begin{minipage}{0.19\textwidth}
        \centering{\footnotesize{
    $S :0.748$ \ $S_{m}: 0.814$\\
    $S_{d}: 0.877$ \ $S_{\alpha}: 0.486$\\
    }}
\end{minipage}
\begin{minipage}{0.19\textwidth}
    \centering{\footnotesize{
    $S :0.727$ \ $S_{m}: 0.843$\\
    $S_{d}: 0.795$ \ $S_{\alpha}: 0.429$\\
    }}
\end{minipage}\\[5mm]
%%%%%%%%%%%%%%%%%%%%%%%%%%%%%%%%%5
%466916 be3f2542-0036-46fe-8491-ca0fc7a17faa Play a game using the left joystick on the shelf of the TV stand

%%%%%%%%%%%%%%%%%%%%%%%%%%%%%%%%%
\begin{minipage}[t]{\textwidth}
    \vspace{0pt}
    \centering{\texttt{TV stand}}
\end{minipage}\\[0pt]
\begin{minipage}[t]{0.19\textwidth}
\vspace{0pt}
    \begin{overpic}[width=1\textwidth]{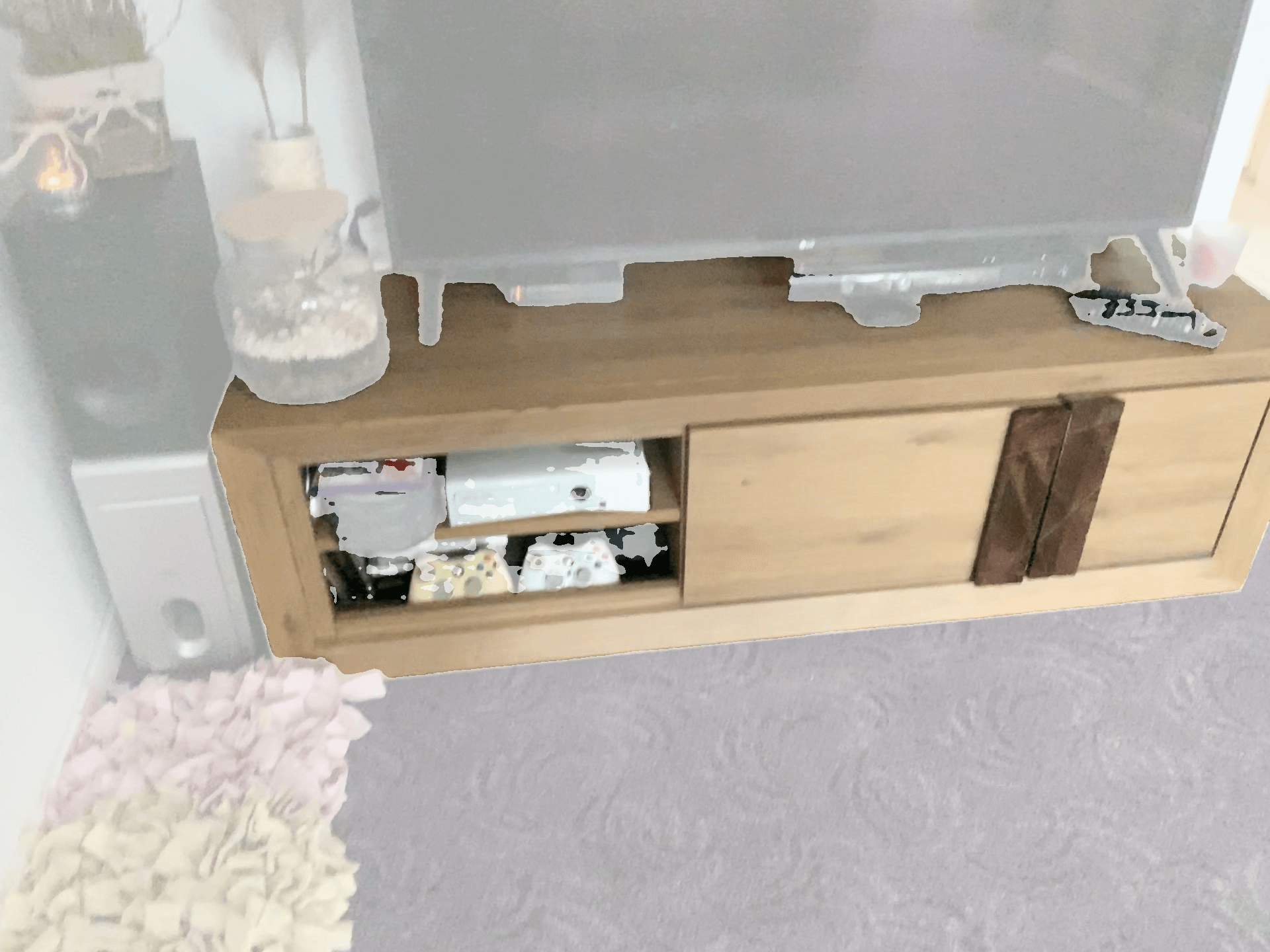}
    % \put(6,43){\small{\textcolor{magenta}{$\mathbf{Q}$}}}
    % \put(54,43){\small{\textcolor{magenta}{$\mathbf{A}$}}}
    \end{overpic}
\end{minipage}
\begin{minipage}[t]{0.19\textwidth}
\vspace{0pt}
    \begin{overpic}[width=1\textwidth]{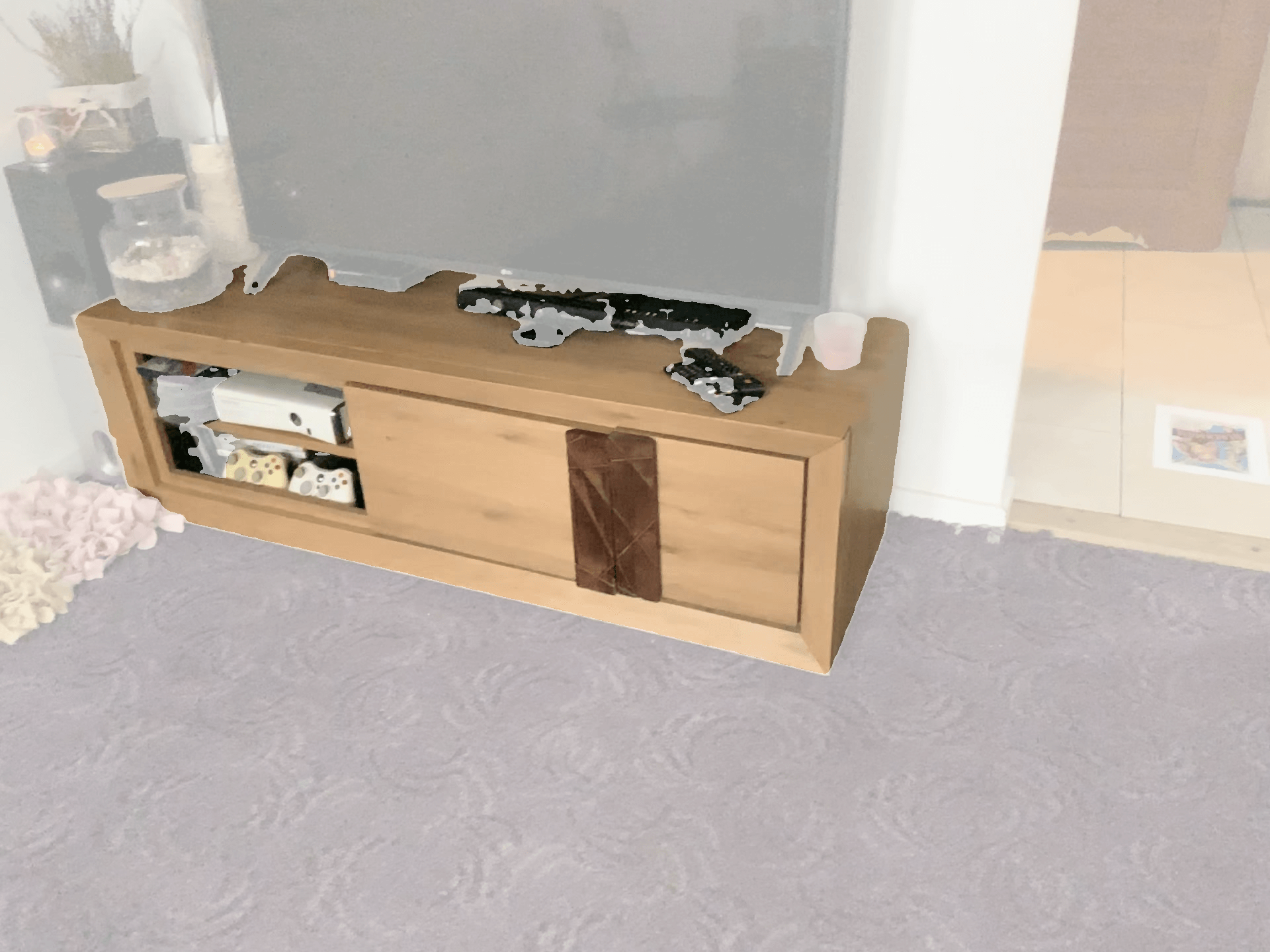}
    % \put(6,43){\small{\textcolor{magenta}{$\mathbf{Q}$}}}
    % \put(54,43){\small{\textcolor{magenta}{$\mathbf{A}$}}}
    \end{overpic}
\end{minipage}
\begin{minipage}[t]{0.19\textwidth}
\vspace{0pt}
    \begin{overpic}[width=1\textwidth]{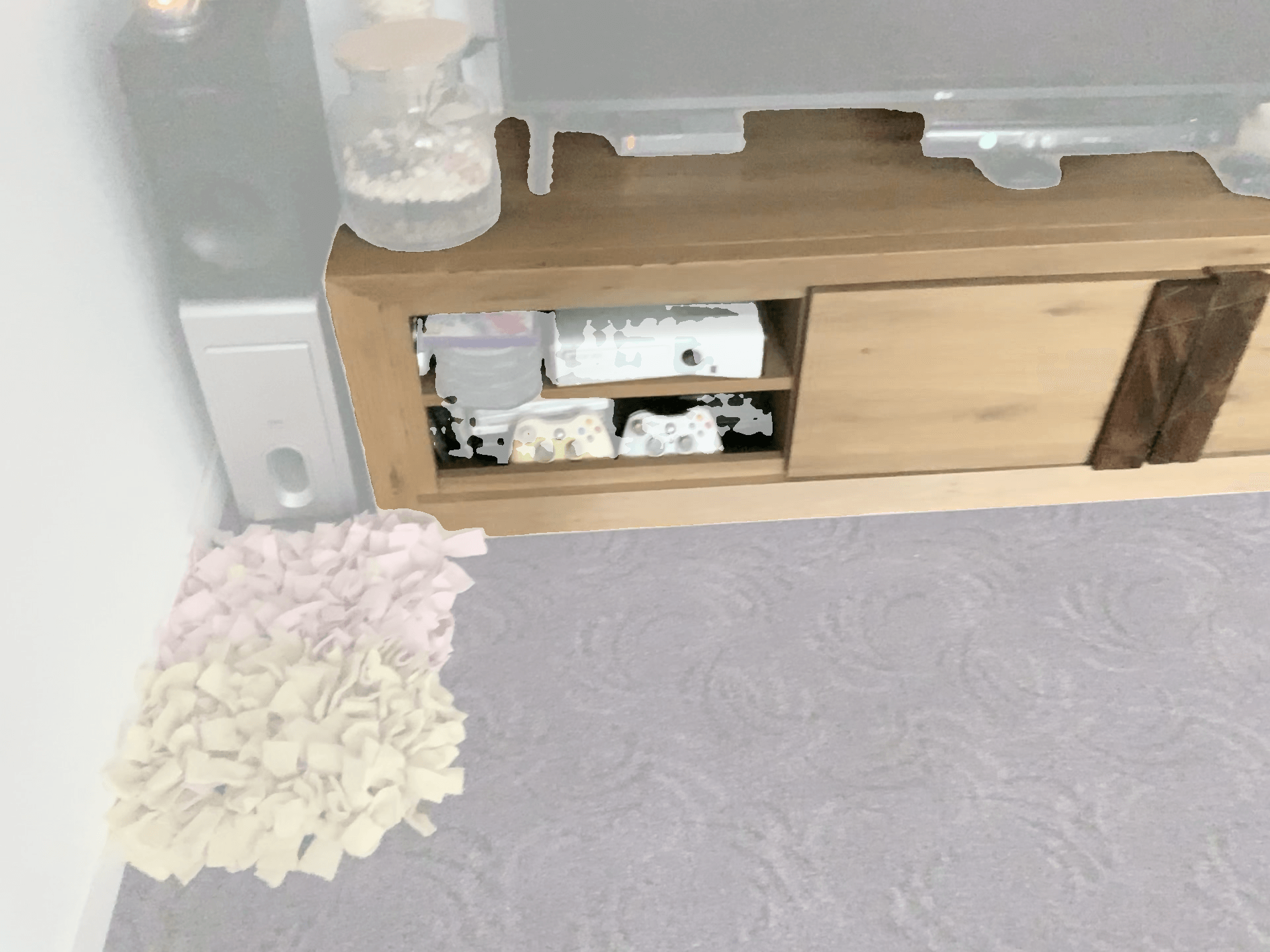}
    % \put(6,43){\small{\textcolor{magenta}{$\mathbf{Q}$}}}
    % \put(54,43){\small{\textcolor{magenta}{$\mathbf{A}$}}}
    \end{overpic}
\end{minipage}
\begin{minipage}[t]{0.19\textwidth}
\vspace{0pt}
    \begin{overpic}[width=1\textwidth]{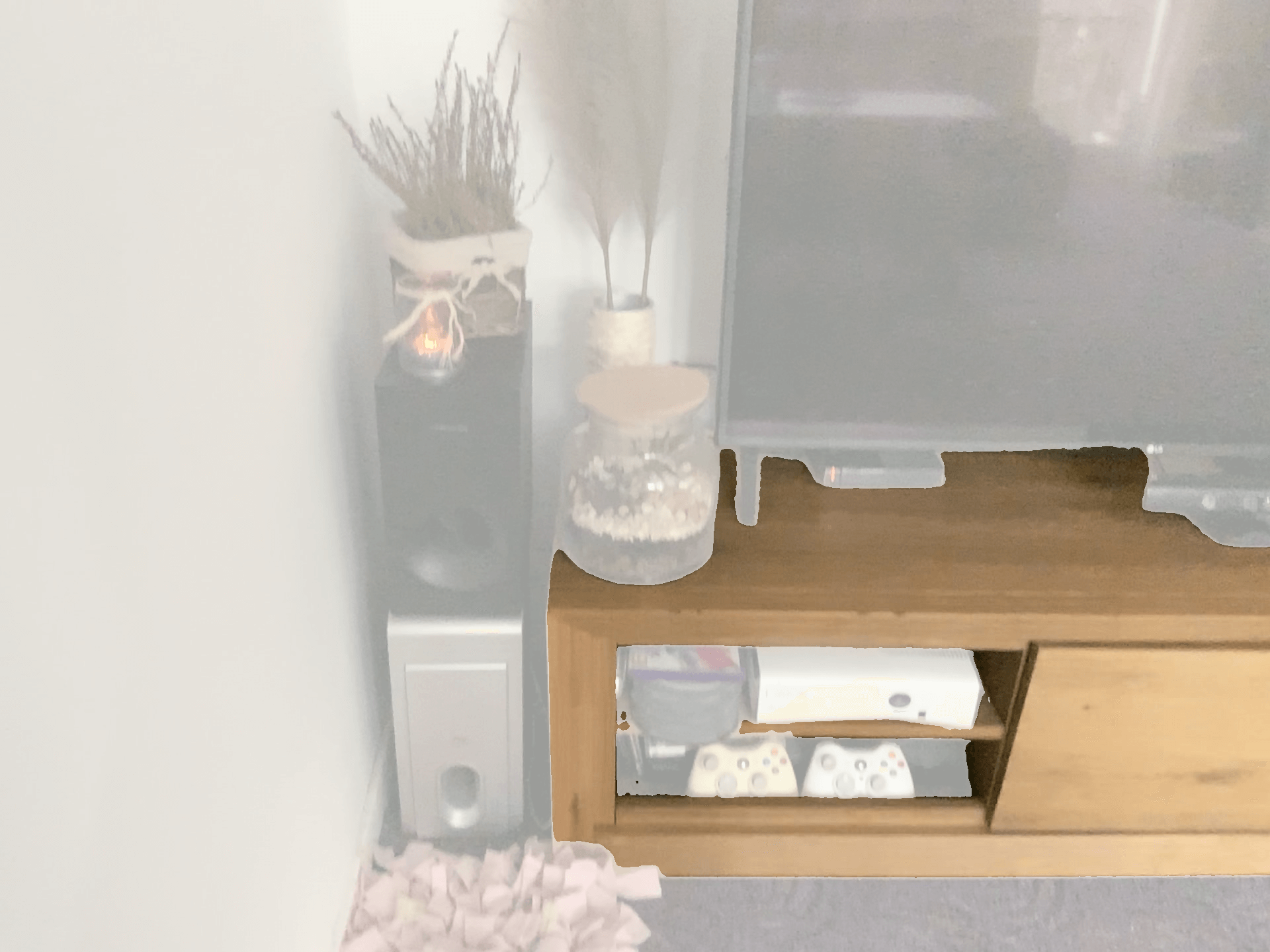}
    % \put(6,43){\small{\textcolor{magenta}{$\mathbf{Q}$}}}
    % \put(54,43){\small{\textcolor{magenta}{$\mathbf{A}$}}}
    \end{overpic}
\end{minipage}
\begin{minipage}[t]{0.19\textwidth}
    \vspace{0pt}
    \begin{overpic}[width=1\textwidth]{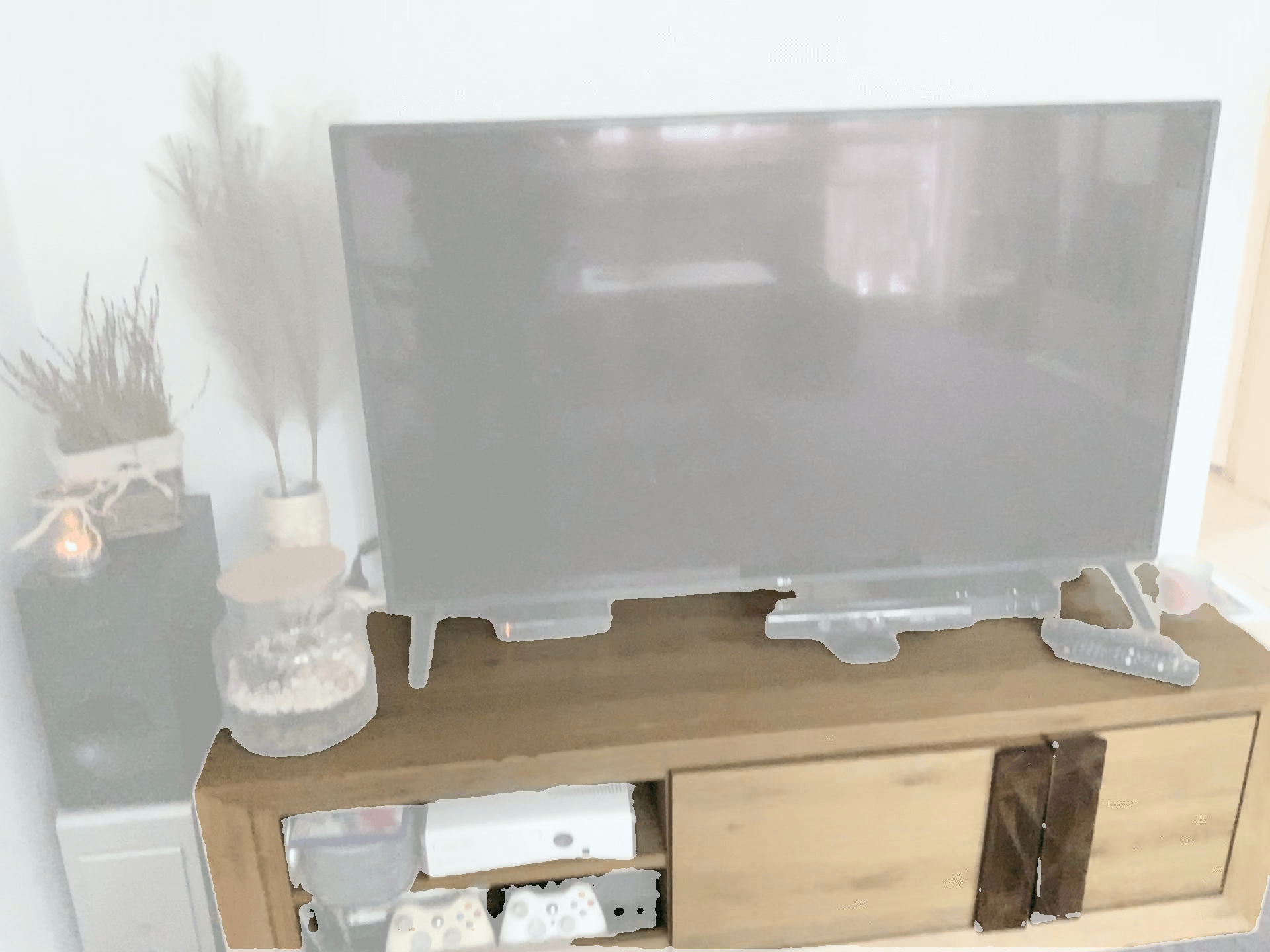}
    % \put(6,43){\small{\textcolor{magenta}{$\mathbf{Q}$}}}
    % \put(54,43){\small{\textcolor{magenta}{$\mathbf{A}$}}}
    \end{overpic}
\end{minipage}\\
\begin{minipage}{0.19\textwidth}
    \centering{\footnotesize{
$S :0.779$ \ $S_{m}: 0.611$\\
$S_{d}: 0.942$ \ $S_{\alpha}: 0.950$\\

    }}
\end{minipage}
\begin{minipage}{0.19\textwidth}
        \centering{\footnotesize{
$S :0.776$ \ $S_{m}: 0.634$\\
$S_{d}: 0.923$ \ $S_{\alpha}: 0.942$\\

    }}
\end{minipage}
\begin{minipage}{0.19\textwidth}
        \centering{\footnotesize{
$S :0.727$ \ $S_{m}: 0.558$\\
$S_{d}: 0.953$ \ $S_{\alpha}: 0.839$\\
    }}
\end{minipage}
\begin{minipage}{0.19\textwidth}
        \centering{\footnotesize{
$S :0.705$ \ $S_{m}: 0.621$\\
$S_{d}: 0.918$ \ $S_{\alpha}: 0.660$\\

    }}
\end{minipage}
\begin{minipage}{0.19\textwidth}
    \centering{\footnotesize{
$S :0.695$ \ $S_{m}: 0.605$\\
$S_{d}: 0.891$ \ $S_{\alpha}: 0.680$\\
    }}
\end{minipage}\\[5mm]
%%%%%%%%%%%%%%%%%%%%%%%%%%%%%%%%%5
%455342 b91e7c76-5d45-4846-a3b7-52b9d7aac5aa Open the top right window
%%%%%%%%%%%%%%%%%%%%%%%%%%%%%%%%%
\begin{minipage}[t]{\textwidth}
    \vspace{0pt}
    \centering{\texttt{Top right window}}
\end{minipage}\\[0pt]
\begin{minipage}[t]{0.19\textwidth}
    \vspace{0pt}
    \begin{overpic}[width=1\textwidth]{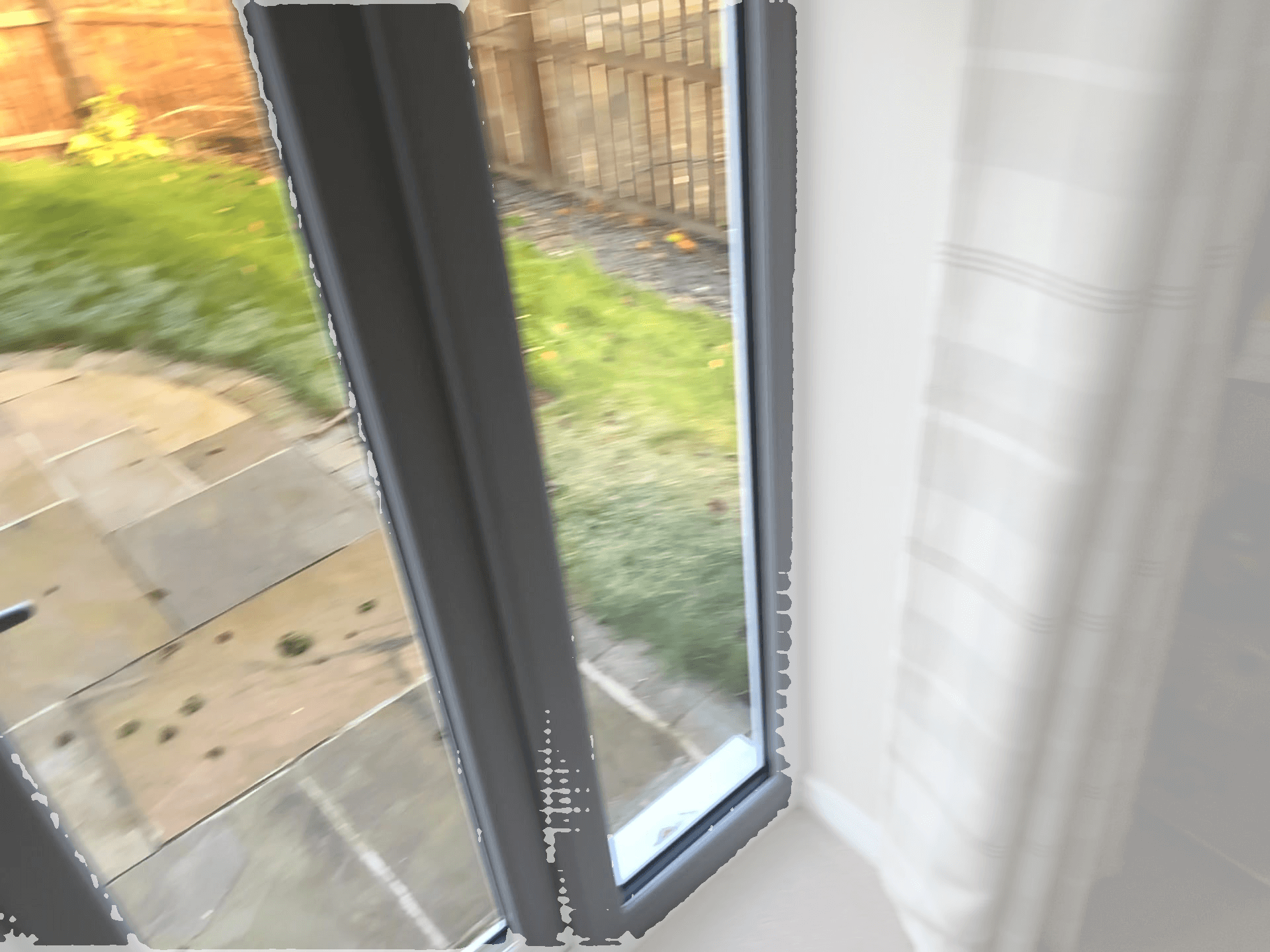}
    % \put(6,43){\small{\textcolor{magenta}{$\mathbf{Q}$}}}
    % \put(54,43){\small{\textcolor{magenta}{$\mathbf{A}$}}}
    \end{overpic}
\end{minipage}
\begin{minipage}[t]{0.19\textwidth}
    \vspace{0pt}
    \begin{overpic}[width=1\textwidth]{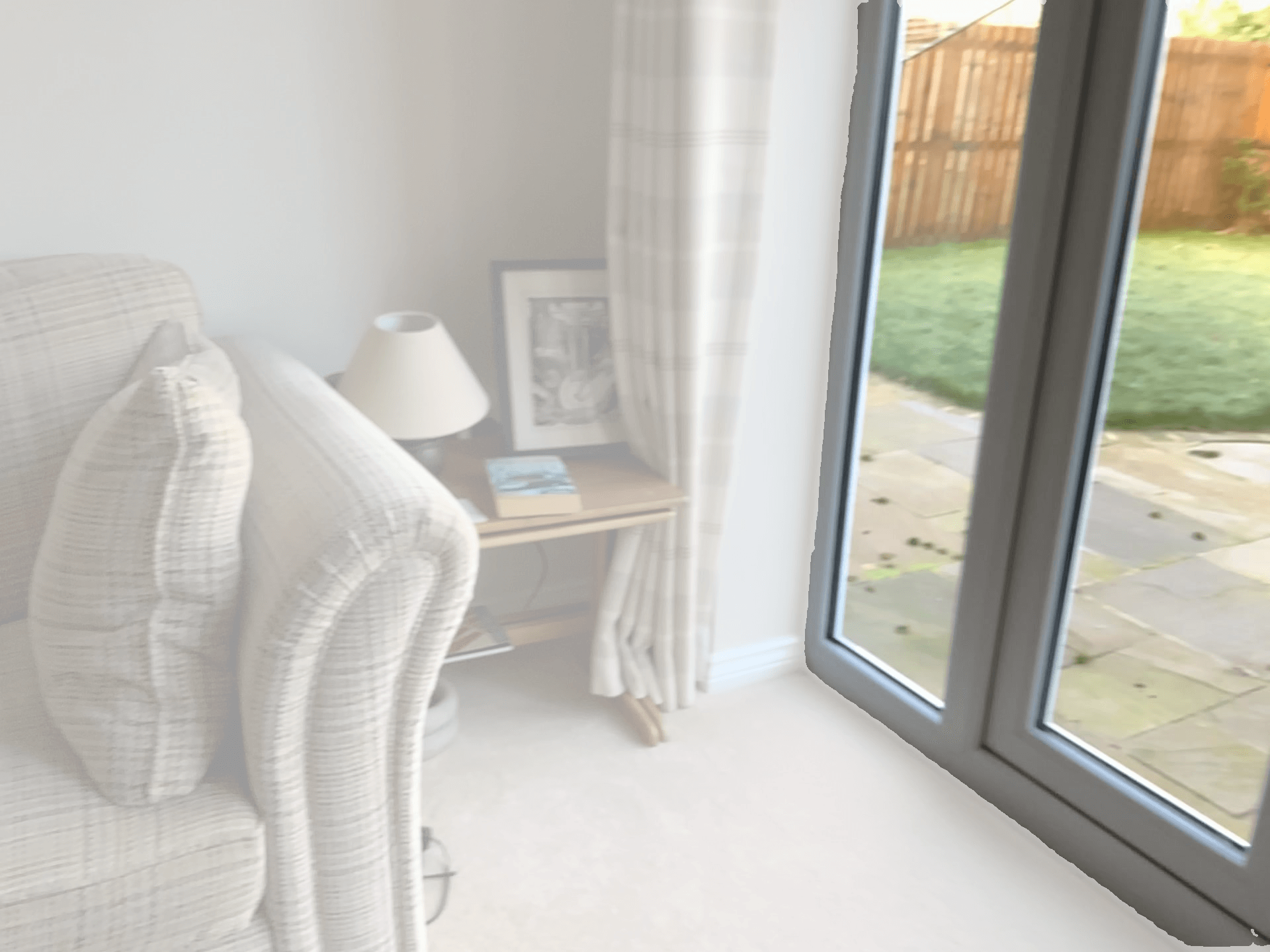}
    % \put(6,43){\small{\textcolor{magenta}{$\mathbf{Q}$}}}
    % \put(54,43){\small{\textcolor{magenta}{$\mathbf{A}$}}}
    \end{overpic}
\end{minipage}
\begin{minipage}[t]{0.19\textwidth}
    \vspace{0pt}
    \begin{overpic}[width=1\textwidth]{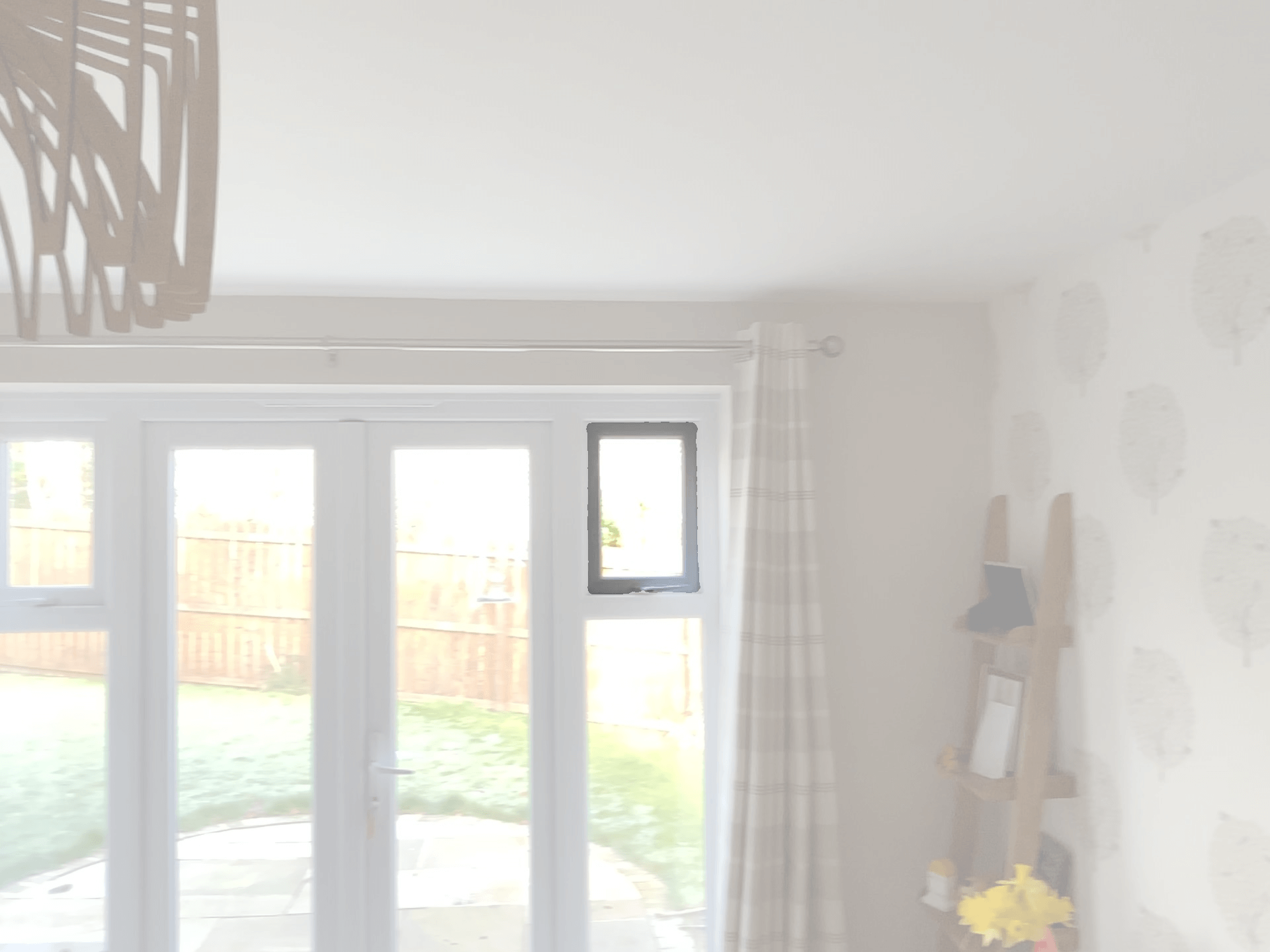}
    % \put(6,43){\small{\textcolor{magenta}{$\mathbf{Q}$}}}
    % \put(54,43){\small{\textcolor{magenta}{$\mathbf{A}$}}}
    \end{overpic}
\end{minipage}
\begin{minipage}[t]{0.19\textwidth}
    \vspace{0pt}
    \begin{overpic}[width=1\textwidth]{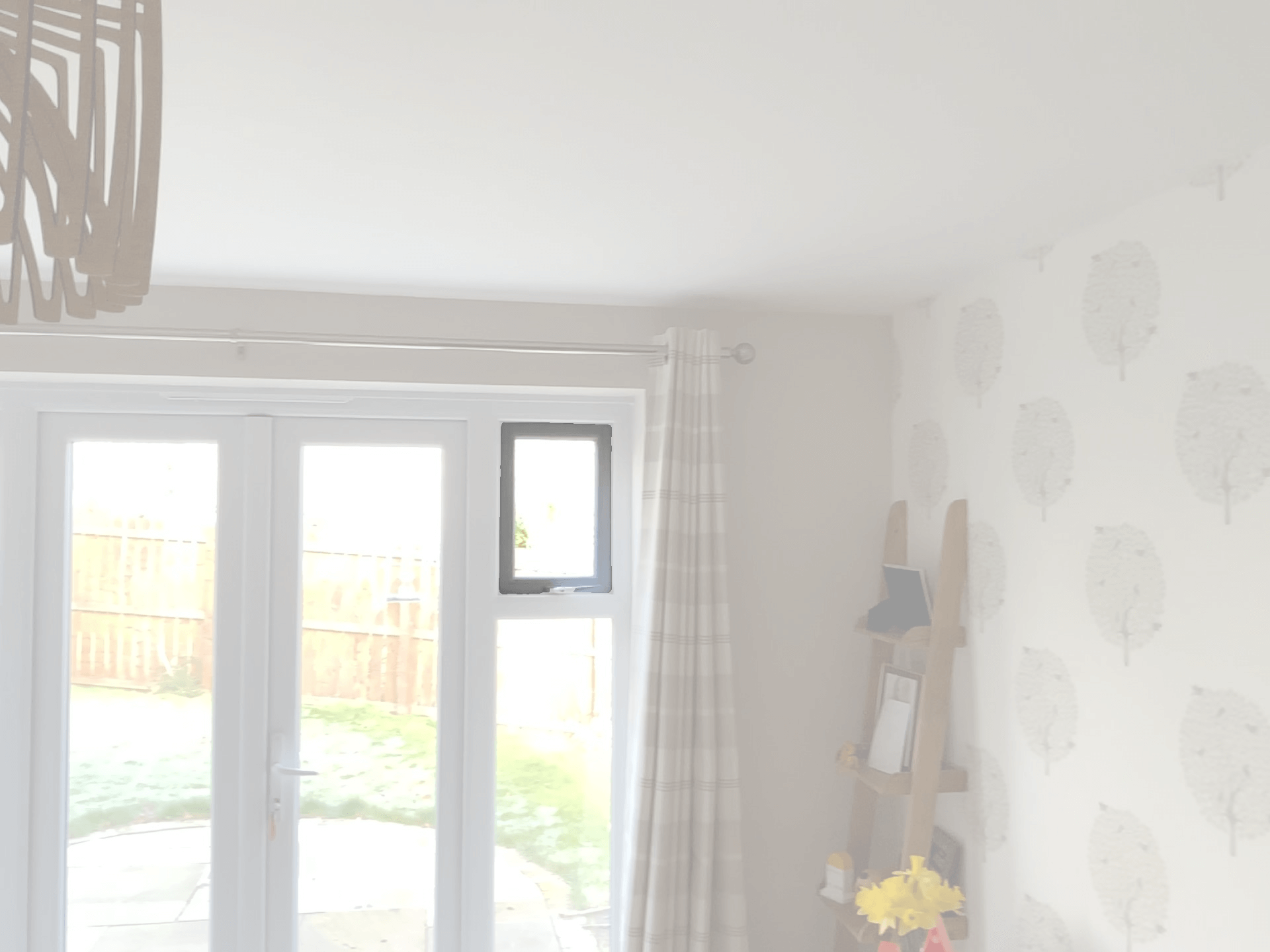}
    % \put(6,43){\small{\textcolor{magenta}{$\mathbf{Q}$}}}
    % \put(54,43){\small{\textcolor{magenta}{$\mathbf{A}$}}}
    \end{overpic}
\end{minipage}
\begin{minipage}[t]{0.19\textwidth}
    \vspace{0pt}
    \begin{overpic}[width=1\textwidth]{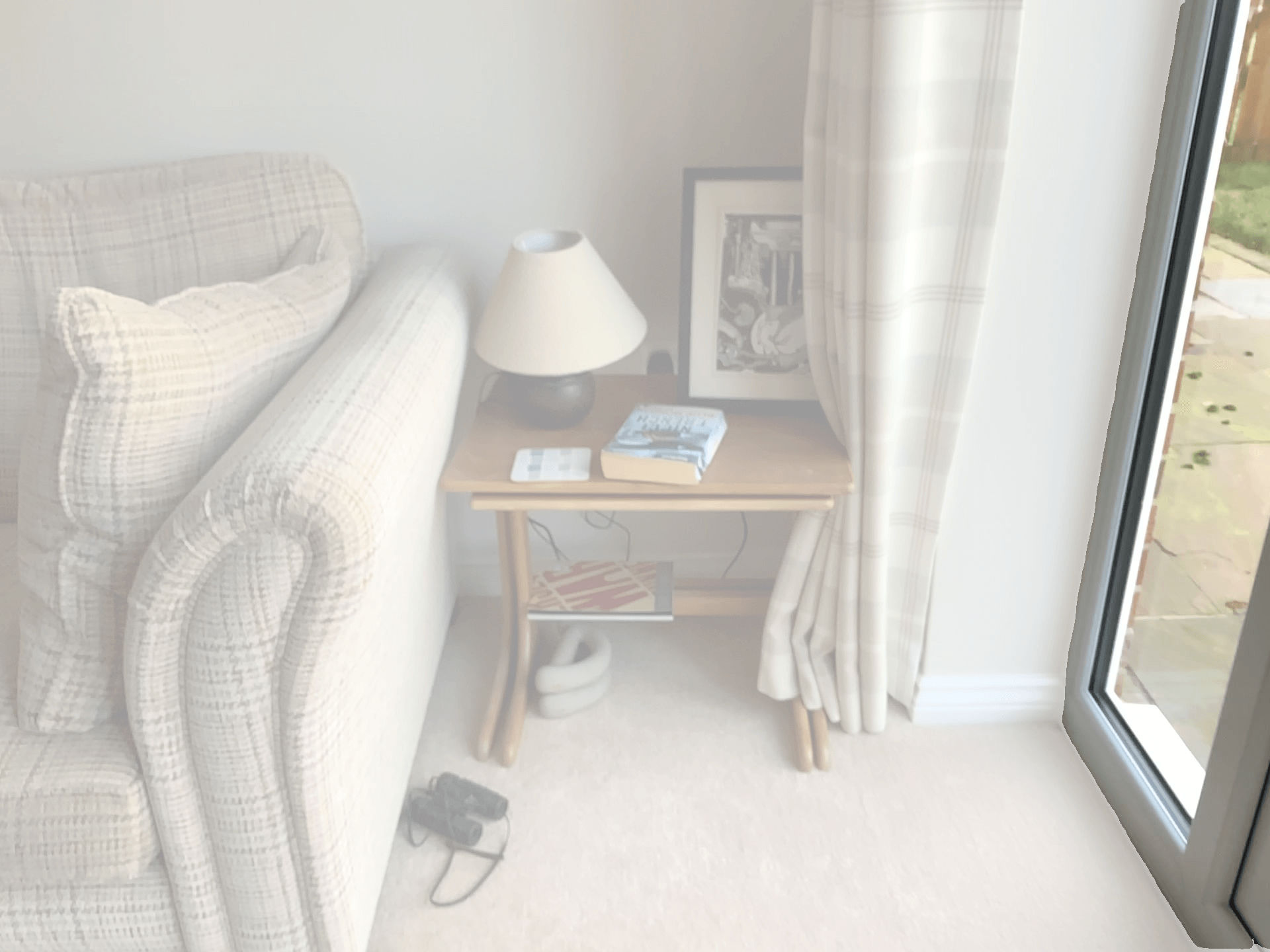}
    % \put(6,43){\small{\textcolor{magenta}{$\mathbf{Q}$}}}
    % \put(54,43){\small{\textcolor{magenta}{$\mathbf{A}$}}}
    \end{overpic}
\end{minipage}\\
\begin{minipage}{0.19\textwidth}
    \centering{\footnotesize{
$S :0.712$ \ $S_{m}: 0.490$\\
$S_{d}: 0.953$ \ $S_{\alpha}: 0.915$\\
    }}
\end{minipage}
\begin{minipage}{0.19\textwidth}
        \centering{\footnotesize{
$S :0.623$ \ $S_{m}: 0.459$\\
$S_{d}: 0.883$ \ $S_{\alpha}: 0.691$\\
    }}
\end{minipage}
\begin{minipage}{0.19\textwidth}
        \centering{\footnotesize{
$S :0.589$ \ $S_{m}: 0.235$\\
$S_{d}: 0.965$ \ $S_{\alpha}: 0.919$\\
    }}
\end{minipage}
\begin{minipage}{0.19\textwidth}
        \centering{\footnotesize{
$S :0.572$ \ $S_{m}: 0.326$\\
$S_{d}: 0.919$ \ $S_{\alpha}: 0.719$\\
    }}
\end{minipage}
\begin{minipage}{0.19\textwidth}
    \centering{\footnotesize{
$S :0.555$ \ $S_{m}: 0.445$\\
$S_{d}: 0.715$ \ $S_{\alpha}: 0.616$\\
    }}
\end{minipage}

%% file: supp/figures/qualit_molmo/qualit_molmo.tex
%%%%%%%%%%%%%%%%%%%%%%%%%%%%%%%%%5
% 421254 Open the bottom drawer of the nightstand with the red table lamp on top
%%%%%%%%%%%%%%%%%%%%%%%%%%%%%%%%%
\begin{minipage}[t]{\textwidth}
    \vspace{0pt}
    \centering{\texttt{Open the bottom drawer of the nightstand with the red table lamp on top}}
\end{minipage}

\begin{minipage}[t]{0.23\textwidth}
\vspace{0pt}
    \begin{overpic}[width=1\textwidth]{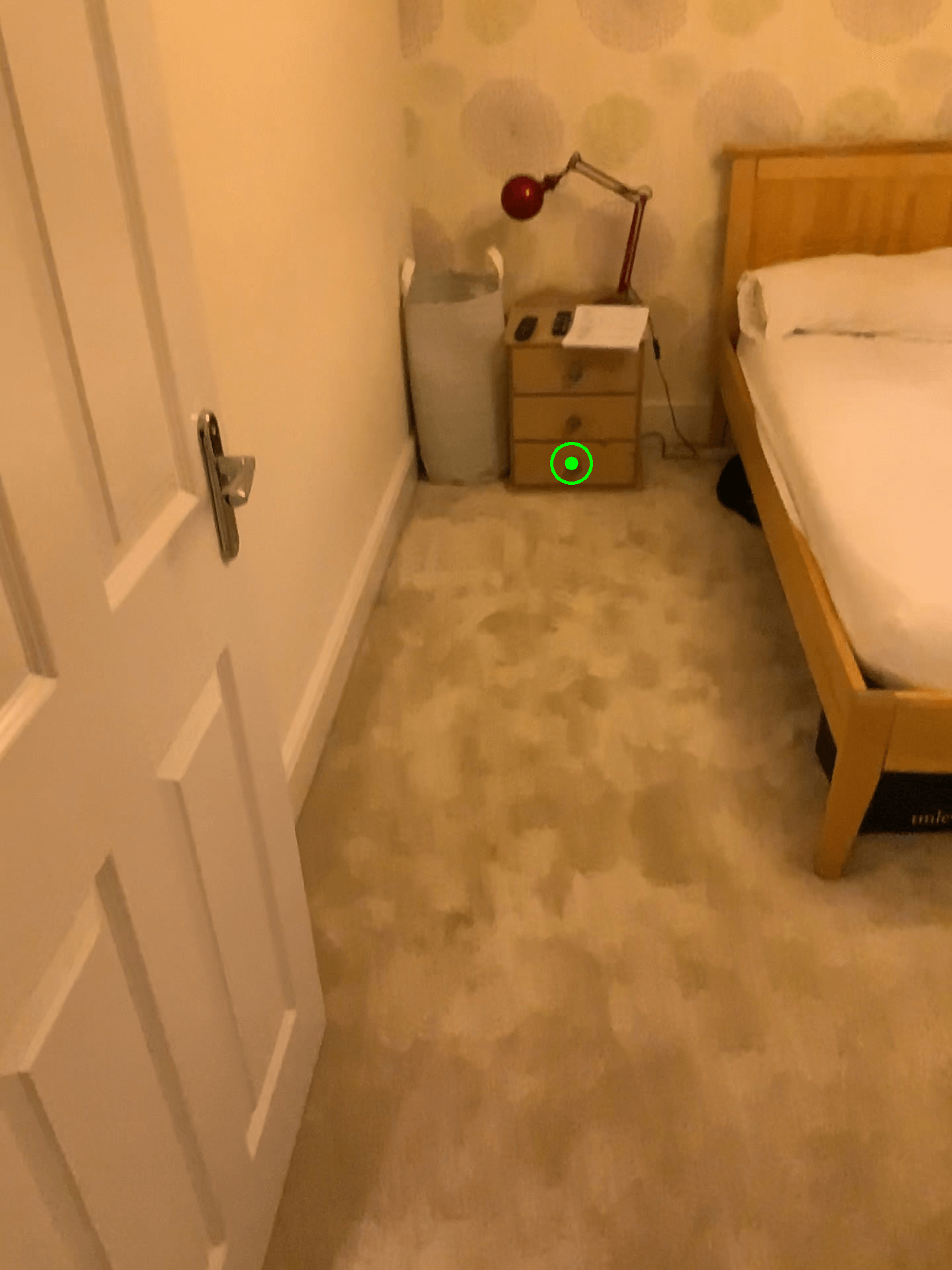}
    
    \end{overpic}
\end{minipage}
\begin{minipage}[t]{0.23\textwidth}
\vspace{0pt}
    \begin{overpic}[width=1\textwidth]{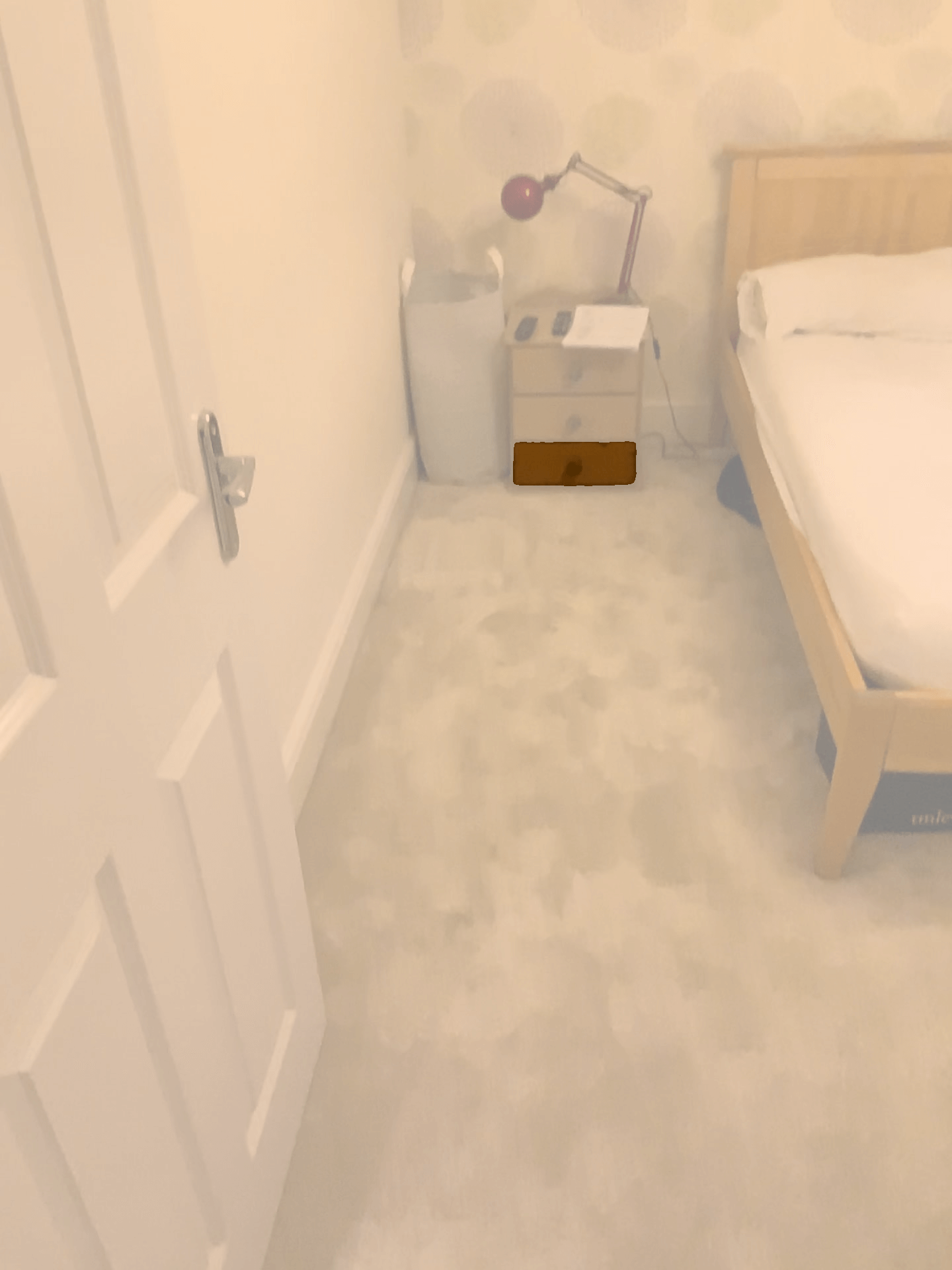}
    
    \end{overpic}
\end{minipage}
\hfill
\begin{minipage}[t]{0.23\textwidth}
\vspace{0pt}
    \begin{overpic}[width=1\textwidth]{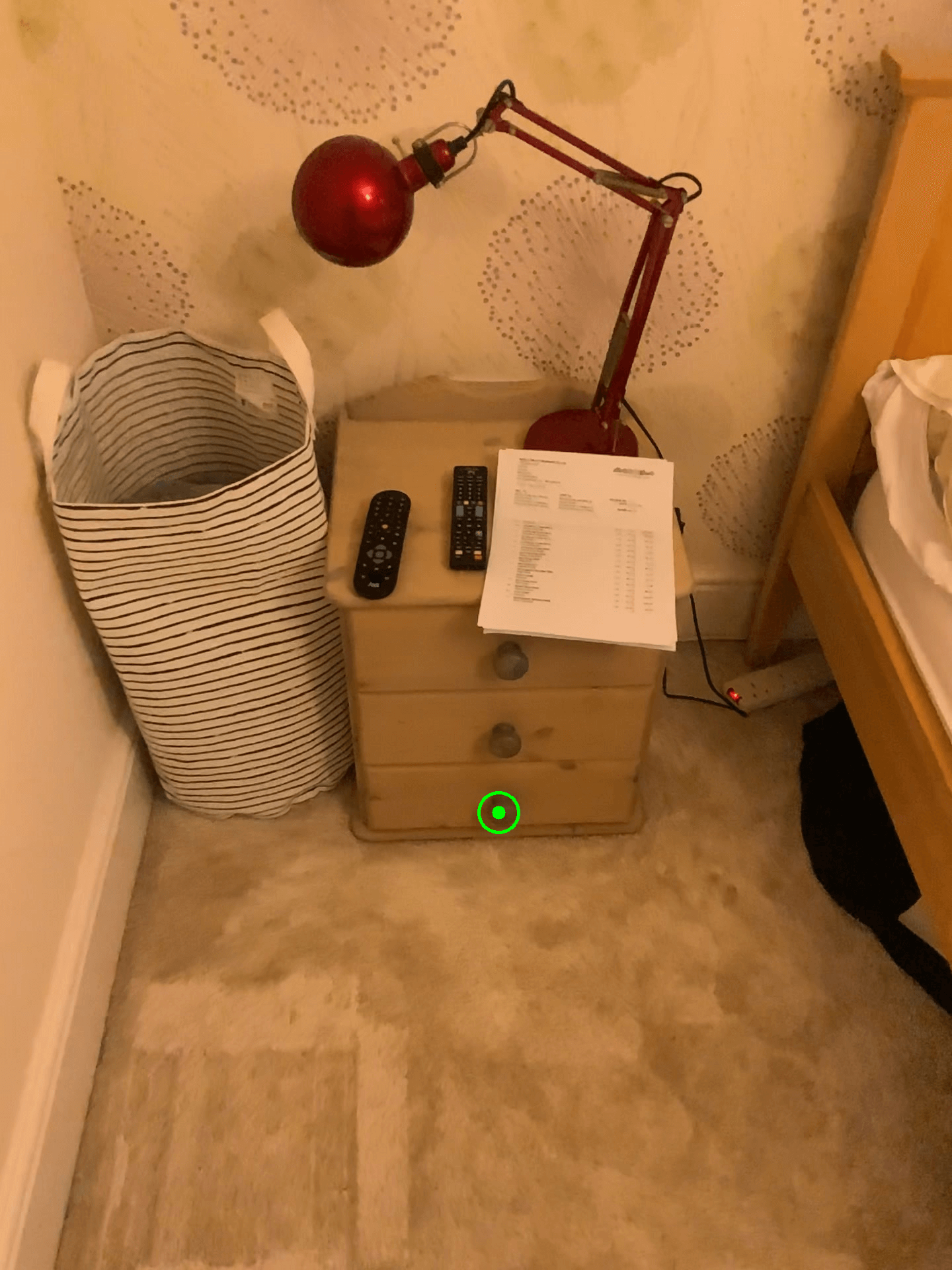}
    
    \end{overpic}
\end{minipage}
\begin{minipage}[t]{0.23\textwidth}
\vspace{0pt}
    \begin{overpic}[width=1\textwidth]{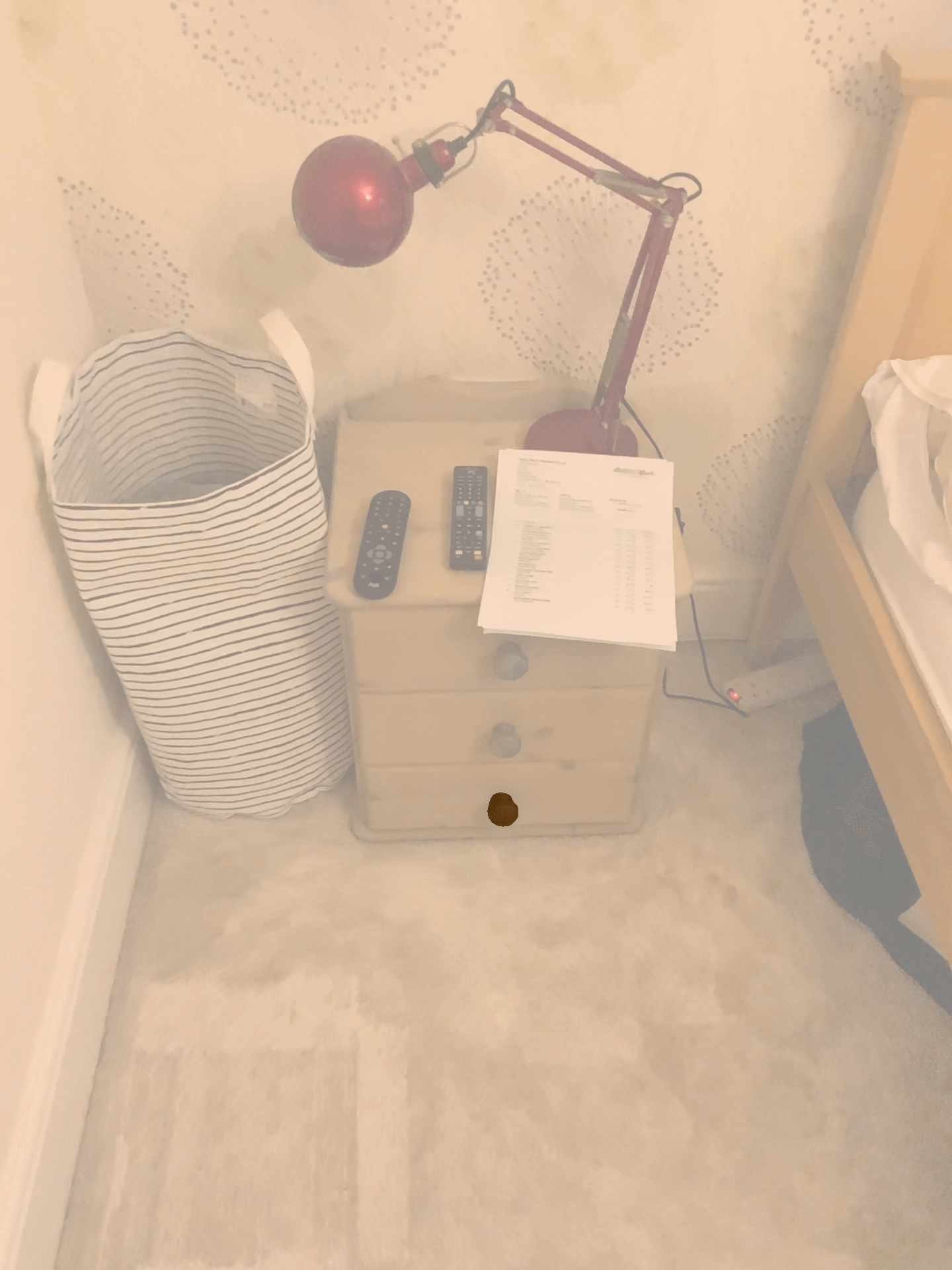}
    
    \end{overpic}
\end{minipage}\\[5mm]
%%%%%%%%%%%%%%%%%%%%%%%%%%%%%%%%%5
% 421393 729dc0d8-571c-44e5-9dc4-05045524dcf5 
%%%%%%%%%%%%%%%%%%%%%%%%%%%%%%%%%
\begin{minipage}[t]{\textwidth}
\vspace{0pt}
    \centering{\texttt{Adjust the room's temperature using the radiator dial}}
\end{minipage}\\
\begin{minipage}[t]{0.23\textwidth}
\vspace{0pt}
    \begin{overpic}[width=1\textwidth]{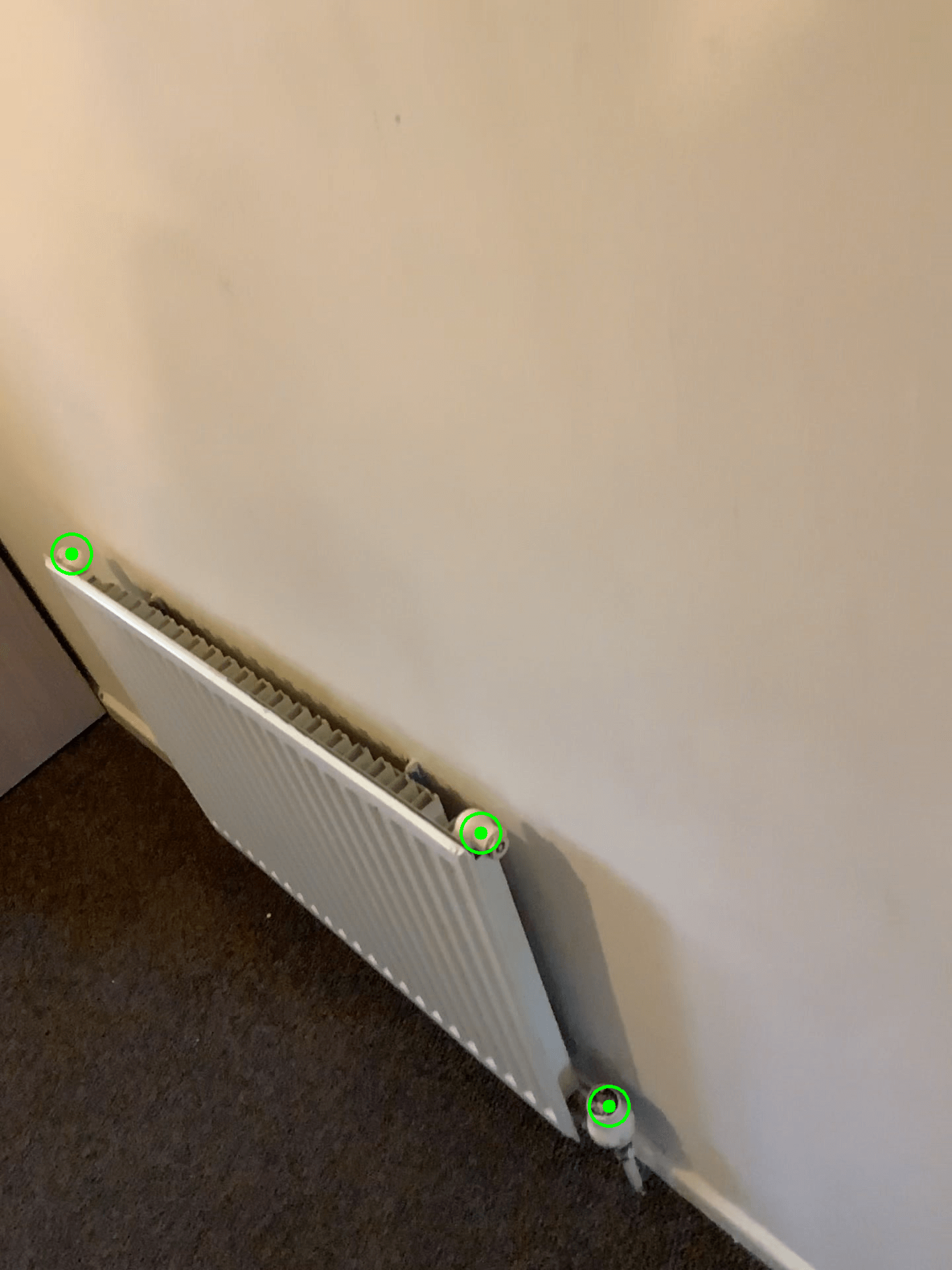}
    
    \end{overpic}
\end{minipage}
\begin{minipage}[t]{0.23\textwidth}
\vspace{0pt}
    \begin{overpic}[width=1\textwidth]{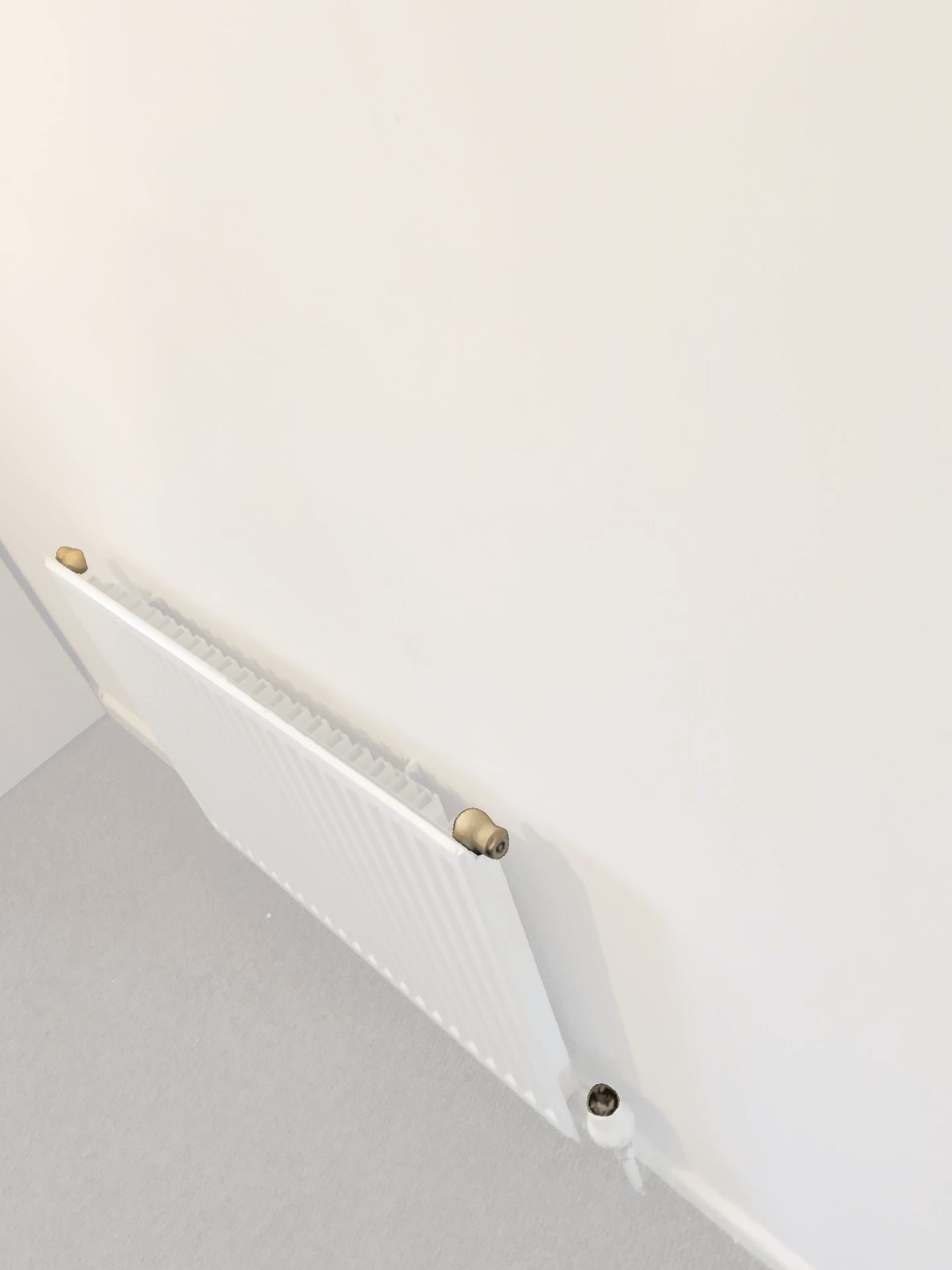}
    
    \end{overpic}
\end{minipage}
\hfill
\begin{minipage}[t]{0.23\textwidth}
\vspace{0pt}
    \begin{overpic}[width=1\textwidth]{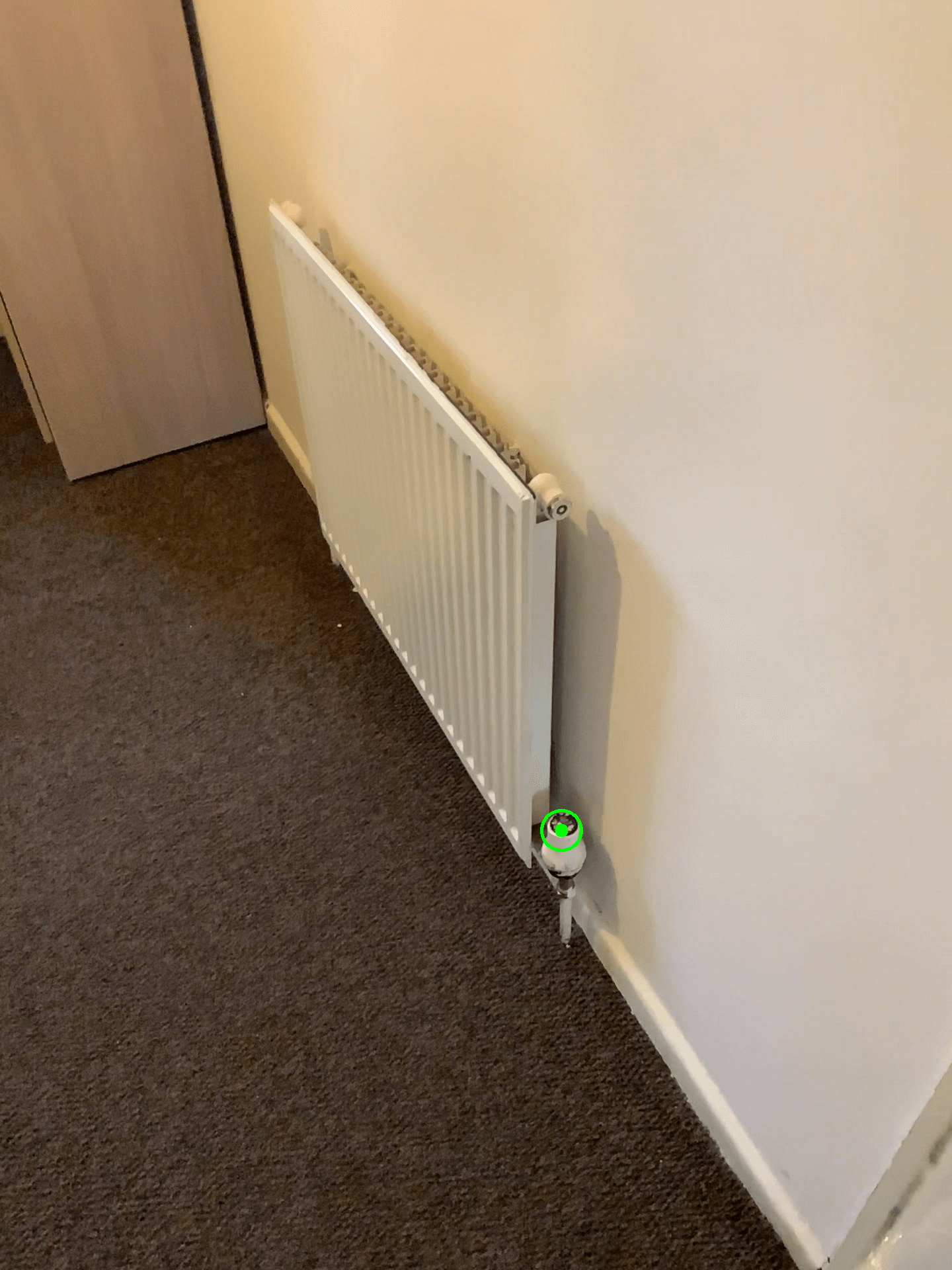}
    
    \end{overpic}
\end{minipage}
\begin{minipage}[t]{0.23\textwidth}
\vspace{0pt}
    \begin{overpic}[width=1\textwidth]{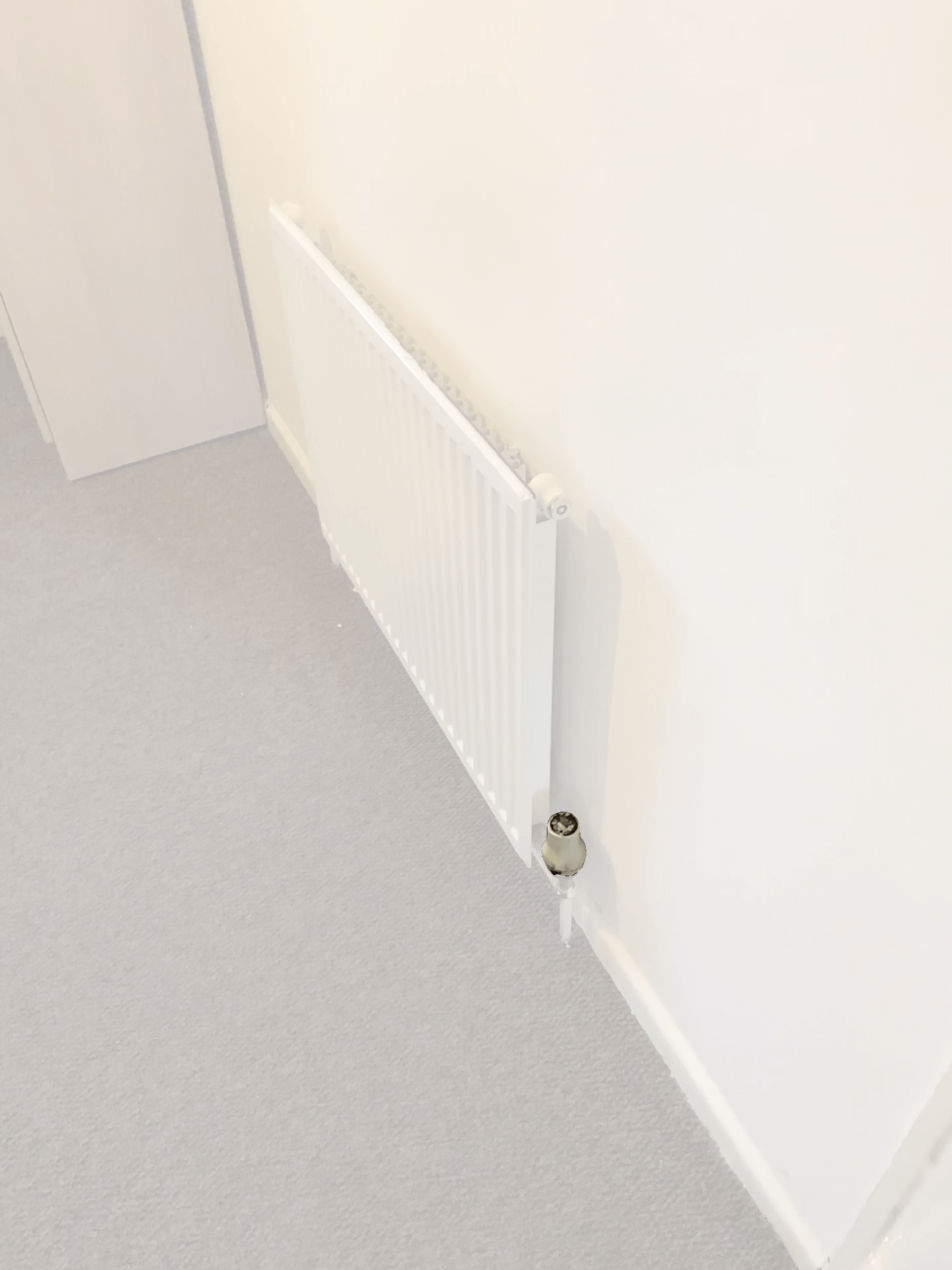}
    
    \end{overpic}
\end{minipage}\\[5mm]
%%%%%%%%%%%%%%%%%%%%%%%%%%%%%%%%%5
% 467330 fea1c7a5-d283-43d3-b674-dd32ed116dc6 Turn on the TV using the remote on the blue couch
%%%%%%%%%%%%%%%%%%%%%%%%%%%%%%%%%
\begin{minipage}[t]{\textwidth}
\vspace{0pt}
    \centering{\texttt{Turn on the TV using the remote on the blue couch}}
\end{minipage}\\
\begin{minipage}[t]{0.23\textwidth}
\vspace{0pt}
    \begin{overpic}[width=1\textwidth]{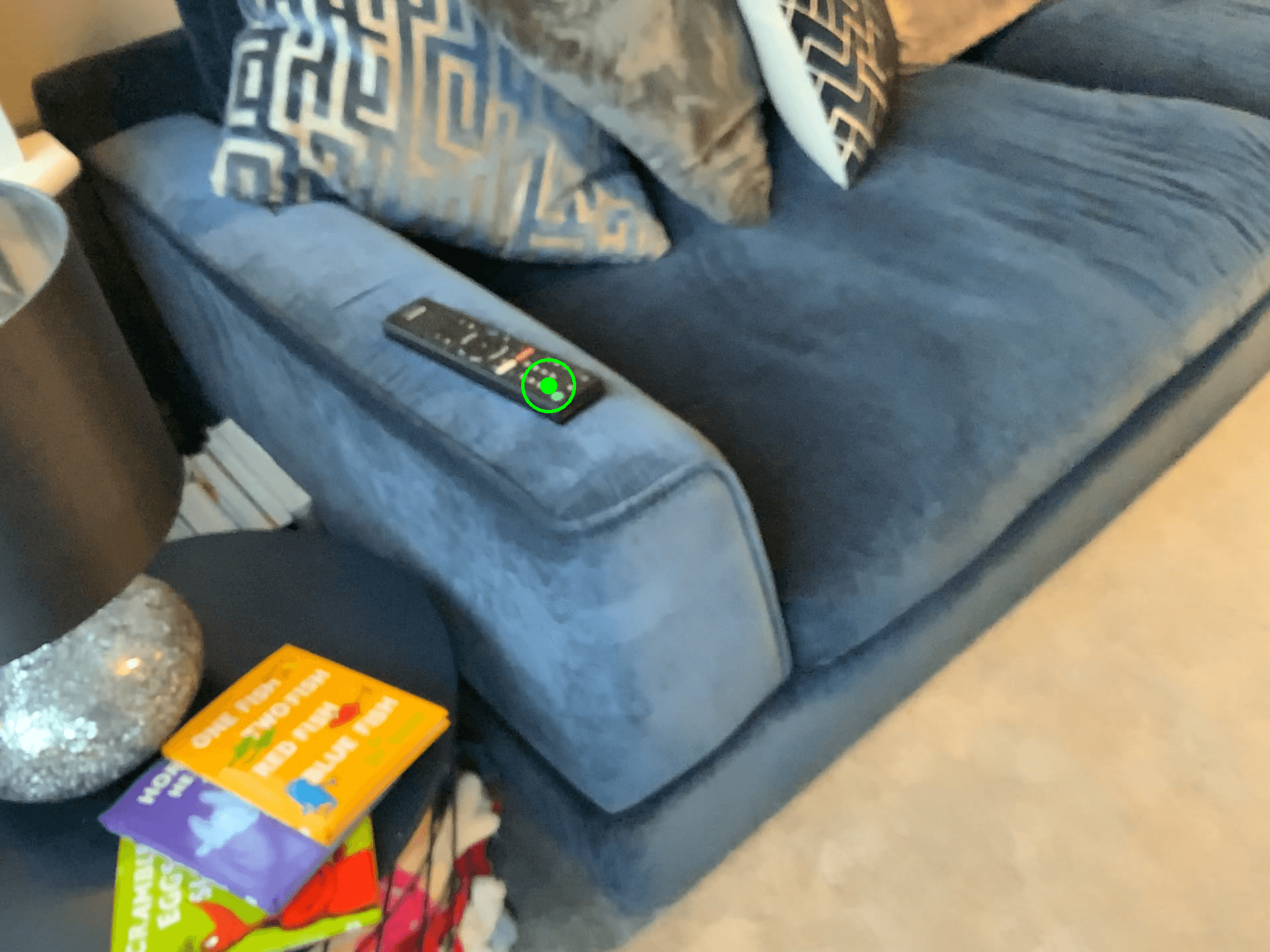}
    
    \end{overpic}
\end{minipage}
\begin{minipage}[t]{0.23\textwidth}
\vspace{0pt}
    \begin{overpic}[width=1\textwidth]{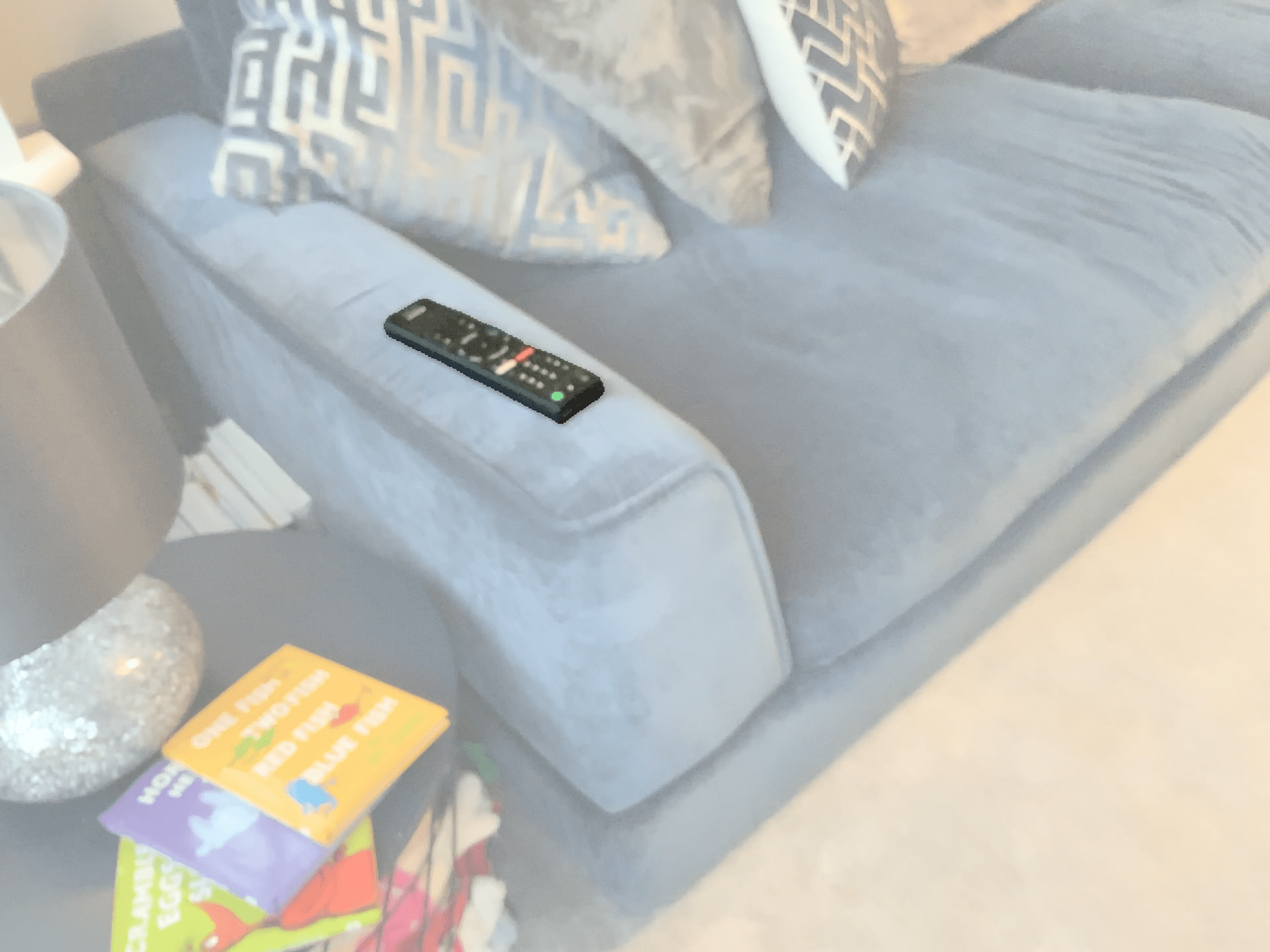}
    
    \end{overpic}
\end{minipage}
\hfill
\begin{minipage}[t]{0.23\textwidth}
\vspace{0pt}
    \begin{overpic}[width=1\textwidth]{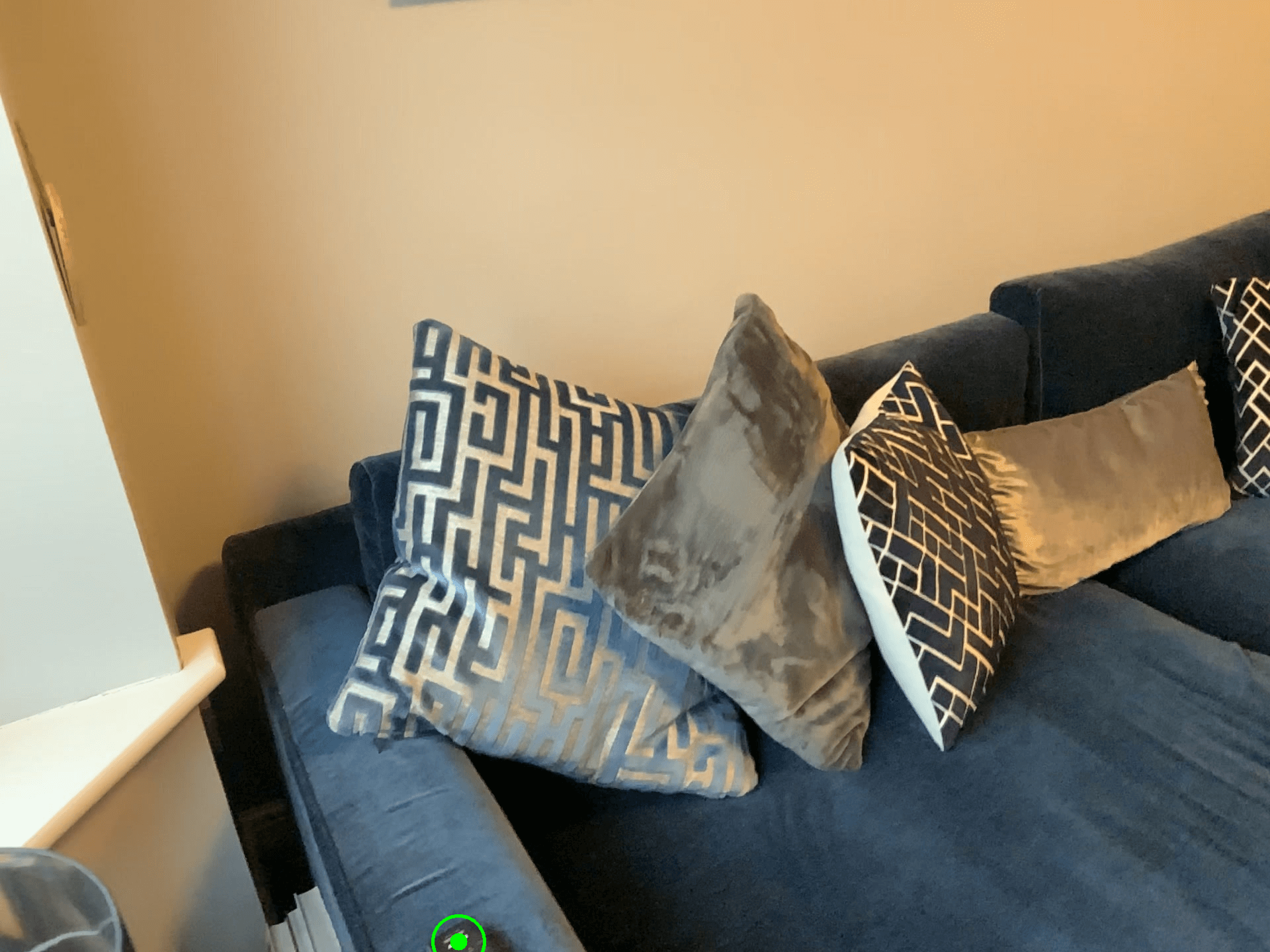}
    
    \end{overpic}
\end{minipage}
\begin{minipage}[t]{0.23\textwidth}
\vspace{0pt}
    \begin{overpic}[width=1\textwidth]{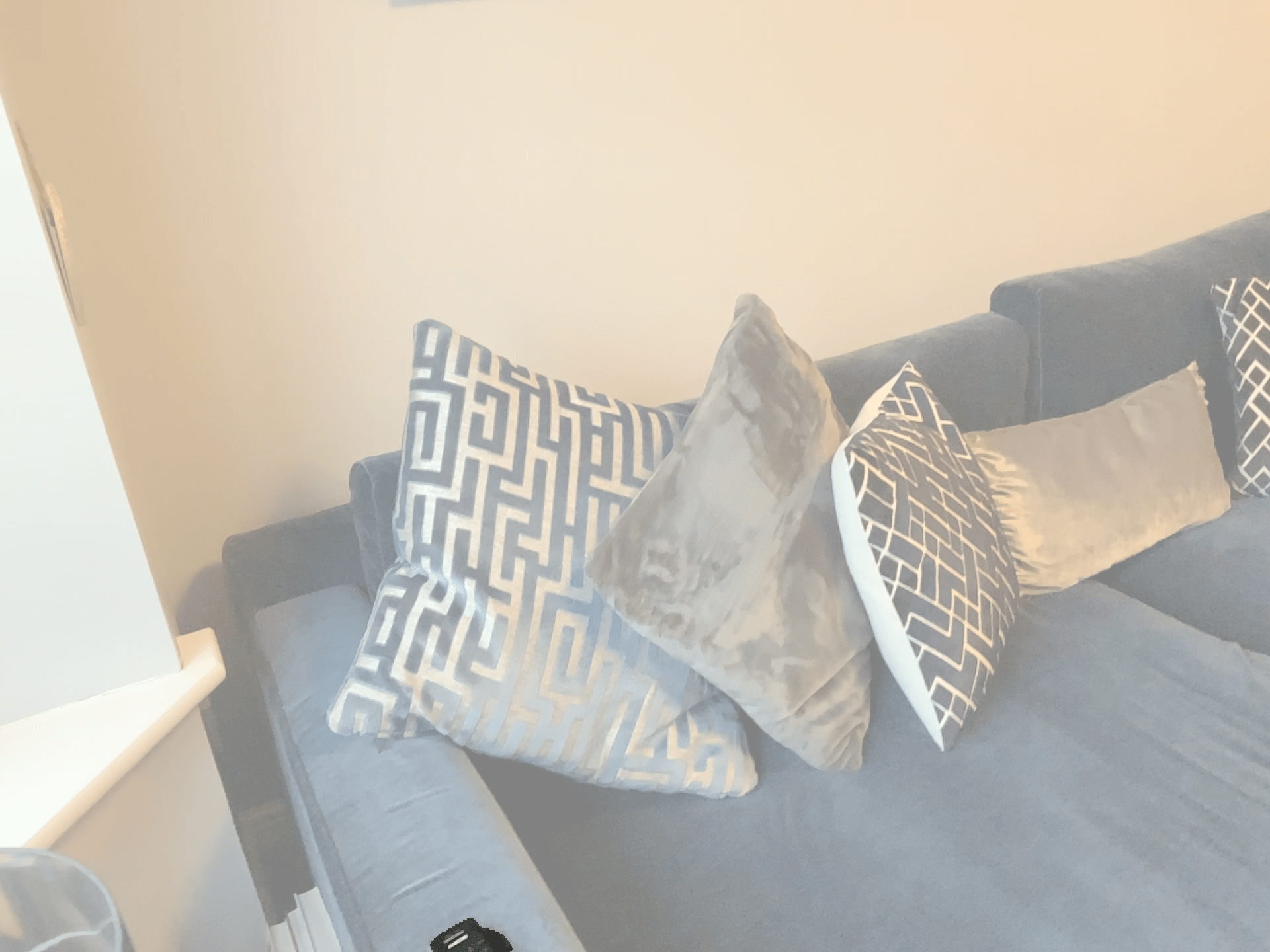}
    
    \end{overpic}
\end{minipage}\\[5mm]
%%%%%%%%%%%%%%%%%%%%%%%%%%%%%%%%%5
% 466916 c49b3990-80da-49e1-b6ec-a5ae7b651504 Adjust the power of the humidifier
%%%%%%%%%%%%%%%%%%%%%%%%%%%%%%%%%
\begin{minipage}[t]{\textwidth}
\vspace{0pt}
    \centering{\texttt{Adjust the power of the humidifier}}
\end{minipage}\\
\begin{minipage}[t]{0.23\textwidth}
\vspace{0pt}
    \begin{overpic}[width=1\textwidth]{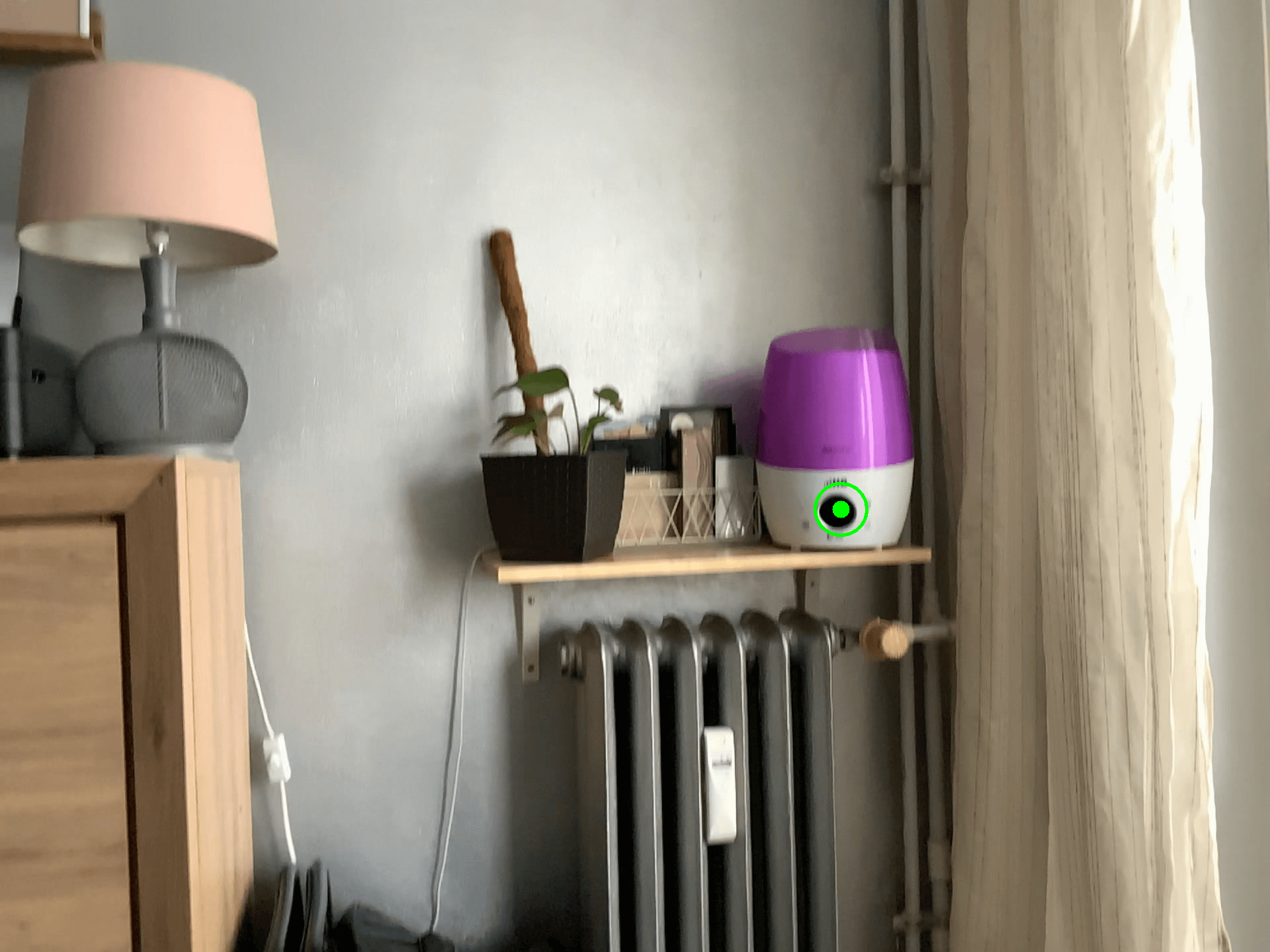}
    
    \end{overpic}
\end{minipage}
\begin{minipage}[t]{0.23\textwidth}
\vspace{0pt}
    \begin{overpic}[width=1\textwidth]{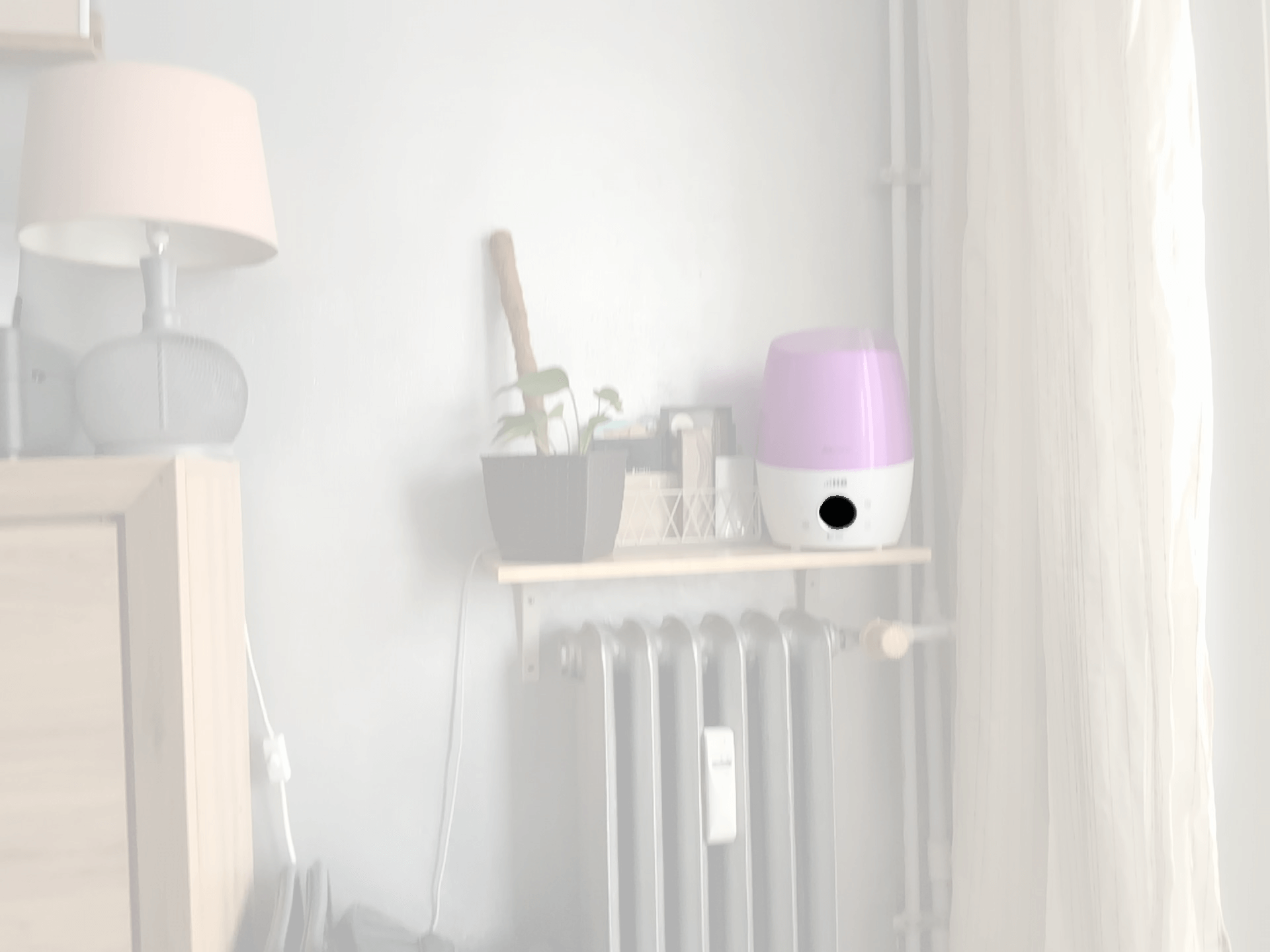}
    
    \end{overpic}
\end{minipage}
\hfill
\begin{minipage}[t]{0.23\textwidth}
\vspace{0pt}
    \begin{overpic}[width=1\textwidth]{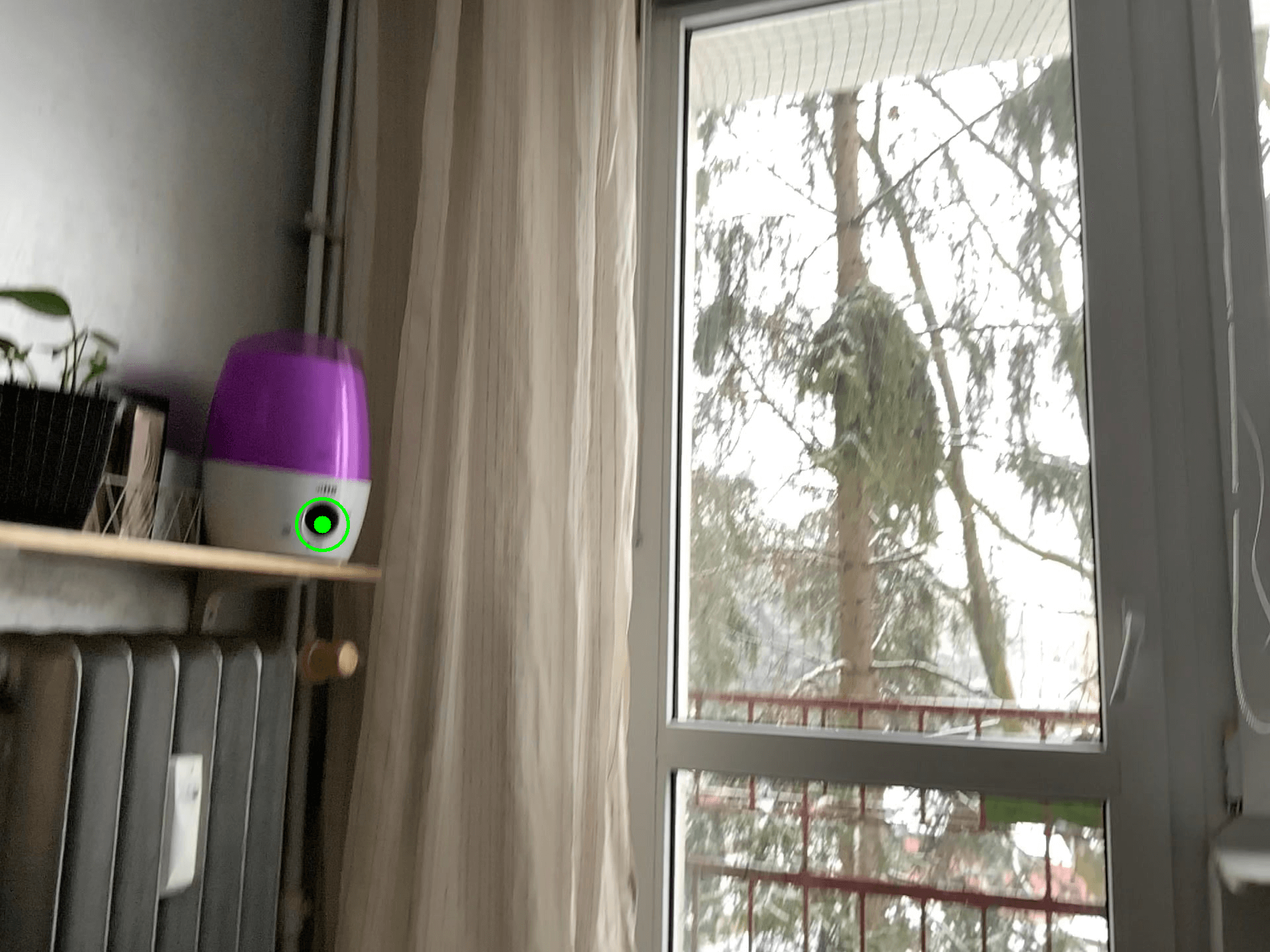}
    
    \end{overpic}
\end{minipage}
\begin{minipage}[t]{0.23\textwidth}
\vspace{0pt}
    \begin{overpic}[width=1\textwidth]{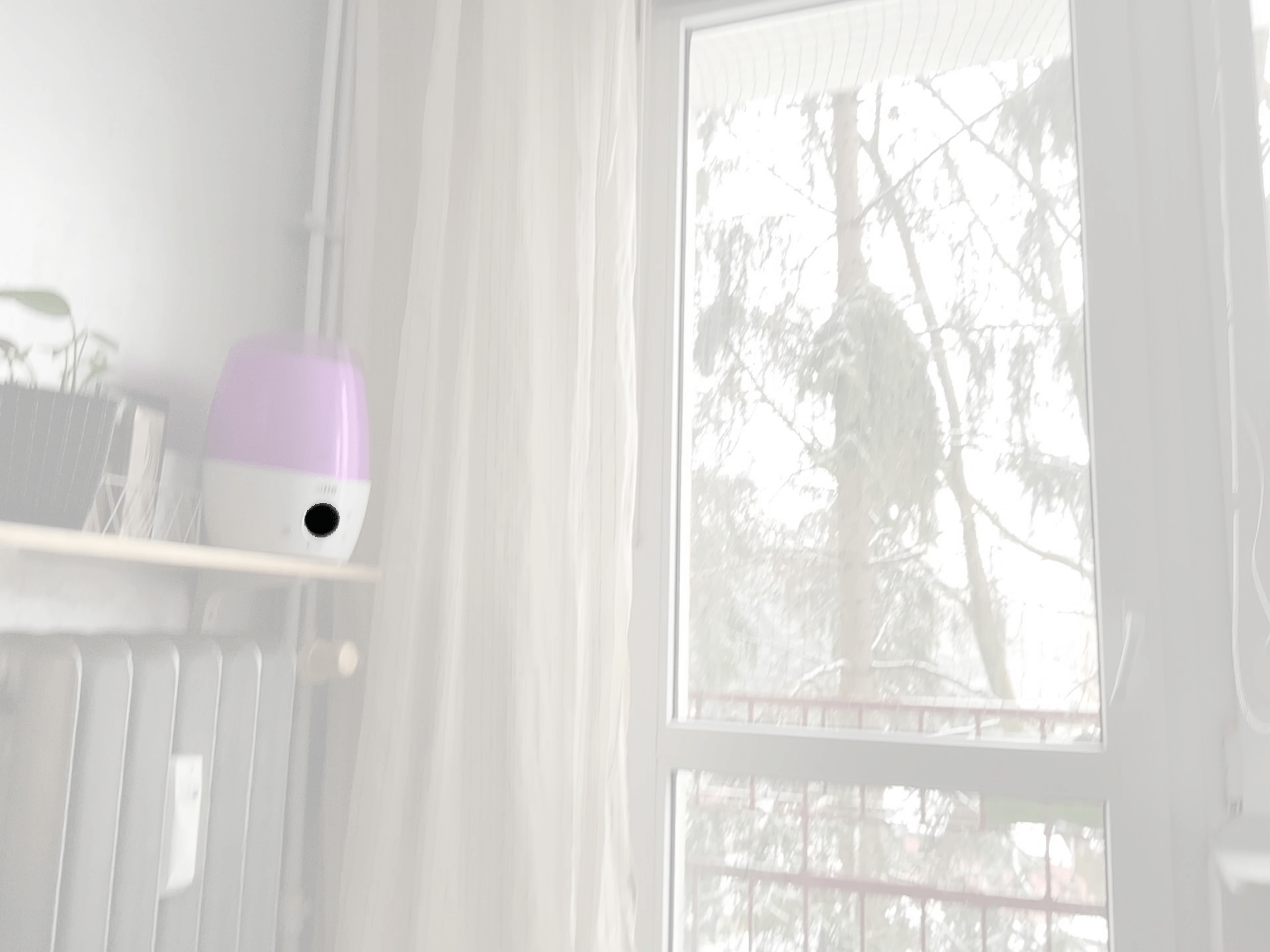}
    
    \end{overpic}
\end{minipage}

%% file: supp/figures/qualit_3d/qualitative_split0.tex
\begin{figure*}[t]
    \centering
    \vspace*{20mm}
    \begin{overpic}[width=\textwidth]{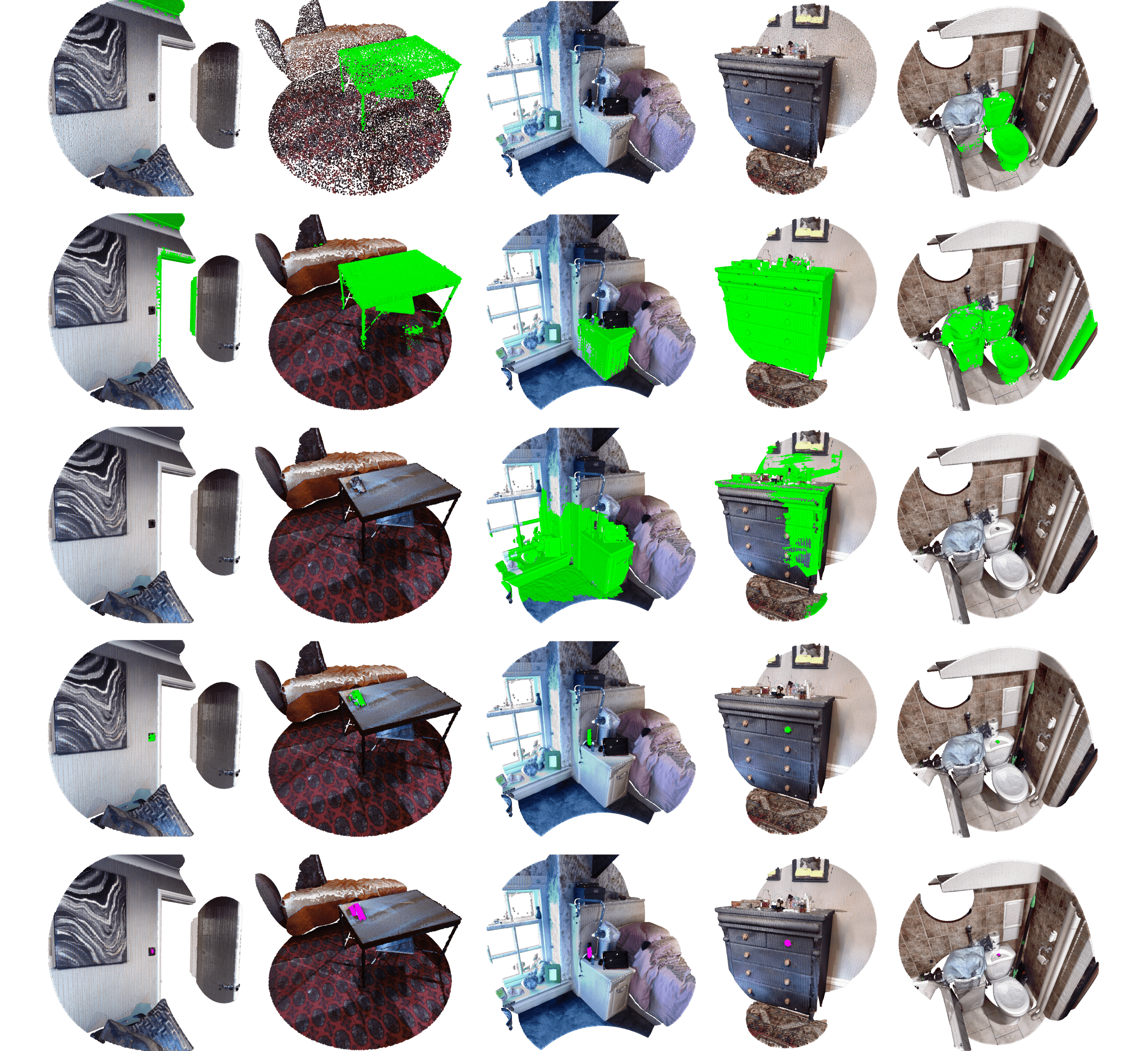}

        % OpenMask
        \miu{\OPxi}{\OPyopenmask}{\ \ 0.00}{\ \ 0.00}{\ \ 0.00}
        \miu{\OPxii}{\OPyopenmask}{\ \ 3.62}{100.00}{\ \ 3.62}
        \miu{\OPxiii}{\OPyopenmask}{\ \ 0.00}{\ \ 0.00}{\ \ 0.00}
        \miu{\OPxiiii}{\OPyopenmask}{\ \ 0.00}{\ 0.00}{\ \ 0.00}
        \miu{\OPxiiiii}{\OPyopenmask}{\ \ 0.73}{\ 98.55}{\ \ 0.73}
        
        % openins
        \miu{\OPxi}{\OPyopenins}{\ \ 0.00}{\ \ 0.00}{\ \ 0.00}
        \miu{\OPxii}{\OPyopenins}{\ \ 2.53}{100.00}{\ \ 2.53}
        \miu{\OPxiii}{\OPyopenins}{\ \ 0.00}{\ \ 0.00}{\ \ 0.00}
        \miu{\OPxiiii}{\OPyopenins}{\ \ 0.19}{\ 96.29}{\ \ 0.19}
        \miu{\OPxiiiii}{\OPyopenins}{\ \ 0.22}{\ 97.85}{\ \ 0.22}

        % lerf
        \miu{\OPxi}{\OPylerf}{\ \ 0.00}{\ \ 0.00}{\ \ 0.00}
        \miu{\OPxii}{\OPylerf}{\ \ 0.00}{\ \ 0.00}{\ \ 0.00}
        \miu{\OPxiii}{\OPylerf}{\ \ 0.46}{100.00}{\ \ 0.46}
        \miu{\OPxiiii}{\OPylerf}{\ \ 0.04}{\ 98.27}{\ \ 0.04}
        \miu{\OPxiiiii}{\OPylerf}{\ \ 0.00}{\ \ 0.00}{\ \ 0.00}

        % ours
        \miu{\OPxi}{\OPyfun}{\ 29.53}{100.00}{\ 29.53}
        \miu{\OPxii}{\OPyfun}{\ 70.35}{\ 82.51}{\ 61.22}
        \miu{\OPxiii}{\OPyfun}{\ 75.78}{\ 91.72}{\ 70.85}
        \miu{\OPxiiii}{\OPyfun}{\ 99.63}{\ 65.84}{\ 65.68}
        \miu{\OPxiiiii}{\OPyfun}{\ 81.37}{\ 70.43}{\ 60.65}

    \end{overpic}

    \vspace{-2mm}
    \caption{
    Qualitative examples of \acronym and the baselines on \valid of SceneFun3D~\cite{delitzas2024scenefun3d}.
    Point clouds are cropped around the functional object for better visualization. We report mask-level Precision (Prc), Recall (Rec), and IoU.
    }
    \label{fig:qualit_3d_split0}
\end{figure*}

%% file: supp/figures/qualit_3d/qualitative_split1.tex
\begin{figure*}[t]
    \centering
    \vspace*{20mm}
    \begin{overpic}[width=\textwidth]{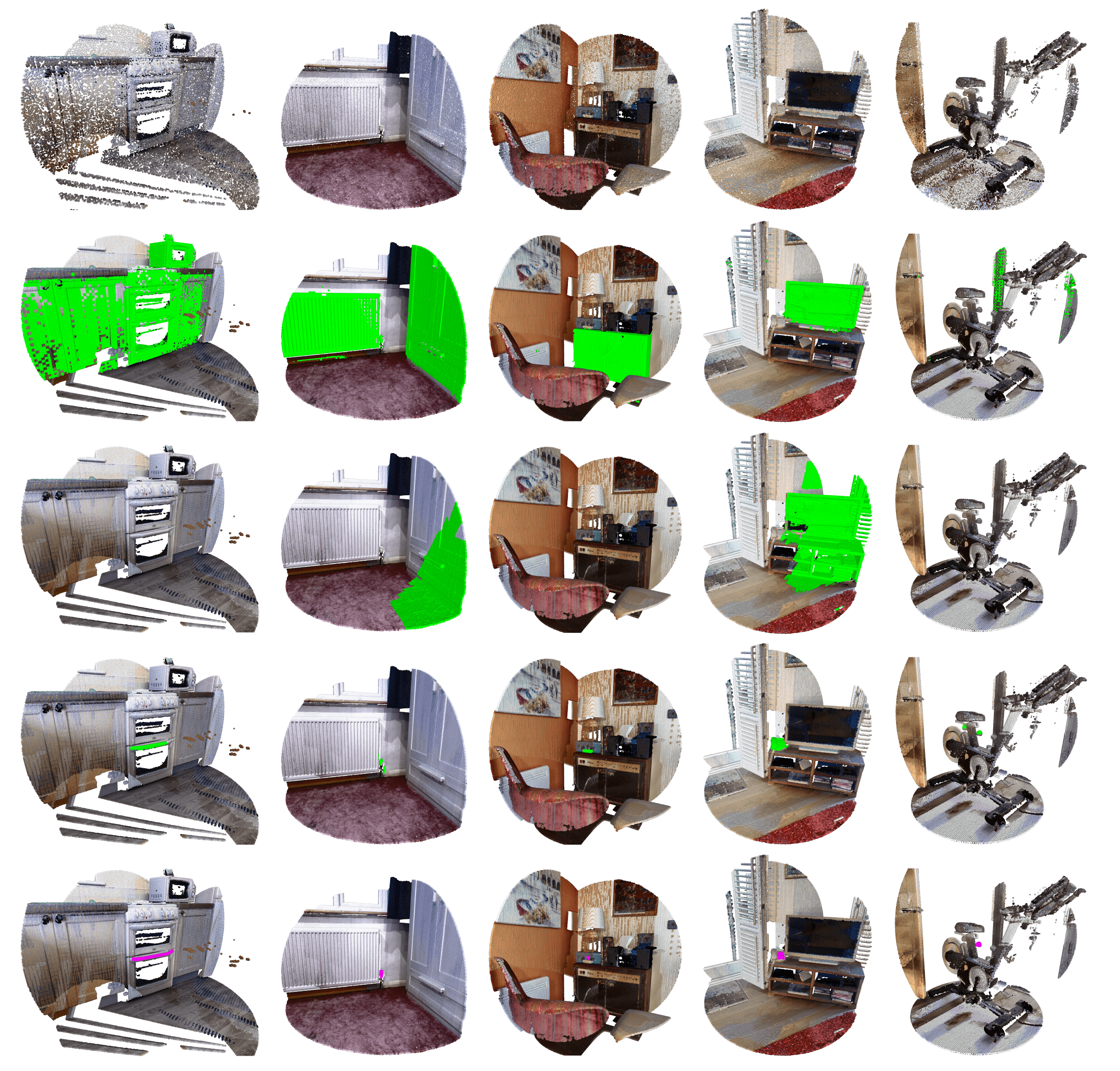}
        % Side names
        \put(0,7){\rotatebox{90}{Ground truth}}
        \put(0,22){\rotatebox{90}{\acronym (ours)}}
        \put(0,44){\rotatebox{90}{LERF \cite{kerr2023lerf}}}
        \put(0,62){\rotatebox{90}{OpenIns3D \cite{huang2023openins3d}}}
        \put(0,80){\rotatebox{90}{OpenMask3D \cite{takmaz2024openmask3d}}}

        % prompts
        \put(4,105){\OPprompt{ Open the bottom door of the oven}}
        \put(24,105){\OPprompt{ Adjust the room's temperature using the radiator dial next to the closet}}
        \put(43,105){\OPprompt{ Open the blue case next to the audio system}}
        \put(61,105){\OPprompt{ Dial a number on the telephone next to the TV}}
        \put(78,105){\OPprompt{ Adjust the height of the seat on the exercise bike}}

        % OpenMask
        \miu{\OPxi}{\OPyopenmask}{\ \ 0.00}{\ \ 0.00}{\ \ 0.00}
        \miu{\OPxii}{\OPyopenmask}{\ \ 0.00}{\ \ 0.00}{\ \ 0.00}
        \miu{\OPxiii}{\OPyopenmask}{\ \ 0.00}{\ \ 0.00}{\ \ 0.00}
        \miu{\OPxiiii}{\OPyopenmask}{\ \ 0.00}{\ \ 0.00}{\ \ 0.00}
        \miu{\OPxiiiii}{\OPyopenmask}{\ \ 0.00}{\ \ 0.00}{\ \ 0.00}
        
        % openins
        \miu{\OPxi}{\OPyopenins}{\ \ 0.41}{\ 79.17}{\ \ 0.41}
        \miu{\OPxii}{\OPyopenins}{\ \ 0.09}{\ 68.36}{\ \ 0.09}
        \miu{\OPxiii}{\OPyopenins}{\ \ 0.00}{\ \ 0.00}{\ \ 0.00}
        \miu{\OPxiiii}{\OPyopenins}{\ \ 0.00}{\ \ 0.00}{\ \ 0.00}
        \miu{\OPxiiiii}{\OPyopenins}{\ \ 0.00}{\ \ 0.00}{\ \ 0.00}

        % lerf
        \miu{\OPxi}{\OPylerf}{\ \ 0.00}{\ \ 0.00}{\ \ 0.00}
        \miu{\OPxii}{\OPylerf}{\ \ 0.00}{\ \ 0.00}{\ \ 0.00}
        \miu{\OPxiii}{\OPylerf}{\ \ 0.00}{\ \ 0.00}{\ \ 0.00}
        \miu{\OPxiiii}{\OPylerf}{\ \ 0.02}{\ 16.30}{\ \ 0.02}
        \miu{\OPxiiiii}{\OPylerf}{\ \ 0.00}{\ \ 0.00}{\ \ 0.00}

        % ours
        \miu{\OPxi}{\OPyfun}{\ 49.81}{\ 70.76}{\ 41.30}
        \miu{\OPxii}{\OPyfun}{\ 57.66}{\ 55.86}{\ 39.61}
        \miu{\OPxiii}{\OPyfun}{\ \ 0.00}{\ \ 0.00}{\ \ 0.00}
        \miu{\OPxiiii}{\OPyfun}{\ 18.21}{\ 99.69}{\ 18.20}
        \miu{\OPxiiiii}{\OPyfun}{\ 54.42}{\ 69.97}{\ 44.12}

    \end{overpic}

    \vspace{-2mm}
    \caption{
    Qualitative examples of \acronym and the baselines on \train of SceneFun3D~\cite{delitzas2024scenefun3d}.
    Point clouds are cropped around the functional object for better visualization. We report mask-level Precision (Prc), Recall (Rec), and IoU.
    }
    \label{fig:qualit_3d_split1}
\end{figure*}

%% file: supp/figures/qualit_3d/qualitative_big.tex
%%%%%%%%%%%%%%%%%%%%%%%%%%%%%%%%%5
% 421254 Open the bottom drawer of the nightstand with the red table lamp on top
%%%%%%%%%%%%%%%%%%%%%%%%%%%%%%%%%
\begin{minipage}[c]{\textwidth}
    \vspace{0pt}
    \centering{\texttt{Dial a number on the telephone next to the TV}}
\end{minipage}\\[2mm]
\begin{minipage}[c]{0.72\textwidth}
\vspace{0pt}
    \begin{overpic}[width=1\textwidth]{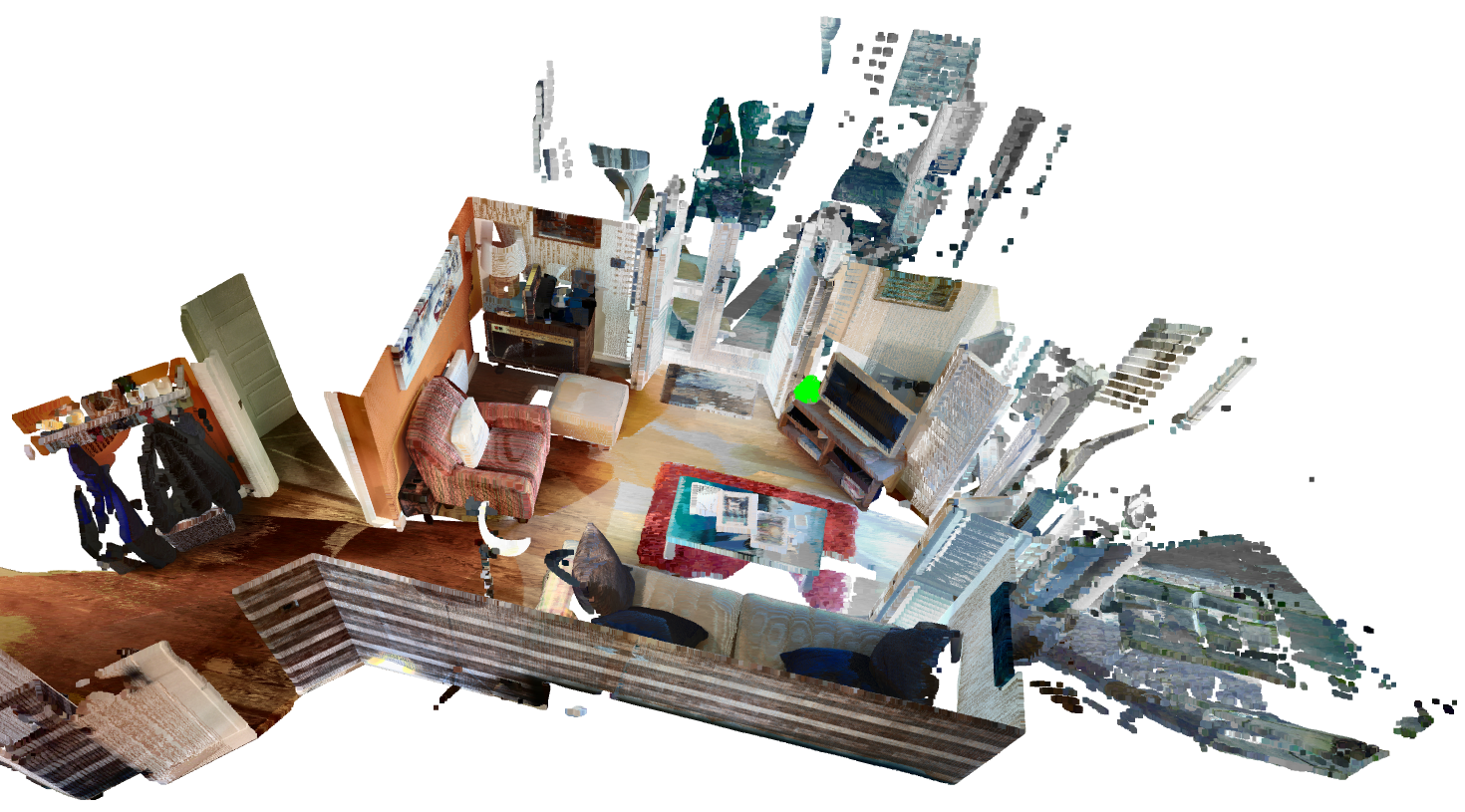}
    \put(58,27){%
        \begin{tikzpicture}[overlay]
            \draw[red, thick] (0,0) circle (20pt);
        \end{tikzpicture}
    }
    \end{overpic}
\end{minipage}
\begin{minipage}[c]{0.27\textwidth}
\vspace{0pt}
    \begin{overpic}[width=1\textwidth]{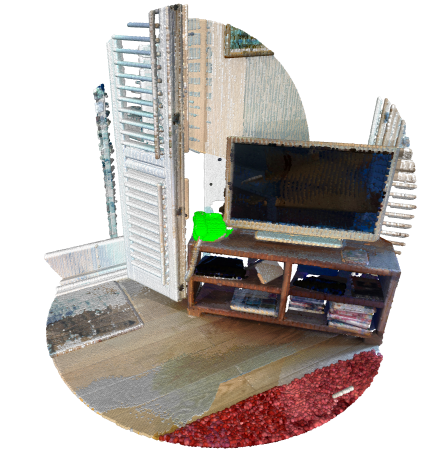}

    \end{overpic}
\end{minipage}\\[5mm]

\begin{minipage}[c]{\textwidth}
    \vspace{0pt}
    \centering{\texttt{ Open the bottom door of the oven}}
\end{minipage}\\[2mm]
\begin{minipage}[c]{0.72\textwidth}
\vspace{0pt}
    \begin{overpic}[width=1\textwidth]{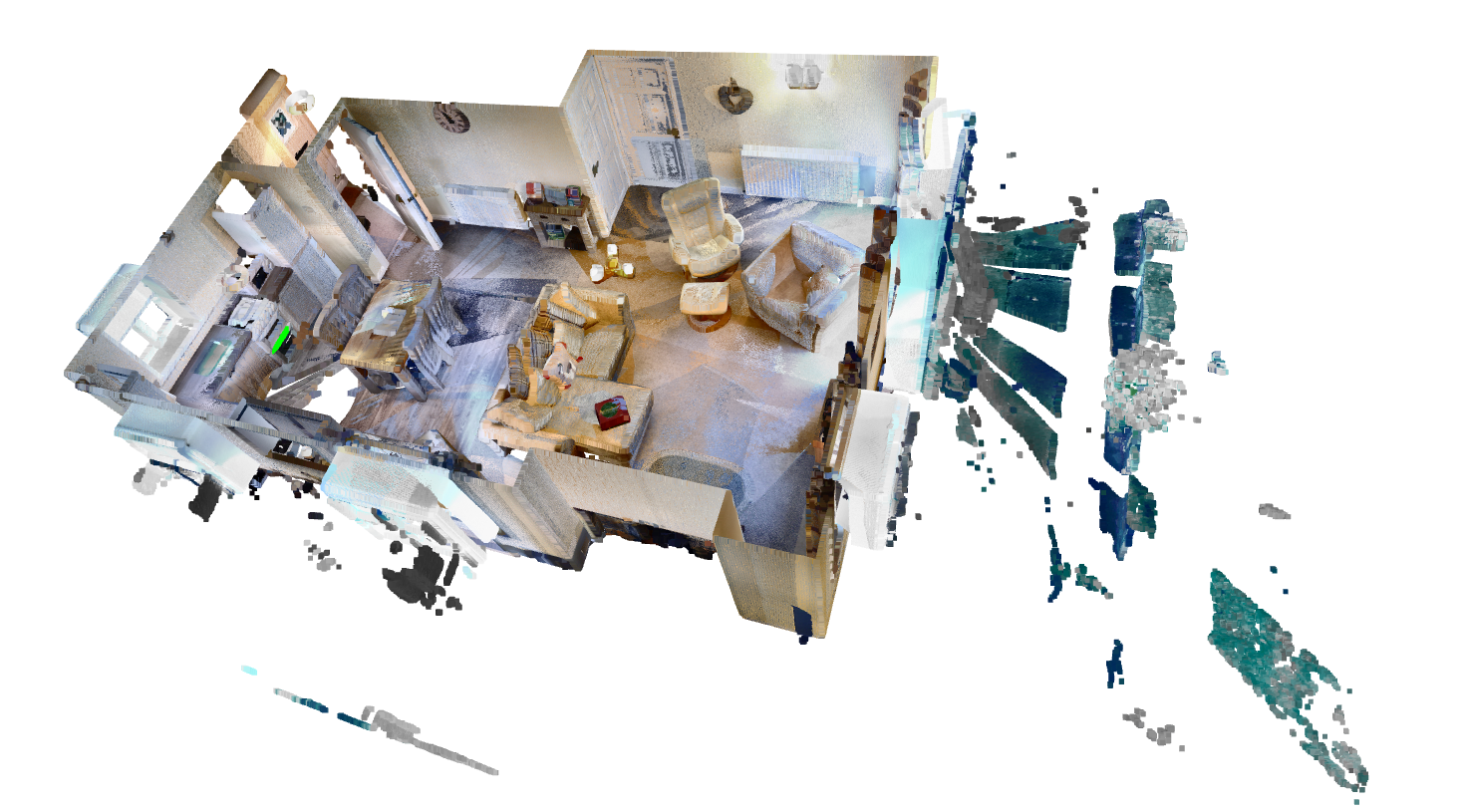}
    \put(18,32){%
        \begin{tikzpicture}[overlay]
            \draw[red, thick] (0,0) circle (20pt);
        \end{tikzpicture}
    }
    \end{overpic}
\end{minipage}
\begin{minipage}[c]{0.27\textwidth}
\vspace{0pt}
    \begin{overpic}[width=1\textwidth]{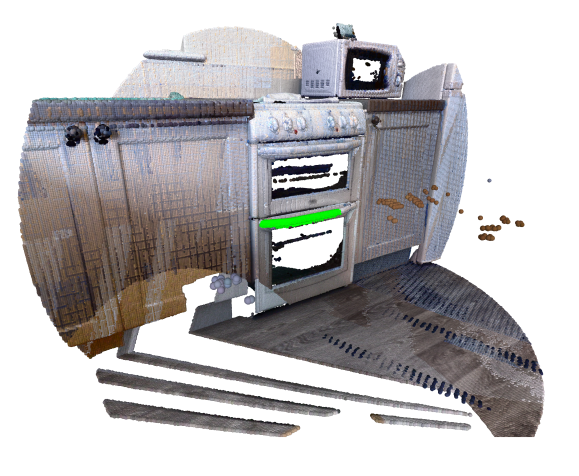}
    
    \end{overpic}
\end{minipage}

%% file: supp/sections/3_implementation.tex
\section{Implementation details}
\label{sec:details}
We implement \acronym in Pytorch and carry out all experiments on an NVIDIA A100 GPU.
We use the public HuggingFace models for Owlv2\footnote{\url{https://huggingface.co/google/owlv2-base-patch16-ensemble}}, SAM\footnote{\url{https://huggingface.co/jadechoghari/robustsam-vit-large}} and Molmo\footnote{\url{https://huggingface.co/allenai/Molmo-7B-D-0924}}.